\documentclass[journal]{IEEEtran}
\usepackage{amsmath,amsfonts}
\usepackage{graphicx}
\usepackage[T1]{fontenc}
\usepackage{tabularx}
\usepackage{subfigure}
\usepackage{wrapfig}
\usepackage{pbox}
\usepackage{bbding}
\usepackage{textcomp}
\usepackage{xcolor}
\usepackage[noend]{algorithm2e}
\usepackage{bbm}
\usepackage{dblfloatfix}
\usepackage{pifont}
\usepackage{bm}

\ifodd 1
\else

\fi

\def\BibTeX{{\rm B\kern-.05em{\sc i\kern-.025em b}\kern-.08em
    T\kern-.1667em\lower.7ex\hbox{E}\kern-.125emX}}
\usepackage{balance}

\newcommand{\name}{t-READi\xspace} %double-check text

\begin{document}
\title{\name: Transformer-Powered Robust and Efficient Multimodal Inference for Autonomous Driving}

\author{
    Pengfei Hu,~\IEEEmembership{}
    Yuhang Qian,~\IEEEmembership{}
    Tianyue Zheng,~\IEEEmembership{}
    Ang Li,~\IEEEmembership{}
    Zhe Chen\IEEEauthorrefmark{1},~\IEEEmembership{}\\
    Yue Gao,~\IEEEmembership{Fellow, IEEE,}
    Xiuzhen Cheng,~\IEEEmembership{Fellow, IEEE,}
    Jun Luo~\IEEEmembership{Fellow, IEEE}

\IEEEcompsocitemizethanks{
    \IEEEcompsocthanksitem
    Yuhang Qian, Pengfei Hu and Xiuzhen Cheng are with School of Computer Science and Technology at Shandong University, China. Email: yhqian@mail.sdu.edu.cn, \{phu,xzcheng\}@sdu.edu.cn, 
    \IEEEcompsocthanksitem
    Tianyue Zheng is with Department of Computer Science and Engineering at Southern University of Science and Technology, China. Email: zhengty@sustech.edu.cn
    \IEEEcompsocthanksitem
    Ang Li is with Department of Electrical and Computer Engineering at University of Maryland College Park, US. Email: angliece@umd.edu
    \IEEEcompsocthanksitem
    Zhe Chen and Yue Gao are with Institue of Space Internet, and School of Computer Science at Fudan University, China. Email: \{zhechen, gao\_yue\}@fudan.edu.cn
    \IEEEcompsocthanksitem
    Jun Luo is with College of Computing and Data Science, Nanyang Technological University, Singapore. Email: junluo@ntu.edu.sg
    \IEEEcompsocthanksitem \IEEEauthorrefmark{1} Corresponding author
}

% \thanks{Manuscript created October, 2020; This work was developed by the IEEE Publication Technology Department. This work is distributed under the \LaTeX \ Project Public License (LPPL) ( http://www.latex-project.org/ ) version 1.3. A copy of the LPPL, version 1.3, is included in the base \LaTeX \ documentation of all distributions of \LaTeX \ released 2003/12/01 or later. The opinions expressed here are entirely that of the author. No warranty is expressed or implied. User assumes all risk.}

}

\markboth{IEEE Transactions on Mobile Computing,~Vol.~XX, No.~XX, September~202X}%
{How to Use the IEEEtran \LaTeX \ Templates}

\maketitle

\begin{abstract}
    Given the wide adoption of multimodal sensors (e.g., camera, lidar, radar) by \textit{autonomous vehicle}s (AVs), deep analytics to fuse their outputs for a robust perception become imperative. However, existing fusion methods often make two assumptions rarely holding in practice: i) similar data distributions for all inputs and ii) constant availability for all sensors. Because, for example,  
    % However, the two assumptions rarely hold in practice (e.g., different cameras and 
    lidars have various resolutions and failures of radars may occur, such variability 
    % The varied sensing data will reshape the distribution and modalities of input, and hence 
    often results in significant performance degradation in fusion.
    %
    % The multimodal sensors, such as camera, lidar, mmWave radar, etc., have been widely applied in autonomous vehicles, and hence pretrained deep neural networks (DNNs) are deployed to fuse them for enabling robust perception at a large scale. The current mainstream multimodal data processing approaches always hold two assumptions, which both the inference input as well as training samples follow the similar data distribution, and all the sensors work well all the time. However, these two assumptions are rare to hold in practice. For example, cameras on autonomous vehicles may capture surroundings under various light conditions, and the lidars  may also generate point clouds with varied sampling rates. In addition, the vehicles may not be equipped with all the sensors. In such cases, the varied sensing data will reshape the distribution and modalities of input, and hence result in the significant performance degradation of DNNs.
    % The mainstream multimodal data processing uses a pretrained deep neural network (DNN) model, which can achieve desired performance with assuming that inference input and training samples follow the independent and identically distribution.
    % However, this assumption is not always true in reality: the cameras on autonomous vehicles may capture surroundings with various light conditions; lidar may also sample points at different rates. In such cases, the varied sensing data will reshape the distribution of input, and hence result in the performance degradation of DNN models. 
    %
    To this end, we present \name, an adaptive inference system that accommodates the variability of multimodal sensory data and thus enables robust and efficient perception.
    \name identifies variation-sensitive yet \textit{structure-specific} model parameters; it then adapts only these parameters while keeping the rest intact.
    %%%%%%%%
    %%%%%%%% note by yuhang: overwrite the first major contribution. The main difference is there is no explicit ``optimization algorithm" to identify these parameters.
    %%%%%%%%
    % Given an inference latency budget, \name employs an optimization algorithm to identify the variation-sensitive model parameters; it then adapts only those 
    % % variation-sensitive 
    % chosen parameters while keeping the rest 
    % % of model parameters 
    % intact. 
    \name also leverages a cross-modality contrastive learning method to compensate for
    % generate complementary information for 
    the loss from missing modalities. Both functions are implemented to maintain compatibility with
    % \name can seamlessly work with 
    existing multimodal deep fusion methods.
    % (e.g., BEVFusion and TransFusion).
    The extensive experiments evidently demonstrate that compared with the status quo approaches, \name not only improves the average inference accuracy by more than $6\%$ but also reduces the inference latency by almost 15$\times$ with the cost of only 5\% extra memory overhead in the worst case under realistic data and modal variations.
    
    %%%%%%%%
    %%%%%%%% note by yuhang: how do we define the comparison of ``inference accuracy"? since the fully fine-tuned model is slightly better than our adaptive tuning, while when compare to pretrained model, we do boost the inference accuracy greatly (by 15% in a moderate case). And the ``inference latency" is evaluated with the comparison of loading time. And ``extra memory overhead" varies from $2%~6%$, it depends on the layout of specific model (i.e., the overhead of a transformer-intensive framework is much less than that of a resnet basicblock-intensive framework(e.g., ResNet18)).
    %%%%%%%%
    % The extensive experiments on benchmark dataset evidently demonstrate that, compared with the status quo approaches, \name improves the average inference accuracy by more than $6\%$ and reduces the inference latency by almost 15$\times$, at the cost of only 1\% extra memory overhead under realistic data and modal variations. 
\end{abstract}

\begin{IEEEkeywords}
Autonomous vehicle, robust perception, multimodal learning, object detection, semantic segmentation.
\end{IEEEkeywords}

% 1.5 - 2 pages
\section{Introduction}\label{sec:introduction} 
\IEEEPARstart{A}{utonomous} driving, with its worldwide developments~\cite{tesla, waymo, uber}, promises to achieve greater safety, less harmful emissions, increased lane capacity, and reduced travel time~\cite{lin2022tracking,shaheen2019mobility,lin2022channel}. At the core of autonomous driving, the perception capability of \textit{autonomous vehicle}s (AVs) leverages the data collected from various sensors (e.g., camera, lidar, and radar) %and the processing of these data 
to better understand AVs' surrounding~\cite{lv2022edge,fang2024ic3m,zheng2022catch}, 
% world around the vehicle, and 
thus enabling applications such as object detection
% ~\cite{zhao2020fusion, hnewa2020object, arnold2019survey} 
~\cite{zhao2020fusion, hnewa2020object, guan2022deepmix} 
and semantic segmentation
% ~\cite{feng2020deep,ha2017mfnet, chen2018importance}. 
~\cite{feng2020deep,AngelaDai20183DMVJ3, HangSu2018SPLATNetSL}.
To power the perception capability, \textit{deep neural network}s~(DNN) are adopted to process and fuse the multimodal data~\cite{li2022motion,tang2024merit,lin2023fedsn}
% ~\cite{Hong2021MoreDM}
into a unified representation. Therefore, maintaining the effectiveness of these AV-DNN models is crucial to autonomous driving.

The lifecycle of an AV-DNN model starts with its design, then it gets pre-trained model by individual manufacturers, before being deployed to respective AVs equipped multimodal sensors for performing perception-related inferences. 
% and trained in the cloud. Then, the pre-trained DNN model is deployed in AVs directly to perform inference, resorting to the  power of edge computing. 
Although DNN-based multimodal fusion for on-device inference in general has 
% become a new and hot research topic paid 
attracted attention from both academia~\cite{grigorescu2020cloud2edge, lin2023pushing,guo2018cloud,lin2024split} and industry (e.g., Google~\cite{lee2019device}, as well as Intel and Ford~\cite{han2021putting}), only part of them have targeted multimodal DNN for autonomous driving (e.g., \cite{liu2022bevfusion,Bai2022TransFusionRL}). Since this latter batch of proposals largely focuses on DNN architecture design at the manufacturer side, 
% such as BEVFusion~\cite{liu2022bevfusion} and TransFusion~\cite{Bai2022TransFusionRL}, most of them 
they often disregard the per-vehicle variations in terms of input data and modality, 
% thus their application to real-world applications remain questionable. 
% focus on network architecture design to improve the performance on respective tasks. 
resulting in a \textbf{missing link}, pertaining to in-vehicle inference, between DNN models and their practical adoptions.

\begin{figure}[t]
    \centering
    \includegraphics[width=\linewidth]{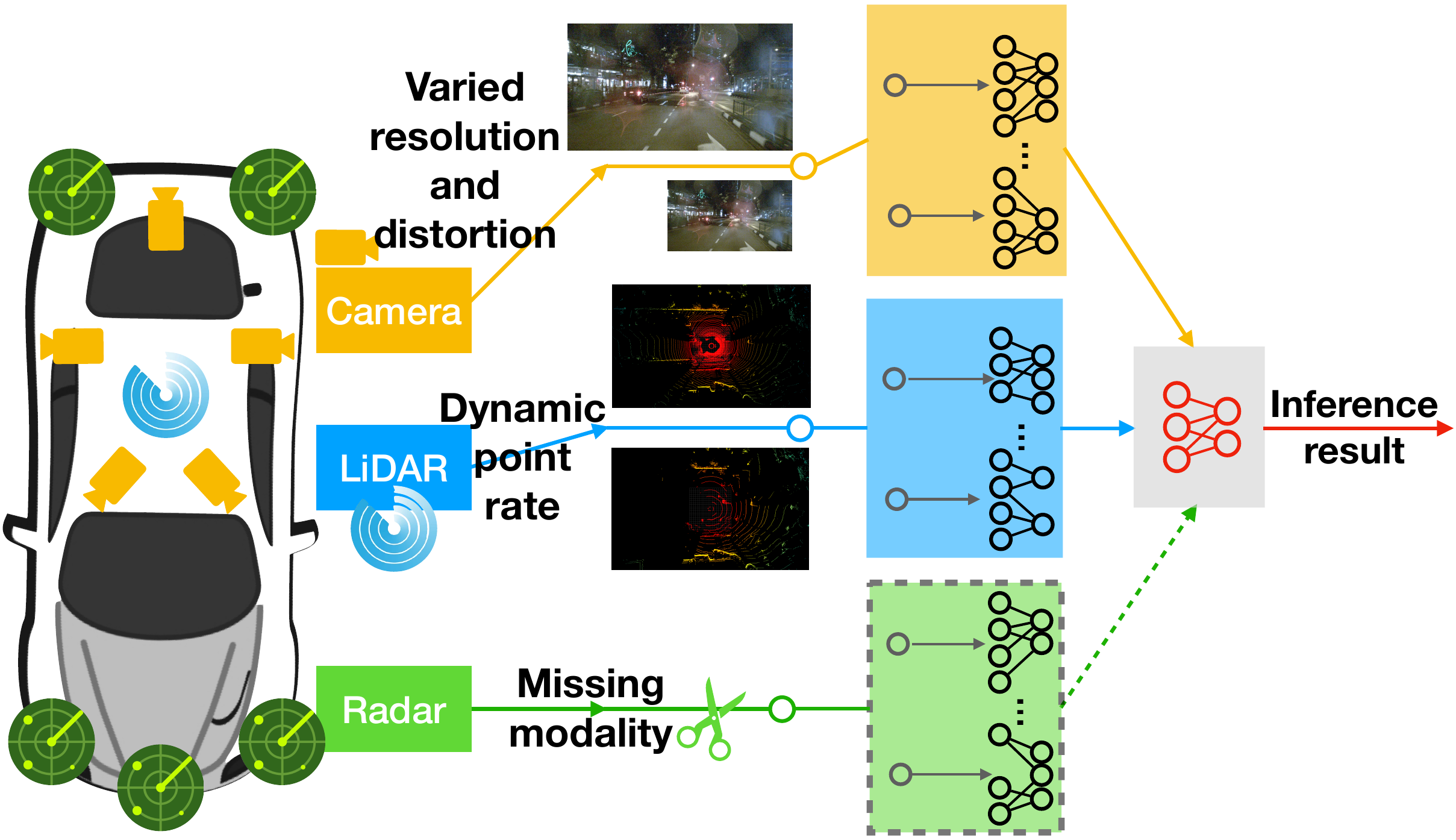}
    \caption{The variations in data and modality.}
    \label{fig:overview}
    %\vspace{-1ex}
\end{figure}
One critical issue with existing proposals is the fixed set of multimodal sensors and environment conditions used for model training. 
% However, in practical scenarios, 
In practice, both sensor modalities and environmental conditions may experience drastic variations~\cite{chen2021octopus,hu2023holofed,lin2022v2i,chen2019m}.
% cannot remain constant. 
For example, different lighting conditions may force the camera to change its exposure, different velocities can result in motion blurs to various degrees, and adverse weather conditions (e.g., rain, fog, and snow) often scatter laser light, thus forcing lidar to perform more intensive scans. Consequently, those two modalities introduce more drastic variations.
Moreover, sensor occlusion and/or malfunction may lead to missing modalities that compromise the multimodal fusion. All these realistic scenarios yield variations in the input data and modality, as illustrated in Figure~\ref{fig:overview}. They often significantly degrade the performance of well-designed DNN models, causing erroneous decision-making and even traffic accidents.
%%%%%%%%%%%%%%%%
% \yuhang{In a nutshell, if high-level autonomous driving is to become a reality, (extra-system) various adverse weather conditions as well as the (intra-system) sensors misfunction should be tackled cautiously.}
% Hence, various adverse exterior conditions as well as the interior sensors malfunction should be tackled cautiously, \rev{as survey~\cite{Texas_2023} points out, adverse weather conditions and faulty sensors are practical common causes of autonomous car accidents, such as Tesla Model 3 accident in 2019~\cite{TWP_2023}.}
%%%%%%%%%%%%%%%%
% Obviously, the rigid DNN model designs in these works make it difficult to maintain good performance that relies heavily on consistent data distributions during the training and inference stages. 
% As a naive solution to mend for the performance degradation in reality, one may pre-train a large set of models targeted for different scenarios and frequently switch among these models during runtime in response to respective data and modality variations. On one hand, this solution cannot be \textit{robust} against the varying sensor and environment conditions, since such varying processes happen in continuous form while we may only have a countable number of models.
% As a naive solution to mend for the performance degradation in reality, one may pre-train a large set of models for different scenarios and frequently switch among these models during runtime in response to respective data and modality variations. 
Therefore, it is crucial to address adverse external conditions and internal sensor malfunctions carefully. As highlighted in the survey by \cite{Texas_2023}, adverse weather conditions and faulty sensors are common contributors to autonomous vehicle accidents, such as the 2019 Tesla Model 3 incident~\cite{TWP_2023}. A naive solution to mitigate performance degradation is to pre-train a large set of models for different scenarios and switch among them during runtime based on the available data and modality variations.
% On one hand, this solution cannot be \textit{robust} against the varying sensor and environment conditions, since such varying processes happen in continuous form while we may only have a countable number of models.
However, this solution is not \textit{robust} against continuously varying sensor and environment conditions since we may only have a countable number of models.
%is and the scarce on-device resource makes it impossible for the system to take care of every possible scenario.
% On the other hand, the solution fails to be \textit{efficient} either, because storing and (re-)loading a large number of model versions (each with many parameters) incur excessive memory footprints and intolerable inference latency. 
Besides, model switching incurs excessive memory footprints and intolerable inference latency. 

%More importantly, the sensor's failure in  the vehicle may occur during driving. Missing modalities always impair the performance of the  DNN fusion model, resulting in wrong decision-making and traffic accidents. Unfortunately,  most existing DNN models do not consider this challenging scenario. %just as the naive solution, it is inefficient to store all the pre-trained DNN models and load them dynamically during runtime to deal with the enormous amount of data and modality variations.  %On one end of the spectrum, we can train a large number of models to cover all possible scenarios, thus achieving maximum robustness. Contrariwise, we can directly employ the model pre-trained by automotive manufacturers without any modifications to achieve maximum efficiency. 

Whereas, it is non-trivial to realize robust and efficient multimodal inference for autonomous driving. First of all, as 
% it is challenging to achieve multimodal inference that strikes a balance between 
robustness and efficiency are often at odds, 
% Although we can achieve maximum robustness or efficiency by employing numerous fine-tuned models or single pre-train models, respectively, how to 
designing a multimodal system that copes with variations without consuming excessive resources remains an uncharted area. 
% Secondly, even if one aims to strike a balance between these two objectives, 
% efficiency and robustness under certain conditions, the constraints of the optimization problem can vary greatly in practice, thus making 
% Secondly, even if one aims to strike a balance between these two objectives, choosing the right tradeoff can be subjected to varying conditions. For example, the latency requirements for DNN models in high-speed cruising and low-speed maneuvering scenarios are different because of the DNN responsiveness needed to match the vehicle speed. 
Secondly, it's hard to strike an adaptive balance between these two objectives since it depends on continuously varying conditions. For instance, the latency requirements for DNN models in high-speed cruising and low-speed maneuvering scenarios are different because of the DNN responsiveness needed to match the vehicle speed. 
Last but not least, since DNN models embedded in vehicles are typically compiled low-level codes optimized for specific DNN architecture~\cite{liu2022bevfusion,Bai2022TransFusionRL,Yu2021AnIE}, it is largely impossible to modify the architecture in response to, for example, missing modalities and in turn their sensory data as part of the input.
% not preferable to directly make changes to the models for better efficiency and robustness. Instead, we should try seek novel approaches (e.g., manipulate the system in terms of training data and dynamic runtime loading). 

%Last but not least, missing modalities unseen during training may cause the DNN models to exhibit unexpected and erroneous inference behaviors. While a method that improves DNN's robustness to missing modalities is urgently needed, the method should not compromise the efficiency of DNN inference.%most existing DNN models lack robustness to missing modalities. Generally speaking, DNN models are trained with a fixed number of working modalities, and missing modalities unseen during training may cause the DNN models to exhibit unexpected and erroneous inference behaviors. %thus causing erroneous decision-making and even resulting in traffic accidents. 
%%%%%%%%
%%%%%%%%note by yuhang: in the newest version, we don't model it as a knapsack problem any longer (which means this is no corresponding ``optimization algorithm" and formulated ``memory constraint" anymore (``memory constraint" driven ``optimization").), but we continue to generate various modality variations.
%%%%%%%%

To address these challenges, we propose \name (i.e., \underline{t}ransformer-powered \underline{R}obust and \underline{E}fficient multimodal inference for \underline{A}utonomous \underline{D}r\underline{i}ving) as a novel AV inference system; it adaptively accommodates the variations in multimodal sensory data and missing modalities. 
Specifically, \name employs a variation-aware model adaptation algorithm to handle data and modality variations under memory constraints while producing multiple variants of models deviated from the pre-trained model. For each variant, we exploit the parameter-efficient fine-tuning techniques of Large Language Models (LLMs)~\cite{hu2021lora}, but extend to widely-used modules in AV-DNN models in addition to transformer modules in LLMs~\cite{qiu2024ifvit,fang2024automated,lin2024splitlora}. For example, \name injects trainable rank decomposition matrices into residual blocks leading to a significantly decreased number of trainable parameters. With multiple variants of models and the pre-trained model, we can switch different variants according to current input indications, such as weather conditions. 

% To be specific, \name handles those variations with a \newrev{robust and efficient \textit{unified} framework} while producing multiple variants of the pre-trained DNN. For each variant, \name identifies only a small portion of the \textit{inplace} neurons sensitive to data/modality variations along with \textit{injects} transparent tiny structures (less than $6\%$ of all neurons in worst cases). This leads to a substantial reduction in memory footprints, hence allowing for loading all these variants into memory for the purpose of robust inference. \needrev{Besides the merits of \textit{robust} and (storage) \textit{efficient}, it is more \textit{agile}, since revising DNN is much more time \textit{efficient} and alleviating data-hungry dependency to a great extent.}
Moreover, \name leverages a contrastive learning framework to overcome the issue of missing modalities, while maintaining a comparable performance as using complete modalities. 
In particular, by contrasting the samples with and without a potential missing modality, \name drives the latent representation under missing modality towards semantically correlated with that under full modalities; this renders the DNN model robust to missing modalities without altering its architecture. Our key contributions can be summarized as follows:

\begin{itemize}
    \item We design \name to address the variations of sensory data and modality in autonomous driving so that \name adapts to various run-time environments with both robust and efficient manners.
    % and integrates with multimodal DNN models seamlessly. 
    %\item \textcolor{red}{Our \name does not need to modify the architectures of the current multimodal DNN models, but only runs on top of them.}
    %
    % \item We propose an optimization framework to perform model adaptations with sparse parameter updates. The technique minimizes memory footprints, hence allowing for loading a large number of model variations into memory and in turn removing extra latency caused by model reloading.
    %%%%%%%%%%%%%%%%
    \item We propose a variation-aware model adaptation algorithm to perform model adaptations with sparse parameter updates. The technique minimizes memory footprints, hence allowing for loading a large number of model variations into memory and in turn eliminating extra latency caused by model reloading.
    %%%%%%%%
    %%%%%%%% note by yuhang: what I mean with ``unified" in this context is, we tune exist/injected parameters in determined places, with determined method, which is not a process of ``optimization".
    %%%%%%%%
    %%%%%%%%%%%%%%%%
    % thus tackling input data variation caused by sensor parameter change and varying environments efficiently and robustly.
    %
    \item We design a cross-modal contrastive learning method to compensate for input data loss due to missing modalities. It avoids runtime modifications on model architecture and thus preserves the overall efficiency.
    %%%%%%%%
    %%%%%%%%note by yuhang: I will dive into some details of ``contrastive learning" to make sure these paragraphs are not abrupt.
    %%%%%%%%
    % sensors in cases such as sensor malfunction and occlusion without compromising the overall efficiency.
    %
    \item We implement a prototype of \name and evaluate it with extensive experiments. The promising results confirm that \name indeed offers robust and efficient multimodal inference for autonomous driving.
\end{itemize}

Though proposals on multimodal fusion for on-device inference do exist~\cite{grigorescu2020cloud2edge, guo2018cloud,han2021putting}, \name is still the first to design such a system for autonomous driving with practical considerations, i.e., robustness and efficiency in driving environments. The rest of the paper is organized as follows. \S~\ref{sec:background_motivation} introduces the background and motivation. The detailed system design and implementation are described in \S~\ref{sec:design} and \S~\ref{sec:implementation}, respectively. \S~\ref{sec:evaluation} reports the evaluation results. Related works are presented in \S~\ref{sec:related_work}, along with limitations and future directions of \name.
%Section~\ref{ssec:discussion}. 
Finally, \S~\ref{sec:conclusion} concludes our paper. 

% 1 page
\section{Background and Motivation}  \label{sec:background_motivation}

In this section, we %provide a brief background on object detection for autonomous driving, and its corresponding evaluation metrics. Then we 
first investigate the impact of sensors' parameter changes (e.g., lidar and camera) 
% \yuhang{Other types of sensors like ultrasonic sensors~\cite{Aeberhard2015ExperienceRA}, mmWave radars~\cite{Maddern20171Y1, Caesar2020nuScenesAM} are not so well-studied in academia and are beyond the scope of this paper}) 
on DNN inference performance. We then show how missing modalities significantly degrade the performance of DNN for autonomous driving. Finally, we explain why model reloading is not feasible for mitigating these negative impacts due to high inference latency and memory usage.

%%%%%%%%%%%%%%%%
% \noteyuhang{I think the following subsection should be \textbf{reorganized} or \textbf{removed}, since it is very hard to evaluate. But we can reserve some insight of this subsection (e.g., some key parameter of sensor adjust dynamically, which should be investigate and verified by me later.).}
% lidar parameter1: spin rate (rotation rate => spin rate)
% datasheet at https://www.manualslib.com/manual/185826/Velodyne-Hd-Hdl-64e-S2-1.html

\begin{figure*}[t]
    \vspace{1ex}
    \centerline{
        \subfigure[Motion distortion.]{
            \includegraphics[width = .2 \linewidth]{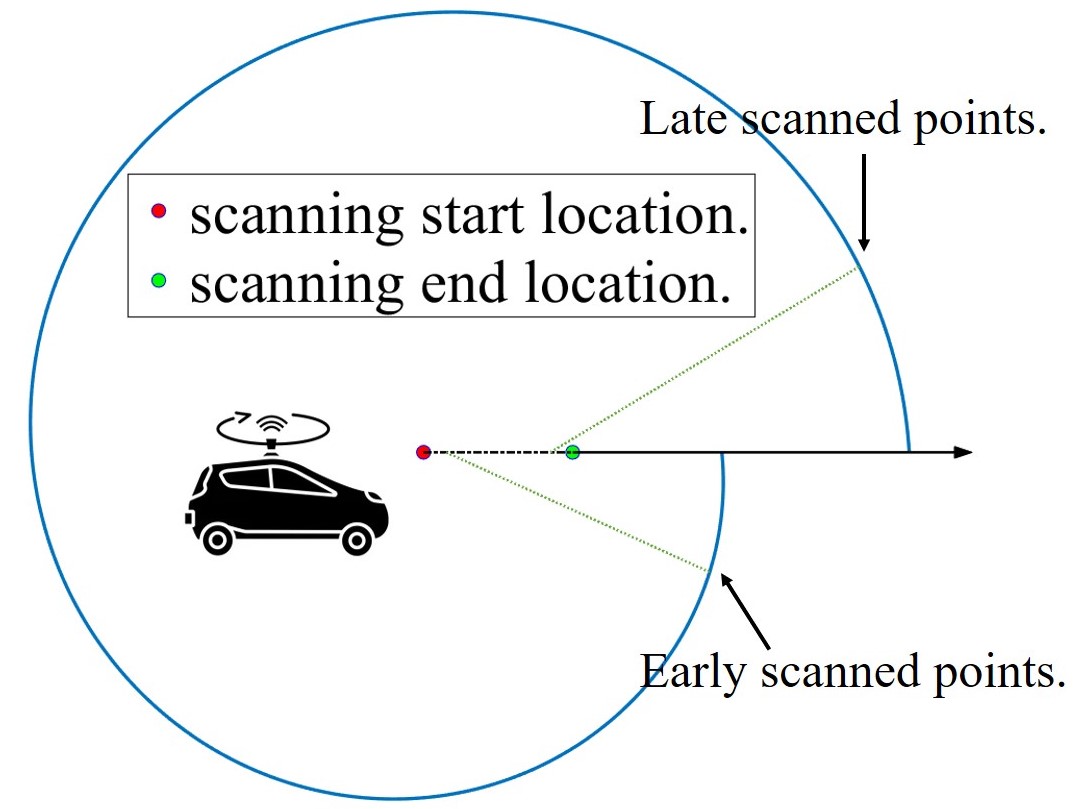}
            \label{subfig: dist_illustrate}
        }
        \subfigure[Reference frame.]{
            \includegraphics[width = .2 \linewidth]{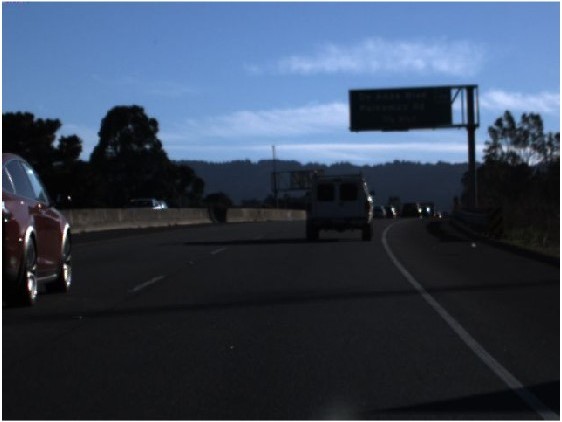}
            \label{subfig: ref_center_cam_frame}
        }
        \subfigure[Delamination effect.]{
            \includegraphics[width = .2 \linewidth]{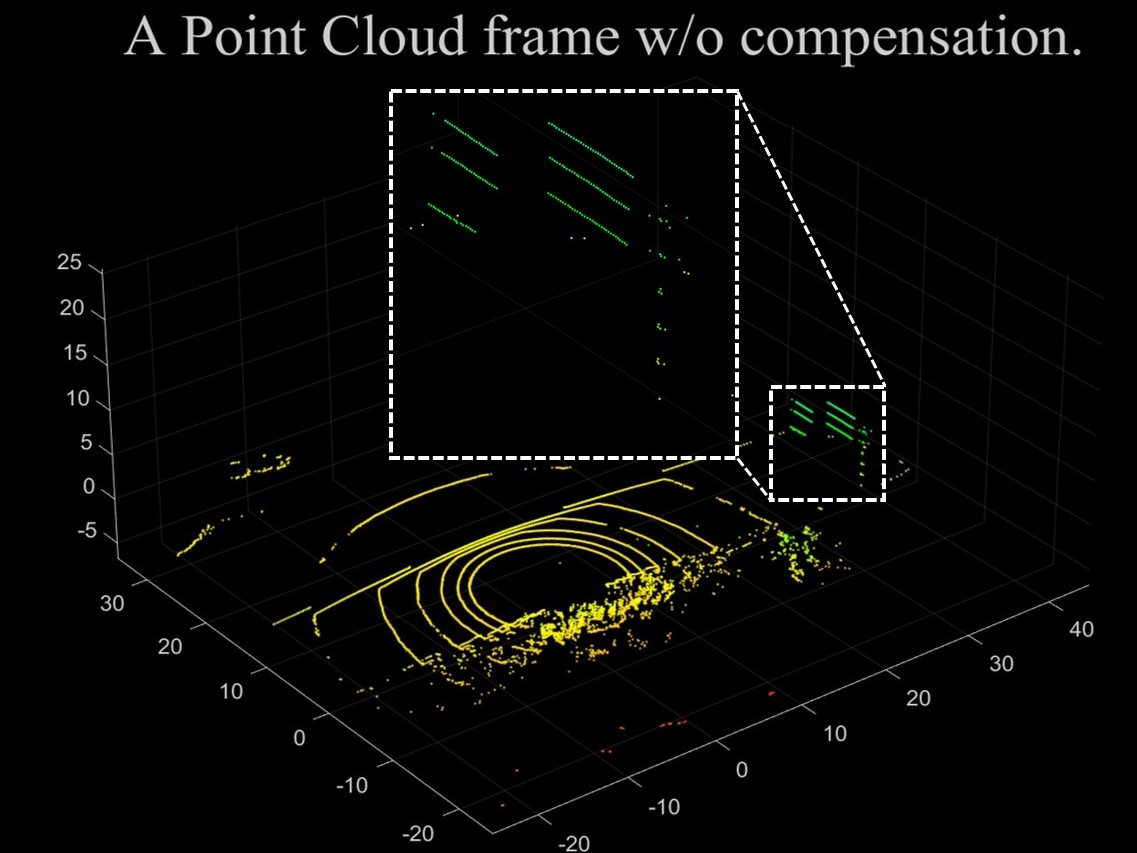}
            \label{subfig: lidar_delaminated}
        }
        \subfigure[Succeed to precept.]{
            \includegraphics[width = .2 \linewidth]{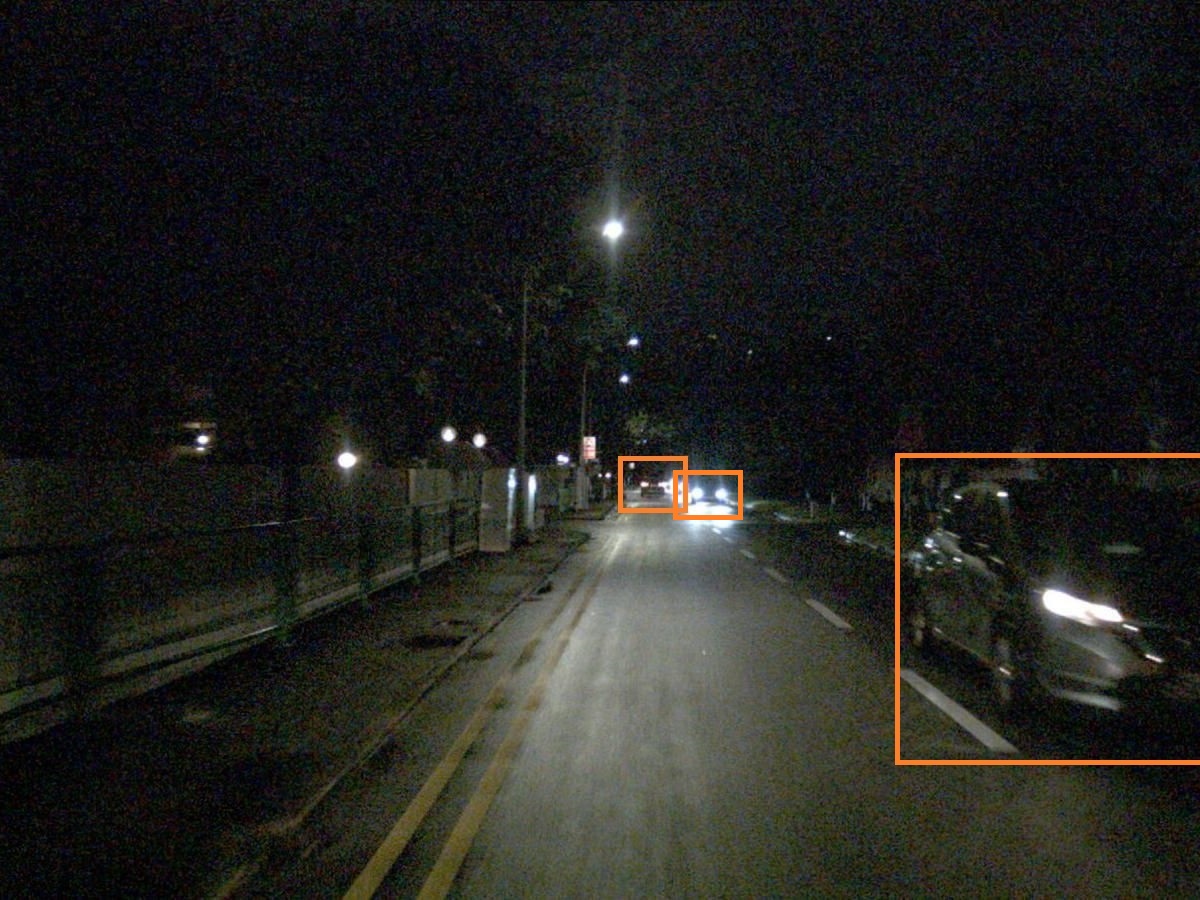}
            \label{subfig: light_flares}
        }
        \subfigure[Fail to precept due to flare.]{
            \includegraphics[width = .2 \linewidth]{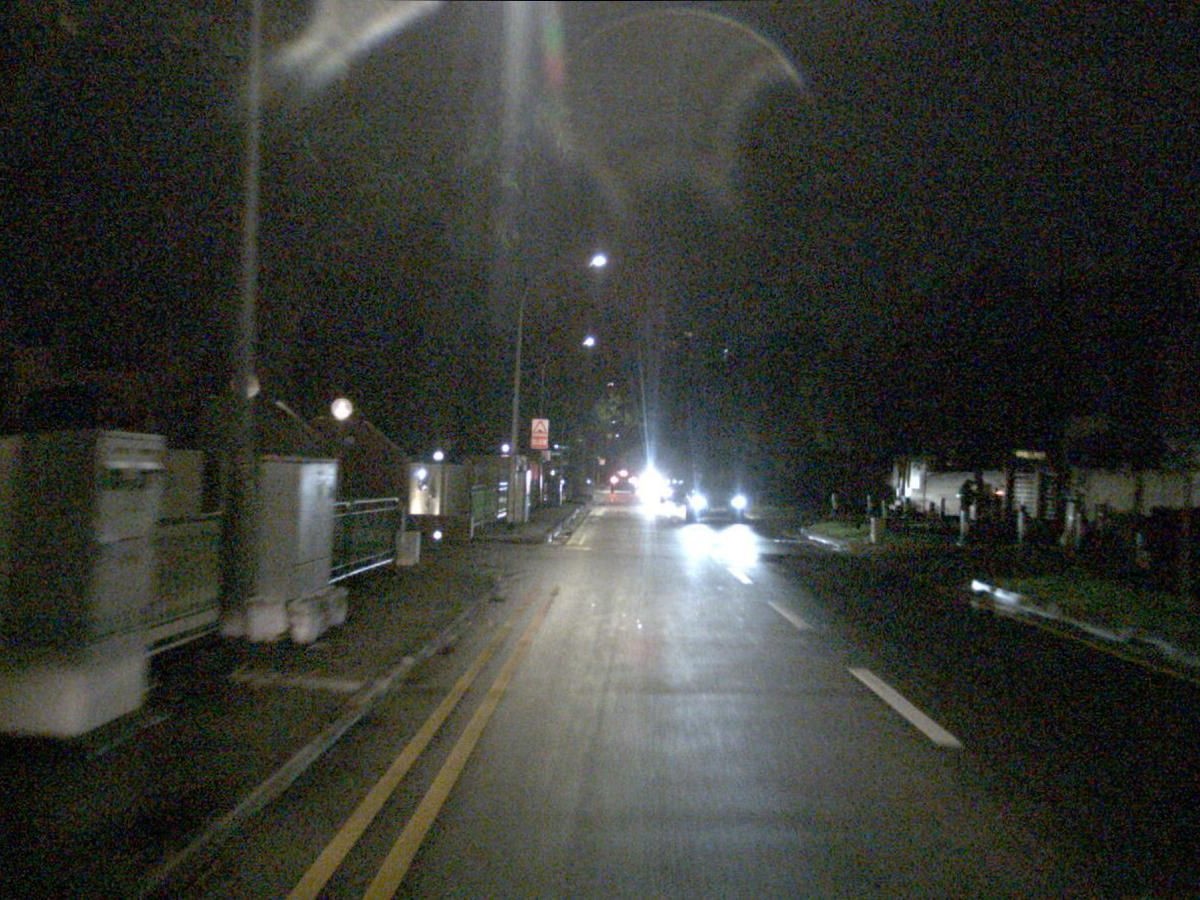}
            \label{subfig: heavy_flares}
        }
    }
    \vspace{-5pt}
	\caption{Sensory data variations dramatically degrade the performance of pre-trained AV-DNN models in reality.~\subref{subfig: dist_illustrate} $\sim$ ~\subref{subfig: lidar_delaminated}: fail to detect the traffic sign due to lidar point cloud variation. ~\subref{subfig: light_flares} $\sim$ ~\subref{subfig: heavy_flares}: fail to detect vehicles due to heavy flare effect on camera.  }
    % Oxford RobotCar~\cite{Maddern20171Y1} andnuScenes~\cite{Caesar2020nuScenesAM}.}
	\label{fig: AV_challenging_scenario}
 
    % Practical AV scenario is challenging, as sensors are not reliable in this cases.
\end{figure*}

\subsection{Sensor Parameter Variation}
\label{ssec:motivation_sensor_variation}

Most existing DNN-powered perception systems use a pre-trained model assuming that inference and training data follow the same probability distribution~\cite{dundar2007learning,liu2022bevfusion,Bai2022TransFusionRL}. However, this assumption does not always hold true under AV scenarios: the camera of an AV may capture video with varying exposure and motion blur; the lidar of the AVs may retrieve data streams with a dynamic density of point cloud. Consequently, the parameter variation will result in a non-iid (independent and identically distributed) distribution that cannot be well handled by the AV-DNN model and hence lead to degraded performance.

Lidar is a vital sensor used in AVs due to its depth estimation capability. Currently, the automotive industry primarily uses rotating lidar sensors, which are powered by mechanical rotation~\cite{lidar_manual}. For lidar sensors, the spinning rate refers to the number of revolutions in a second, and a higher spinning rate results in lower azimuth/elevation angular resolution. For example, when the spinning rate sweeps from $5$~\!Hz to $20$ ~\!Hz, the azimuth angular resolution deteriorates from $0.09^{\circ}$ to $0.36^{\circ}$. However, the pre-setting spinning rate cannot accommodate varying driving speed of the AV on run-time, resulting in a dynamic density of point cloud. The reason is shown in Figure~\ref{subfig: dist_illustrate} that lidar cannot catch point cloud timely, leading to depth difference in each scanning period~(inverse of spinning rate). Figure~\ref{subfig: ref_center_cam_frame} and Figure~\ref{subfig: lidar_delaminated} demonstrate a road sign lies in front of the view, and the reflected lidar points are delaminated significantly, as
the fixed spinning rate cannot catch different driving speeds of the AV.

The camera is another sensor widely used in AVs but much more affordable, which can provide richer semantic information than lidar. However, its performance can be significantly affected by surrounding environments. For example, a high or low intensity of light can cause cameras to malfunction. A common issue is the flare effect. 
Generally, a model's inference performance will degrade for unseen and bad environments with high possibility. As shown in Figure~\ref{subfig: light_flares} and \ref{subfig: heavy_flares}, even though scenarios of two images are similar, the pre-trained object detection model can work well on its training dataset in Figure~\ref{subfig: light_flares}, but fails to detect the target in Figure~\ref{subfig: heavy_flares} due to flare effect.

In a nutshell, input variations of multiple modalities severely affect AV-DNN performance.

\begin{figure}[t]
    \centerline{
        \subfigure[mAP.]{
            \includegraphics[width = .5 \linewidth]{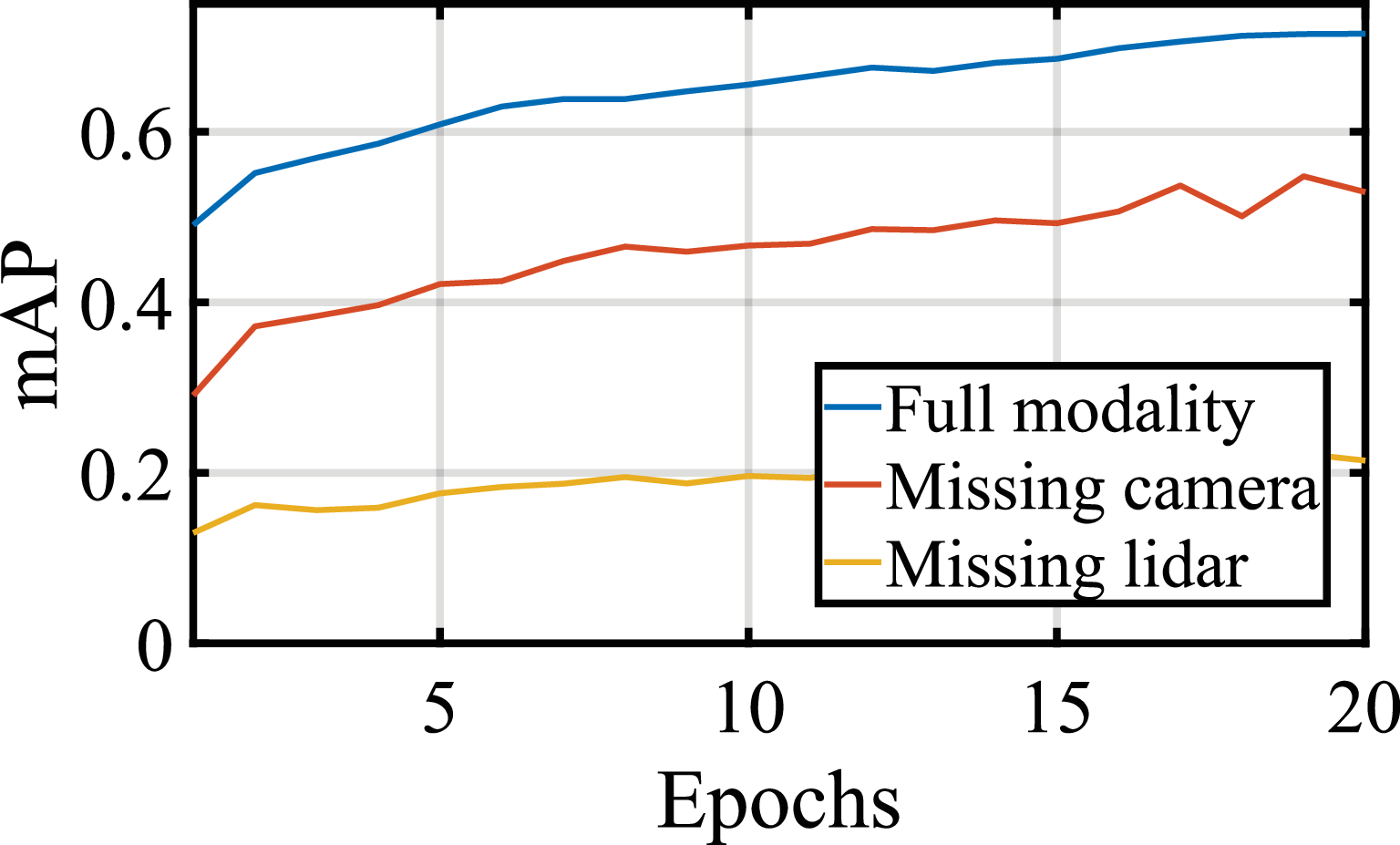}
        }
        \subfigure[NDS.]{
            \includegraphics[width = .5 \linewidth]{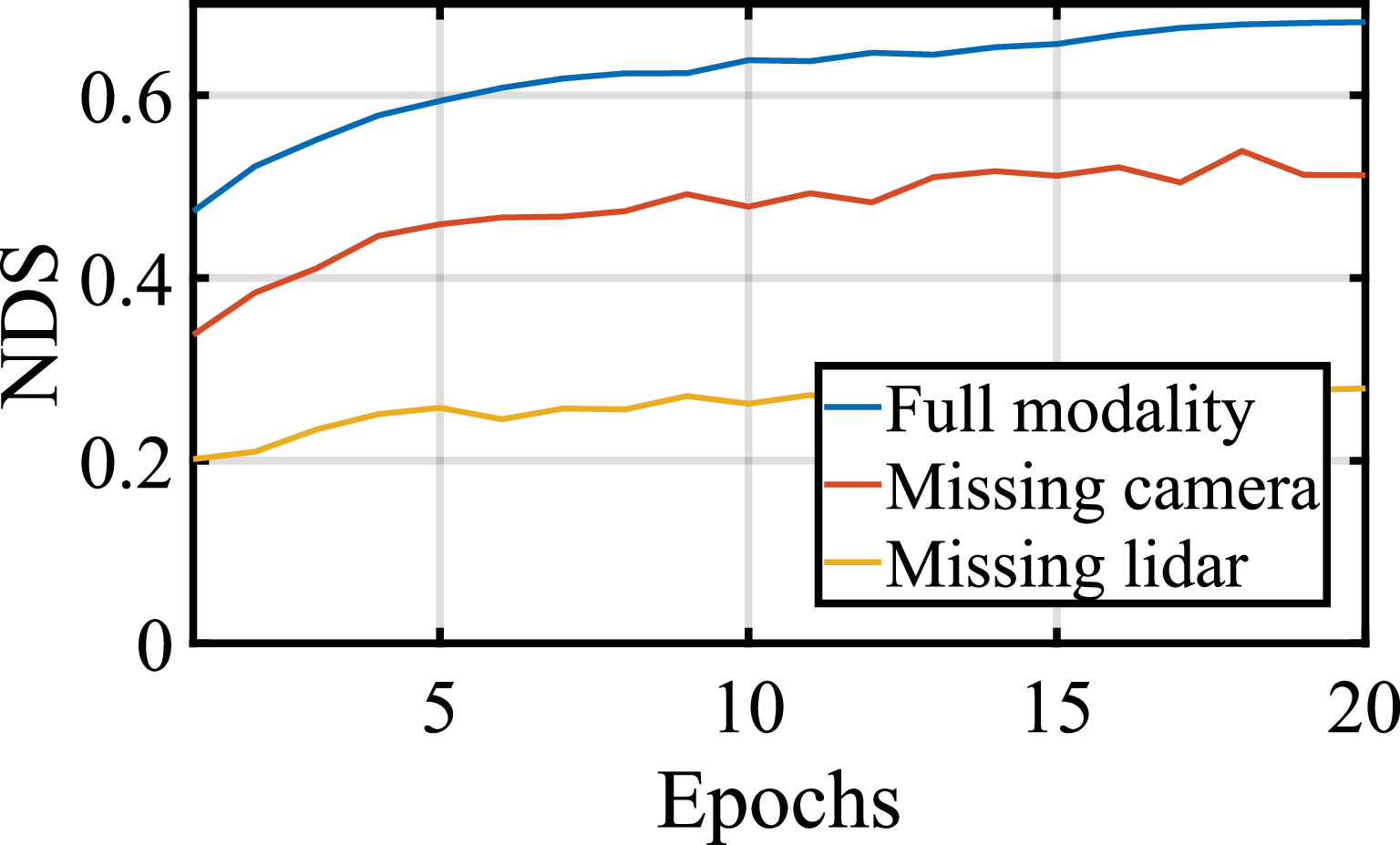}
        }
    }
    \vspace{-5pt}
    \caption{Damaging effects of missing modality.}
    \label{fig:damage_miss_mod}
    \vspace{-5pt}
\end{figure}

\subsection{Missing Modality}
\label{ssec:motivation_missing_modality}
%Now we focus on the modalities of camera and lidar in AV perception tasks. 
% In the realm of multimodal sensing data fusion, most existing research works presume the availability of all modalities during both training and inference stages~\cite{xu2018pointfusion, bijelic2020seeing}. However, this is not always the case, since sensors can malfunction or become obstructed during inference, resulting in absent modalities, \rev{which is proved for RGB-camera modality in~\cite{Ceccarelli2020RGBCF}}. Such failures present significant challenges to DNN-based perception tasks in autonomous driving, primarily due to mismatches between partial input data and the model architecture.
% One common solution is data imputation, which fills missing modalities with predefined numbers, such as zeros~\cite{van2018flexible}. However, doing so incurs the non-iid issue by introducing bias during inference.
In the realm of multimodal sensing data fusion, most existing research presumes the availability of all modalities during both the training and inference stages~\cite{xu2018pointfusion, bijelic2020seeing}. However, this assumption does not always hold true, as sensors can malfunction or become obstructed during inference, leading to missing modalities, as demonstrated for the RGB-camera modality in~\cite{Ceccarelli2020RGBCF}. Such failures pose significant challenges to DNN-based perception tasks in autonomous driving, primarily due to mismatches between the partial input data and the model architecture. A common solution to this problem is data imputation, which involves filling in missing modalities with predefined values, such as zeros~\cite{van2018flexible,Xu2020TouchPassTB}. However, this approach introduces bias during inference, resulting in non-iid issues.

We demonstrate this in Figure~\ref{fig:damage_miss_mod} by showing the mAP~(mean average precision) and NDS~(details
of the metrics will be introduced in \S~\ref{sec:implementation}) of BEVFusion~\cite{liu2022bevfusion} under both full and missing modalities of the nuScene Dataset~\cite{Caesar2020nuScenesAM} with zero-filling. 
%We start experiment and demonstrate this in Figure~\ref{fig:damage_miss_mod}, where train a BEVFusion~\cite{liu2022bevfusion}, an open-sourced state-of-the-art DNN model for AD tasks (more precisely, the object detection task in this context.) from scratch with nuScene Dataset~\cite{Caesar2020nuScenesAM}. 
%Figure~\ref{fig:damage_miss_mod} reveals two metrics, namely mAP and NDS, under both full modalities and missing modalities with zero-filling. Both are represented in percentage bounded by $100\%$ and a higher value is preferred, we reveal more details in Section~\ref{sec:evaluation}. %On the one hand, t
The results reveal that BEVFusion inference with all modalities outperforms inference with missing camera data (filled with zeros) by over 15\%. Moreover, training with missing lidar modality results in significantly lower mAP and NDS, due to the absence of rich geometric information provided by 3D lidar point cloud. %On the other hand, train a multimodal model with a corrupted modality (more precisely, camera modality in this context.) is very challenging and just yield worse performance than train a monomodal model with fine modality (more precisely, lidar modality in this context.), since the performance gap should be much more closer. 
These findings confirm that merely filling missing modalities with zeros falls short. They highlight the need for novel methods that can effectively harness the complementary information from multiple modalities, thereby reducing the performance gap between inference with full and missing modalities.

However, designing an effective mechanism to compensate for missing modalities without additional computational overhead, while maintaining comparable performance to using complete modalities, is challenging. To address this, \name introduces a novel method that effectively utilizes complementary information from multiple modalities to narrow the performance gap between full and missing modalities. This method will be presented in \S~\ref{ssec:contrastive_learning}.

\subsection{Model Reloading is Impractical}
\label{ssec:motivation_reloading}
\begin{figure}[t]
    \centerline{
        \subfigure[Latency.]{
            \includegraphics[width = .5 \linewidth]{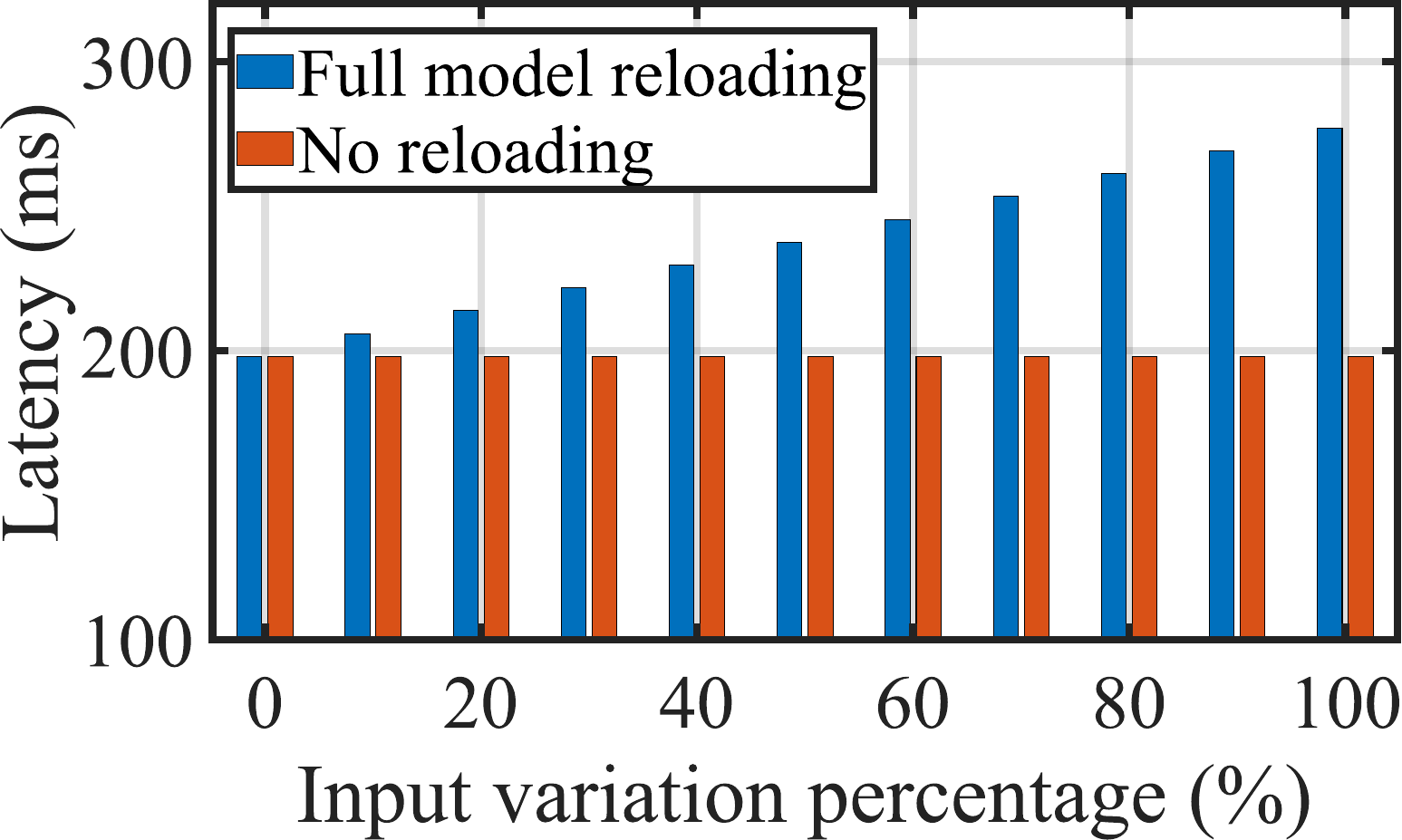}
            \label{subfig:latency}
        }
        \subfigure[Memory.]{
            \includegraphics[width = .5 \linewidth]{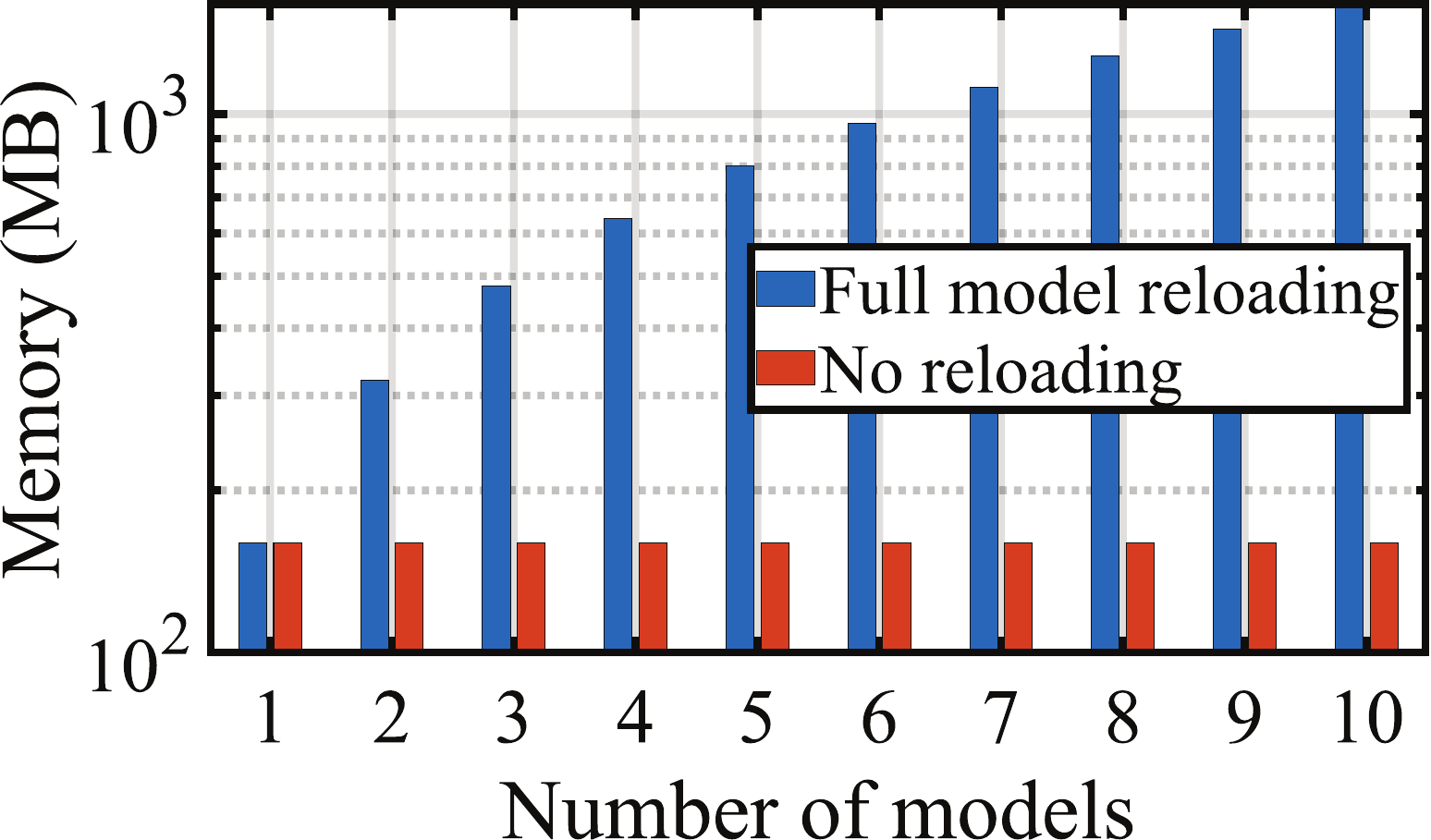}
            \label{subfig:memory}
        }
    }
    \caption{Latency and memory requirements caused by model reloading.}
    \label{fig:reloading}
    \vspace{-2ex}
\end{figure}

%%%%%%%%
%%%%%%%%note by yuhang: I understand ``sensor parameter changes" and ``missing modalities" are echoes to previous subsections, while the former term should be re-organized.
%%%%%%%%
To address the performance degradation caused by environment variance, sensor parameter changes, and missing modalities, a straightforward solution is to pre-train a set of distinct DNN models for different data input variations and missing modalities. These models can be reloaded as needed in corresponding scenarios.
However, changing environments can lead to time-varying parameter changes and missing modalities, requiring frequent model reloading to accommodate these variations. This places a heavy burden on AV embedded systems in terms of memory and latency, making this solution impractical. 
To illustrate the memory and latency demands imposed by frequent DNN model reloading, we present the memory and latency requirements with and without model reloading in Figure~\ref{fig:reloading} using aforementioned pre-trained models.

%
% \setcounter{figure}{4}
% \vspace{-1ex}

As Figure~\ref{subfig:latency} shows, model reloading from hard drives can increase the inference latency by up to 40\% compared to not reloading the model. 
However, as reported in ~\cite{Dixit2016AutonomousVD}, the median driving reaction time in urban street is $370$~\!ms, which makes hundreds of millisecond latency unacceptable indeed. 
In fact, if we can fit all pre-trained models into memory instead of reloading them from hard drives, the inference latency will be comparable to the case of not reloading. %, as the latency caused by memory indexing is negligible. 
However, as Figure~\ref{subfig:memory} demonstrates, each time the AV reloads the model, it requires an additional 160~\!MB of hard disk space (i.e., typical size of our trained model) to store the new model parameters compared to not reloading model.
More importantly, the memory capacity is still limited (e.g., Tesla Model 3 has 8 GB RAM), and the memory needs to run multiple applications simultaneously.
Furthermore, contemporary AV inference engines heavily rely on neural network processors, which use SRAM with capacities comparable to L3 cache at best. The prolonged latency and additional memory requirements are unacceptable for real-time, memory-constrained AV embedded systems. Thus, a more efficient solution for DNN model adaptation is needed to handle variations in input data and modalities. Although parameter-efficient fine-tuning methods~\cite{houlsby19a, hu2021lora, BenZaken2021BitFitSP} have succeeded in LLMs for adaptation, these strategies are not directly applicable to AV-DNN models. Adapting such fine-tuning approaches to address the specific data variation issues in AV-DNN models remains an unresolved challenge. AV-DNN models incorporate diverse modules, such as transformers, convolutional layers, and residual convolution blocks. The challenge lies in developing a versatile fine-tuning method that integrates seamlessly with these diverse structures. To this end, we will present \name in Section~\ref{sec:design} and tackle the challenge in \S~\ref{ssec:variation_aware}.

\subsection{Transformer for Autonomous Driving}
DNN in computer vision has long been dominated by CNN (convolutional neural networks), and these architectures are enhanced with greater scale, more extensive connection, and more sophisticated form of convolution. 
Recently, the Transformer architecture ~\cite{Vaswani2017AttentionIA} is adapted from NLP~(natural language processing) to vision community~\cite{Dosovitskiy2020AnII, Liu2021SwinTH}. 
Vision transformer provides the capability to encode distant dependencies or heterogeneous interactions, which is crucial for autonomous driving scenario, and is qualified to be a powerful backbone as achieves better performance with similar complexity against convolutional-based backbone counterparts.
The attention module, as a component of transformer, plays a critical role in modeling the interactive relation. Mathematically speaking, it is computed as:
\begin{align}\label{eq: classic_attn}
    \bm{A_{X}} = \bm{X}^{T}\bm{W_{Q}}^{T}\bm{W_{K}}\bm{X}/\sqrt{d}
\end{align}
Where $\bm{X} \in \mathbbm{R}^{f\times n}$ denotes $f$-dimensional $n$ features, usually as the intermediate results translated from sensor data with encoders.
$\bm{W_{Q}, W_{K}} \in \mathbbm{R}^{d\times f}$ are feature projection matrices which project vectors to $d$-dimensional ones. Many attention modules split relative large projected dim $d$ into pieces as known as ``multi-head attention" to obtain effective performance, which implies $d\ll n$.

Transformer-powered DNNs are notoriously difficult to train from scratch, particularly in the presence of noisy data, often due to ill-conditioned attention modules. \name effectively tunes these AV-DNNs by addressing and correcting issues within the attention modules.

% 3.5 - 4 pages 
\section{System Design}\label{sec:design} 

This section introduces the design of \name. First we give an overview, then we introduce the variation-aware model adaptation and cross-modal contrastive learning. Finally, we put everything together and summarize the training strategy.

\begin{figure*}[t]
    \setlength\abovecaptionskip{6pt}
	\centering
	\includegraphics[width=0.92\linewidth]{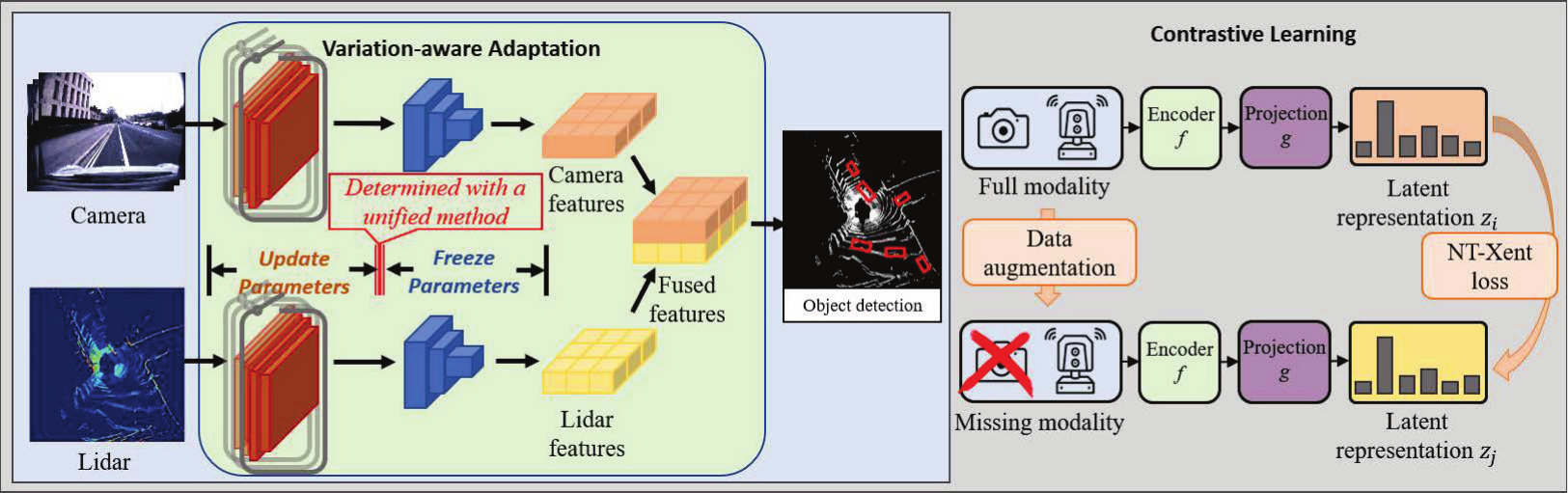}
	\caption{\name architecture: Optimized multimodal inference with contrastive data augmentation.}
	\label{fig:arch}
\end{figure*}

\subsection{Overview}
Motivated by the observation in \S~\ref{sec:background_motivation}, we design \name, a system consisting of two key components: i) a variation-aware model adaptation mechanism for efficient multimodal inference under memory and latency constraints, and ii) a cross-modal contrastive learning algorithm that addresses missing modalities and improves inference robustness. 
% In the following sections, we first provide an overview of \name, and then present the details of the two modules, before putting them together in the end.
% robust and efficient multimodal inference for autonomous driving. Finally, we show how we combine the modules to create \name.
%
% \begin{figure*}[h]
%     \setlength\abovecaptionskip{6pt}
% 	%\centering
% 	\includegraphics[width=0.92\linewidth]{tmp_imgs/arch.pdf}
% 	\caption{\name architecture.}
% 	\vspace{-1ex}
% 	\label{fig:arch}
% \end{figure*}
%
% \subsection{System Overview}
%
% Our goal is to design a multimodal DNN for autonomous driving that can handle variations of sensor data and  modalities, while achieving high efficiency and robustness. 
% One potential solution to overcome this challenge is to prepare multiple models for each case and switch among them. However, this would significantly impair the inference efficiency of the DNN model. 
%While preparing multiple models for each case and switching among them could in theory overcome input variations, it would significantly impair the inference efficiency of the DNN model. Therefore, 
% we prioritize \textit{inference efficiency} in terms of memory, latency, and energy consumption. 
%
% To achieve both robust and efficient inference, we propose \name, 
As shown in Figure~\ref{fig:arch}, \name maintains compatibility with existing  multimodal DNNs for AVs, while incorporating two aforementioned components that will be elaborated in \S~\ref{ssec:variation_aware} and \S\ref{ssec:contrastive_learning}, respectively. 
Given a multimodal DNN (e.g., BEVFusion) and sensory data~(e.g., lidar and camera), the first component % variation-aware model adaptation mechanism 
%%%%%%%%
%%%%%%%%note by yuhang: how do we re-organize the function of first component?
%%%%%%%%
identifies, selects and injects the variation-aware model parameters. The second component compensates for information loss due to missing modalities.
% leverages an optimization algorithm to identify and select the variation-sensitive model parameters.
% avoiding excessive reloading overhead. For the case of missing modalities, 

The design of \name resolves three key technical challenges. 
(i) Adapting techniques from LLM to AV-DNN to manage data variations (\S\ref{ssec:variation_aware}), as introduced in \S~\ref{ssec:motivation_reloading}.
(ii) Narrowing the performance gap between full and missing modalities (\S\ref{ssec:contrastive_learning}), as discussed in \S~\ref{ssec:motivation_missing_modality}.
(iii) As most training samples are collected in a normal environment, we may only have limited types of data variations for fine-tuning the pre-trained model. However, the AV-DNN models always encounter unseen variations in reality, which may result in unexpected performance drop. Therefore, our final challenge is to expand the capability of fine-tuned models to deal with unseen variations (\S\ref{ssec:together}).

\vspace{-2ex}
\subsection{Variation-Aware Model Adaptation} \label{ssec:variation_aware} 
%%%%%%%%
%%%%%%%%note by yuhang: Three major challenges concluded by Prof Li:
    % \begin{itemize}
    %     \item where (which layers) to tune (We previous start from tuning the dense layers (i.e., FC), which didn't work)?
    %     \item how to tune then?
    %     \item how to jump out of the scope of ``transformer" (to be more specify, attention mechanism) and make it more general? 
    % \end{itemize}
%%%%%%%%
% We introduce a variation-aware model adaptation mechanism to tackle sensory data variations.
%
% Training a large DNN is computationally expensive and relies on a huge amount of data, especially for transformer-based models~\cite{Liu2021EfficientTO}.
%%%%%%%%
%%%%%%%%note by yuhang: A not very fair comparison provided by the above literature. (A): ViT trained with ``303million" high resolution img. (B): ResNet with similar complexity, trained with ``1.3million" ImageNet samples. (A) yields worse performance than (B) in ``classification" task.
%%%%%%%%
%%%%%%%%
%%%%%%%%note by yuhang: warm-up plays a critical role in train a transformer successfully, see the ``Introduction" in the above literature.
%%%%%%%% 
% Parameter-efficient fine-tuning has shown promising performance on LLMs, and various techniques~\cite{houlsby19a, hu2021lora, BenZaken2021BitFitSP} have been introduced, which could be utilized to fine-tune the pre-trained model for addressing the data variations in our concerned scenarios.

Existing parameter-efficient fine-tuning techniques~\cite{houlsby19a, hu2021lora,Kong2022m3TrackMM,BenZaken2021BitFitSP} are mostly designed for downstream tasks of LLMs, and how to effectively apply those approaches to AV-DNN models is barely studied. To investigate the effectiveness of the parameter-efficient fine-tuning method for AV-DNN models, we start with the models with the same transformer modules as LLMs, and then extend to other widely-used modules in AV-DNN models.
Drawing inspirations from BitFit~\cite{BenZaken2021BitFitSP}, which focuses on fine-tuning lightweight inductive-bias terms only, we begin by tuning all the normalization layers and task-specific heads in the AV-DNN model. This foundational operation of \name is elaborated in \S~\ref{ssec: ablation}, and results demonstrate that it significantly outperforms the conventional practice of only fine-tuning task-specific heads.
% As \S\ref{ssec:motivation_missing_modality} shows, training a model from scratch is hyperparameter-sensitive and has a high likelihood of corruption in the early stage.  Our insight is that the vanilla tuning (i.e., update every parameters) from a well-pretrained model is entirely hopeless towards efficiency.
% Inspired by the success of \textit{adapter}~\cite{houlsby19a} in Natural Language Processing (NLP), we start with partially fine-tuning the model. Specifically, we tune \textbf{all} the layers in prediction heads (i.e., parallel modules to regression detection information, such as scale and translation) and \textbf{only} the normalization layers in other upstream modules.

% It is intuitive: a well-pretrained model implies it can extract features of fed data clearly given that they are informative, which means its heavyweight convolution-based \textit{extractor}, which is fundamental for AV DNNs, is effective. 
% Tuning these \textit{extractor} with variation data incurs mountain in labour, in contrast, normalization layers are lightweight and widespread.

\begin{figure}[t]
    \centerline{
        \setcounter{subfigure}{0}
        \subfigure[Eigenvalues distribution.]{
            \includegraphics[width = .47 \linewidth]{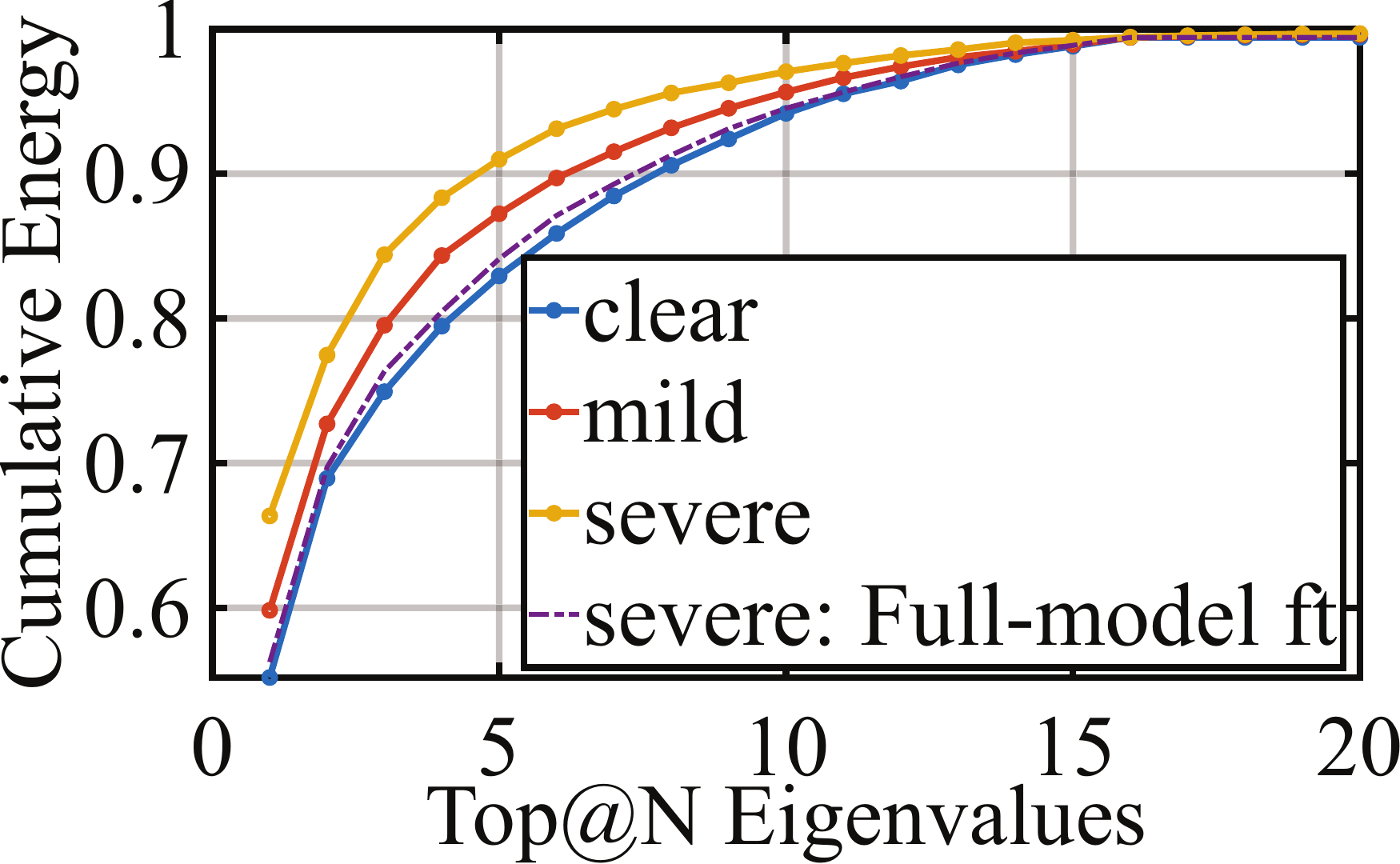}
            \label{sfig: low_rank_demo}
        }
        \subfigure[Low-rank structure.]{
            \includegraphics[width = .53 \linewidth]{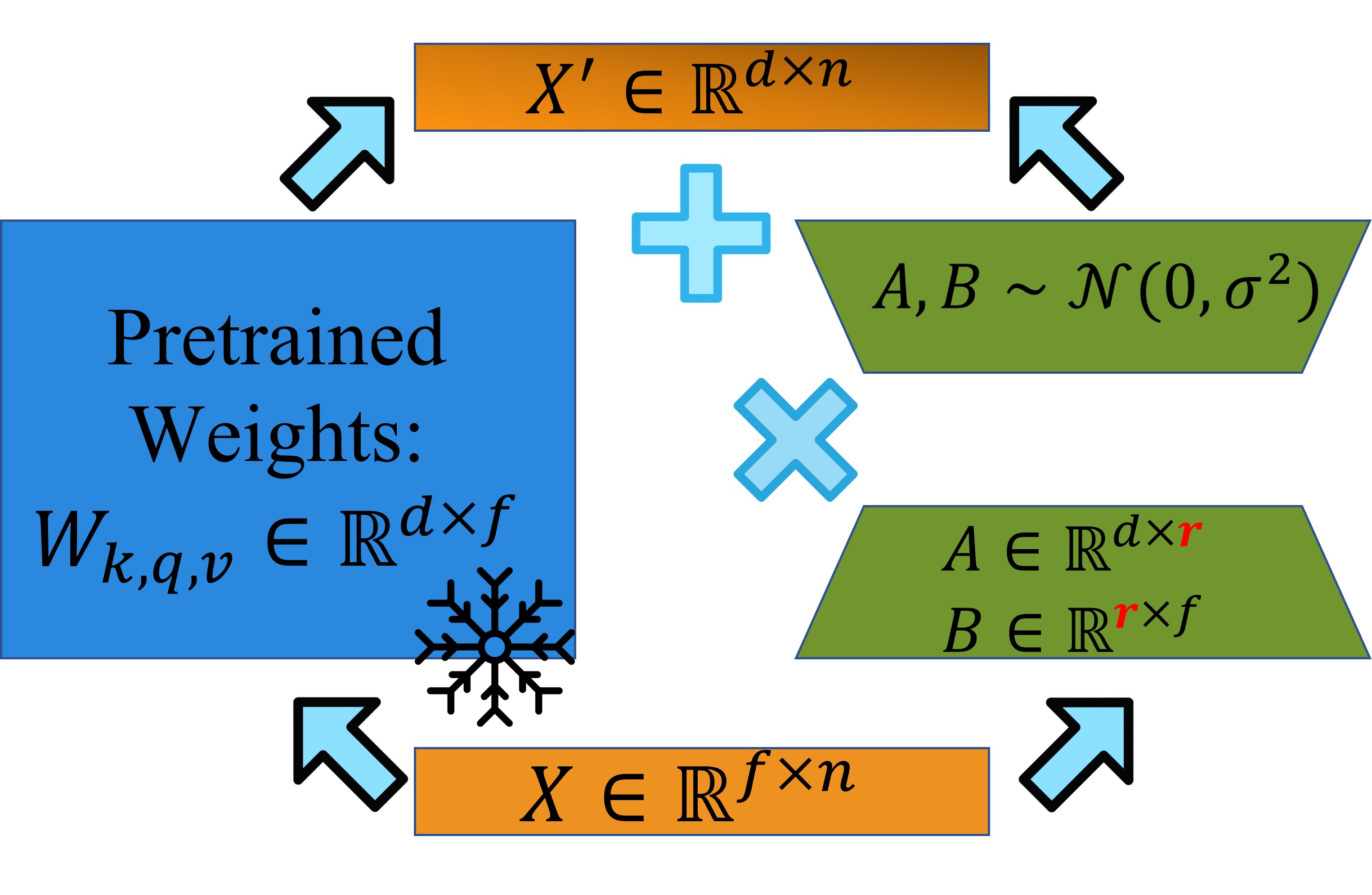}
            \label{sfig: lora_module}
        }
    }
    \vspace{-2.5ex}
    \caption{~\subref{sfig: low_rank_demo} Even severer low-rank problem. ~\subref{sfig: lora_module} Proposed LoRa module to alleviate the adverse effect.}
    % \caption{A: Eigenvalues of distribution a representative  attention module. The rank will even lower when input data is under more severe distortion. B: LoRa structure to alleviate low-rank effect.}
    %%%%%%%%
    %%%%%%%%
    \label{fig: low_rank}
    \vspace{-2.5ex}
\end{figure}

\paragraph{Low-Rank Adaptation} 
Our exploration continues into the transformer module, where the attention mechanism is a critical component. 
% To go a step further and focus on the transformer attribute of these DNN, we start from a observation: The ``attention" mechanism, as a major merit rendered by attention module, seems to fail when input data is under distortion. 
We conduct an experiment by setting $n=200, d=16$ in Equation~\ref{eq: classic_attn}, calculating the cumulative sum of eigenvalues of $\bm{A}{x}$ for each sample, and averaging them to gauge its rank. 
As Figure~\ref{sfig: low_rank_demo} shows, it is noticeable that even on clean data that the attention module is most familiar with, it already exhibits a low-rank instance as the top $3\%$ eigenvalues account for more than $85\%$ energy. This bias amplifies as sensor data distortion intensifies. In other words, the matrix $\bm{A}{x}$, inherently low-rank due to multi-head operations, degrades when processing distorted data. However, this is not a consequence of low-quality input data since comprehensive fine-tuning can substantially mitigate this effect.
The observed rank collapse phenomenon was first recognized in NLP, specifically when fine-tuning the pre-trained model for distinct tasks. To overcome this, a Low-Rank Adaptation method~\cite{hu2021lora} is proposed by of injecting rank-decomposition matrices to transformer blocks. Drawing on this concept, we design low-rank modules to fine-tune transformer blocks in AV-DNNs based on sensory data variations. Figure\ref{sfig: lora_module} presents its detailed architecture.
%
%We take this idea to design so called Low-rank modules, with the intention of adapting transformer modules in AV DNN efficiently. Figure~\ref{sfig: lora_module} depicts the simple architecture of it.
%
In general, the low-rank module comprises a pair of matrices $A, B$, which coexist parasitically for each inherent projection weight matrix. Each matrix has a low rank, bounded by a hyperparameter $r$, and is transparent due to their near-identity initialization. During fine-tuning, we keep the bulky pre-trained weights frozen, allowing only updates to these low-rank modules. As our objective is to restore the pre-trained matrix, and its rank is relatively low in optimal scenarios (as indicated in Figure~\ref{sfig: low_rank_demo}), we can efficiently bound $r$ by a small integer. Additionally, the two distinct paths that are demonstrated in Figure~\ref{sfig: lora_module} can operate in parallel, causing no distinctive latency overhead.

\paragraph{Generalization to Non-Tranformer Modules} It is 
worth noting that not all AV-DNNs rely on transformers. To make our adaptation approach general and not specific to particular models, we take a closer look at the similarities and distinctions between transformer-heavy DNNs and others. 
Our key insight is that transformers or residual convolution blocks widely adopted cutting-edge AV-DNNs are deeply interconnected. 
For example, they both utilize the skip connection, significantly mitigates the rate of rank degeneration~\cite{Dong2021AttentionIN}. 
Furthermore, it has been demonstrated that a convolution-only network with a relatively large kernel size can achieve performance on par with its transformer-based counterparts~\cite{Liu2022ACF}, suggesting that the attention mechanism of transformers can be emulated with larger convolution kernels.
Moreover, we find parallels in viewing the attention module as an ensemble of shallow networks, as studies of ResNet point out~\cite{Veit2016ResidualNA}.
%
% Our insight is, most advanced AV DNN is either transformer intensive or residual convolution blocks intensive, and they have profound relationships. For example, the skip connection technique which transformer adopts from the other, lags the degeneration of rank greatly~\cite{Dong2021AttentionIN}. Beside, attention module can be viewed as ensembles of shallow networks, while similar phenomenon for ResNet is studied even before~\cite{Veit2016ResidualNA}.
%

\begin{figure}[t]
    \centerline{
        \includegraphics[width = .6 \linewidth]{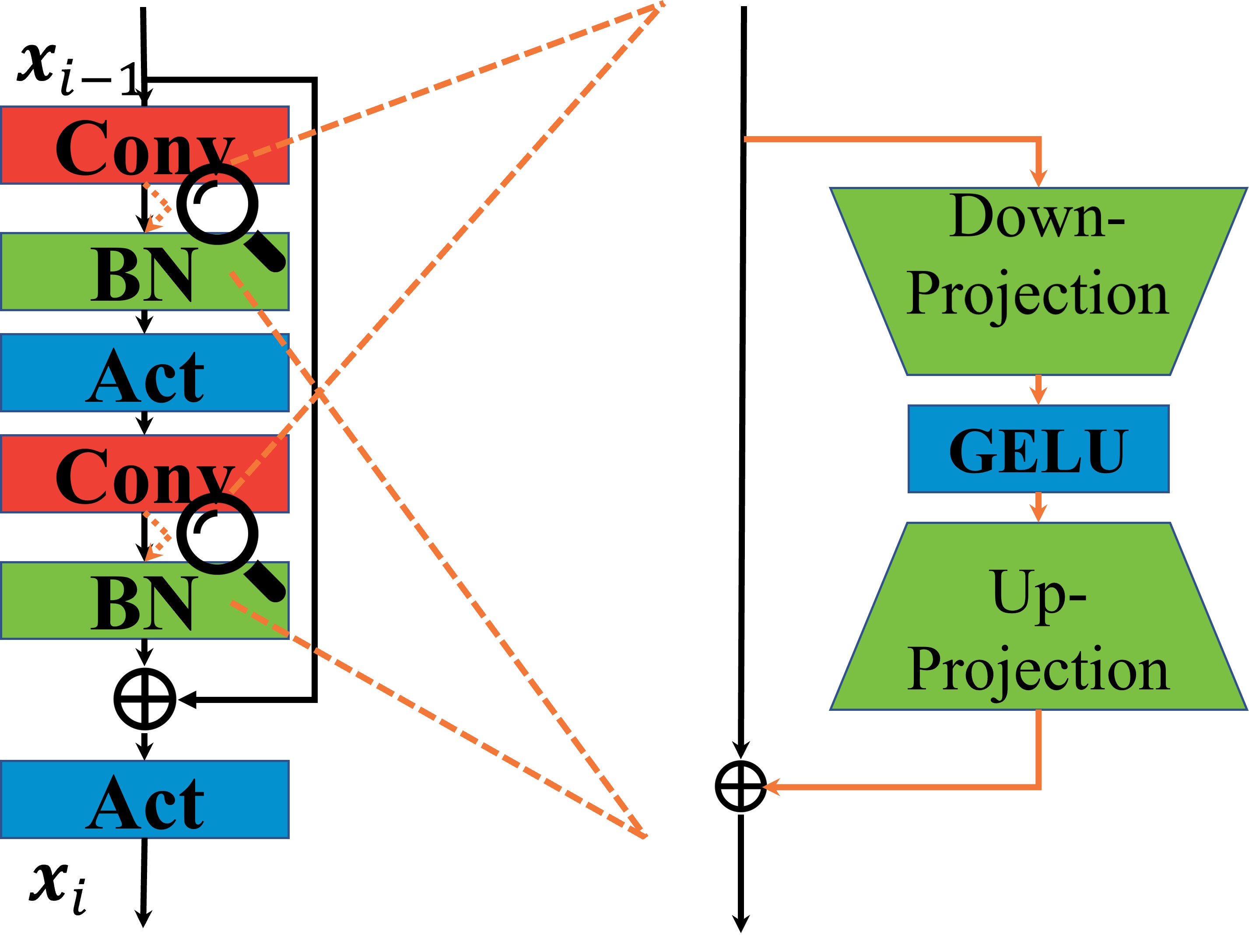}
    }
    \vspace{-2.5ex}
    \caption{A variant of widely-adopted residual convolution block, we inject such transparent module.}
    \label{fig: basicblock_adapter}
    \vspace{-2.5ex}
\end{figure}

Based on these insights, with the intention of reviving convolution kernels, we inject a module with a similar structure as the low-rank module into residual blocks, resulting in a different style of adaptation in the attention module~\cite{houlsby19a}. 
This module, demonstrated in Figure~\ref{fig: basicblock_adapter}, is placed before batch normalization layers with skip connection, a decision informed by a similar layer arrangement in transformers.
We complement this setup with an activation layer, given the absence of an apparent low-rank attention matrix. By harmonizing the above three techniques, we ensure our adaptation method remains efficient and universally applicable.

\subsection{Cross-Modal Contrastive Learning}\label{ssec:contrastive_learning}

% step 1, idea of contrastive learning 的结构
% step 2, loss function

%A powerful approach for training models on a large amount of data without requiring a large amount of labels is semi-supervised learning (SSL). SSL mitigates the requirement for labeled data by providing a means of leveraging unlabeled data. Since unlabeled data can often be obtained with minimal human labor, any performance boost conferred by SSL often comes with low cost. This has led to a plethora of SSL methods that are designed for deep networks
Recall that the preliminary experiments in \S\ref{ssec:motivation_missing_modality} showed that the conventional data imputation methods, such as filling the missing modalities with zeros, incur information loss and even introduce biases into the DNN model. 
To compensate for the information loss caused by missing modality, we resort to representation learned from full modalities (e.g., camera and lidar) to guide uni-modal data (e.g., camera) towards a unified multimodal representation space. 
To this end, we design a cross-modal contrastive learning framework. The key idea behind the this framework is that even when some modalities are missing due to sensor occlusion or malfunction, the latent representation learned from missing modalities should be as similar as possible to the representation extracted from complete modalities.  
%there may be some missing modalities due to sensor occlusion and malfunction, the extracted latent representation with missing modalities should be as close to the one with full modalities as possible. 
For example, if only one camera on an AV, which are equipped with multiple cameras and lidars, is missing, the inference performance can be preserved if the missing camera's latent representation can be compensated by the remaining sensors, which typically have overlapping fields of view (FoV).
%For example, if only one camera on the AV (equipped with multiple cameras and lidars) is missing, the inference performance could remain largely the same if the camera's missing latent representation can be compensated by the remaining sensors. 

To make the latent representations with and without missing modalities as similar as possible, follow work~\cite{zheng2023AutoFed,Kong2021ContinuousAT}, \name employs four components (i.e., data augmentation, feature extractor, projection head, and contrastive loss) in its contrastive learning framework, as shown in the right side of Figure~\ref{fig:arch}. 
First, \name uses a stochastic data augmentation module to remove each sample of some modalities with a probability of 10\%,  resulting in one missing-modality dataset paired with the original full-modality one. The full-modality dataset and the one with missing modality are  denoted as $x$ and $\alpha(x)$, respectively. We consider $(x_i,a(x_i))\in\{(x_i,a(x_i)): i\in|x|\}$ as a positive pair, and $(x_i,a(x_j))\in\{(x_i,a(x_j)): i, j\in|x|, and\ i\neq j\}$ as a negative pair since $x_i$ and $a(x_j)$ describe different scenes.
Then, \name leverages a neural network based encoder $f(\cdot)$ to extract representation vectors from a pair of samples from $x$ and $\alpha(x)$. 
Our framework allows various choices of network architecture without any constraints, and we opt for simplicity, and reuse the feature extractor used in the object detection network. 
Thereafter, we use a multilayer perceptron with one hidden layer as $g(\cdot)$, which is used to map representations to the latent space where the contrastive loss is applied.
Last but not least, we use a contrastive loss function, termed NT-Xent~\cite{chen2020simple}, i.e., the normalized temperature-scaled cross-entropy loss, to train the encoder $f(\cdot)$. The loss function enforces that the features between a positive pair as similar as possible, while enlarging the distance between the features of a negative pair. 

We randomly sample a mini-batch of $N$ samples from $x$ and one sample from $a(x)$, i.e., $a(x_i)$ and $x_i\in \{x_1,\dots,x_N\}$ resulting in $N+1$ samples. Given a positive pair $(x_i,a(x_i))$, we consider the other $N-1$ samples from $x$ as negative samples with regard to $a(x_i)$. Then the contrastive loss for a positive pair $(x_i, a(x_i))$ can be formally expressed as:
\begin{align} \label{eq:ntxent}
\mathcal{L}_{i}=-\log \frac{\exp \left(\mathbf{sim}\left(\boldsymbol{z}_i, \hat{\boldsymbol{z}_i}\right) / \tau\right)}{\sum_{k=1}^{N} \mathbbm{1}_{[k \neq i]} \exp \left(\mathbf{sim}\left(\boldsymbol{z}_k, \hat{\boldsymbol{z}_i}\right) / \tau\right)}
\end{align}
where $z_i = g(f(x_i))$ is the projected feature for scene $x_i$ and $\hat{z_j} = g(f(\alpha(x_j)))$ is the projected feature for scene $x_j$ where some missing modality happens, $\mathbf{sim(\cdot)}$ is a function that calculates cosine similarity of two latent representations,  $\mathbbm{1}_{[k \neq i]} \in\{0,1\}$ is an indicator function evaluating to 1 if $k \neq i$, and $\tau$ denotes a temperature hyper-parameter, whose appropriate tuning can help the model learn from hard negatives as it controls the penalties on hard negative sample. The final loss is computed across all positive pairs in a $N$ size mini-batch. 
Minimizing the contrastive fusion loss will force the projected features from the same scene but modality-missing conditions are different (i.e., $(\boldsymbol{z}_i, \hat{\boldsymbol{z}_i})$ pair) together, while pushing projected features from different scenes (i.e., $(\boldsymbol{z}_k, \hat{\boldsymbol{z}_i}),k\neq i$ pairs) apart. A previously trained but not robust model is incorporated into the contrastive learning framework, during the subsequent adaptation phase, these additional components are removed. The contrastive learning framework approximately doubles the forward propagation time, however, we find that a relatively small number of epochs is sufficient, and it remains completely transparent during inference. Therefore, we consider it introduces an affordable computational cost.

%where $\theta$ is the encoder, $\alpha$ is a stochastic function that randomly replaces certain modalities with 0's with a probability of 10\%.
%%%%%%%%
%%%%%%%%note by yuhang: should we re-write the algorithm? And I find the next section hard to re-write.
%%%%%%%%
\subsection{Putting Everything Together}\label{ssec:together}
% We carefully summarize the training strategy of the \name, and describe the overall workflow in \needrev{\textbf{Algorithm~\ref{alg:fed}}}. The algorithm takes the pre-trained DNN model $\mathcal{D}$ and memory constraint $\epsilon$ as inputs, and outputs the fine-tuned subset $\mathcal{A}_q$ of the DNN model. By loading the layers in runtime, the DNN models can achieve robustness and efficiency by adapting to sensor parameter change and missing modalities. In the algorithm, we fine-tune the DNN model for each possible parameter change. The function $\FuncSty{coarse-to-fine}\left(\boldsymbol{\delta}^q, \boldsymbol{n}, \epsilon \right)$ is implemented based on \S\ref{ssec:variation_aware} to provide the list of selected layers, $\hat{\mathcal{D}}$ for the fine-tuned model. According to the $\hat{\mathcal{D}}$, we fine-tune the DNN model again, but keeping the rest of model intact. With multiple fine-tuned results $\mathcal{A}_{q,p}$  of the selected layers and a single copy of the base loaded in memory, \name switches between $\mathcal{A}_{q,p}$, following a current input parameter change (e.g., exposure and motion blur), while reusing the base across all possible inputs.

%%%%%%%%
%%%%%%%%note by yuhang: overwrite the workflow.
%%%%%%%%
We summarize the training strategy of the \name, and present the overall workflow as follows.
With the strategy presented in \S\ref{ssec:contrastive_learning}, \name re-trains the model under missing modality settings, allowing each sensor to fail independently. During the following tuning phase, which addresses various scenarios with variation, \name is constrained by two parameters, namely rank bound $k$ and projection squeeze ratio $r$, as we have discussed their insights in \S\ref{ssec:variation_aware} above.

In reality, a cocktail of domain-specific variations muddles the situation, rendering the tuning of the entire domain space anything from unmanageable to impossible. Yet, by employing tuned parameters from individual variations, we squeeze the combined domain's tuning scope from $M \times N$ to a more manageable $M + N$ (respecting lidar and camera modalities). Furthermore, we find unseen domains like nighttime often share traits with seen domains, such as underexposure. We propose a conditional melding of two tuned variants $\mathcal{L}{c}^{'}$ and $\mathcal{L}{l}^{'}$: for exclusive layers $(\cup_{i \in \mathbf{N}}\mathcal{L}{c{i}}^{'}) \oplus (\cup_{j \in \mathbf{N}}\mathcal{L}{l{j}}^{'})$, where $\oplus$ indicates a symmetric difference operation, we opt for a straightforward update. More complicated are the overlapping layers $(\cup_{i \in \mathbf{N}}\mathcal{L}{c{i}}^{'}) \bigcap (\cup_{j \in \mathbf{N}}\mathcal{L}{l{j}}^{'})$, for which we choose interpolation using $\lambda_{c} * \mathcal{P}{c}\left(t\right) + \lambda_{l} * \mathcal{P}{l}\left(t\right)$, where $t$ denotes overlapping layers and $\mathcal{P}{{c, l}}\left(t\right)$ represents the model-specific parameter set for $t$, with the additional constraint $\lambda_{c} + \lambda_{l}=1$.

%While in realistic scenarios, a bunch of domain-specific variations overlap together, and tune the whole domain space is unscalable at best and infeasible at worst. To this end, we leverage tuned elements from individual variations, the combined domain's tuning scope can be narrowed from $M \times N$ to $M + N$ (lidar and camera modalities respectively). Additionally, unseen domains (e.g., nighttime) may share similarities with seen domains (e.g., underexposure). We consider a conditional combination of of two tuned variants $\mathcal{L}_{c}^{'}$ and $\mathcal{L}_{l}^{'}$: For the exclusive layers $(\cup_{i \in \mathbf{N}}\mathcal{L}_{c_{i}}^{'}) \oplus (\cup_{j \in \mathbf{N}}\mathcal{L}_{l_{j}}^{'})$ (where $\oplus$ operator indicates symmetric difference operation), updating them is straightforward. For the more complicated overlapping layers $(\cup_{i \in \mathbf{N}}\mathcal{L}_{c_{i}}^{'}) \bigcap (\cup_{j \in \mathbf{N}}\mathcal{L}_{l_{j}}^{'})$, we employ interpolation using $\lambda_{c} * \mathcal{P}_{c}\left(t\right) + \lambda_{l} * \mathcal{P}_{l}\left(t\right)$, where $t$ belongs to the overlapping layers and $\mathcal{P}_{\{c, l\}}\left(t\right)$ represents the model-specific parameter set for $t$, we apply the additional constraint $\lambda_{c} + \lambda_{l}=1$.

While it's possible to overwrite the tuned layers, \name takes a different approach for the sake of efficiency. It simultaneously loads all the layers that are  tuned under different variations, and further refines the less significant bits within the "shared" layers through pruning. These layers are organized as the \textit{values} in a \textit{map}. During inference, \name utilizes the information provided by various sensors (e.g., brightness) to encode the \textit{key} that switches to the desired layer set. While a more self-contained method that only uses input data to switch parameters might be ideal, it can be more specific to certain modalities and less general. For example, it might be easier to implement for cameras but more challenging for lidar. Simple solutions like a lightweight filter are capable of rating images, mapping the ratings to keys, and selecting the tuned parameters to switch. However, it is not as straightforward when dealing with lidar data. The complexity of lidar data, with its 3D point cloud representation, makes it challenging to use the same methodology to rate, map, and switch parameters as easily as we do with images. Therefore, we consider the development of a more self-contained approach to be part of our future work. This would enhance \name's adaptability to different variations in sensory data and modalities.

\section{Implementation and Experiment Setup} \label{sec:implementation}

In this section, we first present the details of \name's implementation, then we apply \name to develop two widely-used applications. Finally, we describe the metrics that we use to comprehensively evaluate the performance of \name.

%\subsection{Tasks}
     % For evaluation tasks, at this time we consider \textbf{3D Object Detection} and \textbf{Semantic Segmentation}. To notice it, the tasks that a dataset orients for is determined by the annotation structure to a great degree, which means \textbf{Tracking} tasks ~\cite{Pang2021SimpleTrackUA, Yin2021Centerbased3O} can be easily attended to these time serialized scenes where dozens of frames are snapshotted in a row, while is unlikely to work for \textbf{Place Recognition} task due to the low-frequency re-visit scheme~\cite{Kim2020MulRanMR}.
\subsection{Implementation}
We implement the vehicle detection application on a server equipped with an Intel Xeon Gold 6226 CPU~\cite{intel}, 128~\!GB RAM, and NVIDIA GeForce RTX 3080 Ti GPU~\cite{nvidia}.  As for the software, Python 3.7 and PyTorch 1.9.1~\cite{pytorch} are used for implementing the application. Our object detection and segmentation model is built upon mmDetection~\cite{chen2019mmdetection}, which is an open-source toolbox that provides state-of-the-art OD models. In particular, the model components and settings for \name are as follows:
%%%%%%%%
%%%%%%%%note by yuhang: I re-write some items which is not precise enough. And comment some items which is too detail (e.g., detailed configuration in OD/Seg tasks).
%%%%%%%%
%
\begin{itemize}
        \item The encoder $f(\cdot)$ consists of two modality-specific encoders. For the camera and lidar modalities, Swin Transformer-T~\cite{Liu2021SwinTH} and VoxelNet~\cite{Zhou2018VoxelNetEL} are used as the encoders, which is transformer intensive and residual block intensive respectively. 
        % For camera encoder, there are $12$ transformer blocks totally scatter into $4$ stages, where each stage downsamples the feature map with a ratio of $2$ and doubles the feature dimensions. For lidar encoder, efficient sparse 3D convolution is performed at a cost of memory. every last convolution layer in each stage downsamples the feature map by half via convolution with a stride of $2$, after each convolution layer, batch normalization and ReLU operations are applied. 
        %%%%%%%%
        %%%%%%%%note by yuhang: in fact, most modern camera encoders don't use pure ResNet anymore. In contrast, transformer backbone model like Swin Transformer has qualified this capability of being a  backbone of general usage, like these models we based on. However, it is very heavy weight.
        %%%%%%%%
    
    % \item The encoder $f(\cdot)$ consists of two modality-specific encoders. For the camera and lidar modalities, ResNet~\cite{he2016deep} \noteyuhang{infect, most modern camera encoders don't use pure ResNet anymore. In contrast, transformer backbone model like Swin Transformer has qualified this capability of being a  backbone of general usage, like these models we based on. However, it is very heavy weight.} and VoxelNet~\cite{Zhou2018VoxelNetEL} are used as the encoders, respectively. The encoders have three blocks of convolutional layers. The first layer of each block downsamples the feature map by half via convolution with a stride of 2, followed by a sequence of convolutions of stride 1. After each convolution layer, batch normalization and ReLU operations are applied.

    \item  We choose to project both camera's and lidar's data to a unified bird's-eye view. For lidar, we flatten the sparse lidar features along the height dimension, hence not creating geometric distortion. For camera, we cast each camera feature pixel back into a ray in the 3D space, which can result in a feature map that retains full semantic information from the cameras.

    \item Even though all sensory inputs are converted to a unified representation, the lidar features and camera features can still be spatially misaligned to some extent due to the inaccurate depth in the view transformer. To this end, we apply a fully convolutional encoder (with a few residual blocks) to compensate for such local misalignments.

    % \yuhang{
    %     \item We consider two major tasks in AV, namely the objection and semantic segmentation task. For the object detection task, the number of proposals during each inference is 200, the NMS (Non-Maximum Suppression) kernel size is set to 3, the dropout rate is set to 0.1, ReLU is used as the activation function, Gaussian focal loss~\cite{lin2017focal} is used as the classification loss, L1 loss is used for bounding box regression. For the segmentation task, we interpolate the map to $200\times200$ grids with bilinear since the resolution is task-specific, and simply perform Gaussian focal loss after \textit{grid-wise} sigmoid operation.
    % }

    % \yuhang{
    %     \item We consider For the object detection task, the number of proposals during each inference is 200, the NMS (Non-Maximum Suppression) kernel size is set to 3, the dropout rate is set to 0.1, ReLU is used as the activation function, Gaussian focal loss~\cite{lin2017focal} is used as the classification loss, L1 loss is used for bounding box regression.
    % }

    %\item For the semantic segmentation task, 
    %The angles of the rotated anchors used by the RPN are set to -90$^{\circ}$, -45$^{\circ}$, 0$^{\circ}$, and 45$^{\circ}$.
    %
    %\item Both lidar and radar feature extractors are composed of four consecutive convolutional layers with a kernel size of 3 and padding of 1.
    %
    %\item The IoU threshold \newrev{(defined later)} of NMS for removing excessive proposals during testing is set to 0.2.

    \item In the tuning process, the AdamW optimizer, which employs the decoupled weight decay regularization~\cite{loshchilov2017decoupled} is used by setting a fixed learning rate of $5 \times 10^{-5}$ and weight decay as 0.01. We also enabled gradient clip, which confines the L2 norm to be less than 35.

    % \item In the local training process, the AdamW optimizer, which employs the decoupled weight decay regularization algorithm~\cite{loshchilov2017decoupled} is used by setting a cyclic learning rate (with an initial learning rate of 0.0001) and weight decay as 0.01. We also enabled gradient clip, which confines the L2 norm to be less than 35. 
%
\end{itemize}

\subsection{Tasks and Dataset}
%%%%%%%%
%%%%%%%%note by yuhang: shall we reserve the DENSE dataset? Since we don't conduct OD or Seg task in current version. We gain the insight of generating physical rational variation partly from DENSE.
%%%%%%%%
We perform two tasks to evaluate of \name, i.e., \textit{object detection} and \textit{semantic segmentation}, which are primary tasks for autonomous driving perception. 
We evaluate our scheme on two large-scale public datasets for autonomous driving: nuScenes ~\cite{Caesar2020nuScenesAM} and DENSE~\cite{Bijelic2020SeeingTF}. Details of the two datasets are given in Table~\ref{tbl:dataset}.
% Object detection takes sensory inputs and produces bounding boxes with class labels around different objects including car, pedestrian, etc. Semantic segmentation presents the exact outline of distinct classes in the input through a pixel-wise annotation.
% car, truck, bus, trailer, construction vehicle, pedestrian, motorcycle, bicycle, barrier, and traffic cone. 

% Semantic segmentation is an extension of object detection \yuhang{(while corresponding classes is much more limited)} which presents the exact outline of an object in the input through a pixel-wise annotation.

% \needrev{
% % PS: DENSE dataset is a future work, which I have to do tons of preparation work to re-organize it. comment it if necessary.
% % PS: related information of nusCenes lies in page3-page6 in that paper.
% % PS: related information of DENSE lies in page4-page5 in that paper.
% We evaluate our method on nuScenes ~\cite{Caesar2020nuScenesAM} and DENSE~\cite{Bijelic2020SeeingTF}\footnote{to be continue in late Dec and Jan.} large-scale outdoor datasets. For nuScenes, it has well-organized structure and rich annotation which support diverse kinds of tasks. The total 40160 annotated ``samples'' belong to 1000 ``scenes'' disjointly, and split into train/val/test set with a proportion of $\text{7:1.5:1.5}$. For DENSE, The total 13000 samples, where over $1/4$ is collected under challenging weather conditions (\textit{e.g.}, dense fog, snow, heavy rain.), which is also featured with the \textit{diversity} rather than the \textit{amount} of sensors.
% }
\begin{itemize}
    \item nuScenes includes 1000 driving scenes under different weather and illumination conditions in Boston and Singapore, which are known for dense traffic and challenging driving situations. There are approximately 1.4M camera images and 390k lidar frames which are annotated with 1.4M accurate bounding boxes for 23 classes.
    \item DENSE is captured during two test drives in February and December 2019 for two weeks, each under different weather (i.e., rain, snow, light/dense fog). There are approximately 104K camera images and 104K lidar frame which are annotated with 13.5K accurate bounding boxes.
    % for \needrev{23} classes.
\end{itemize}

\begin{table}[t]
\caption{Datasets used for \name}
\vspace{-1.5ex}
\renewcommand{\arraystretch}{0.8}
\small
\centering
\begin{tabular}{lll}
\hline
\textbf{Dataset}            & \textbf{NuScenes}\cite{Caesar2020nuScenesAM} & \textbf{DENSE}\cite{Bijelic2020SeeingTF}     \\ \hline
\textbf{Sensor Setup}       &                   &                    \\ \hline
Cameras                     & 6                 & 2                  \\
Camera Resolution           & 1600 $\times$ 900 & 1920 $\times$ 1024 \\
lidars                      & 1                 & 2                  \\
lidar Resolution            & 32                & 32, 64                 \\ \hline
\textbf{Dataset Statistics} &                   &                    \\ \hline
Images                      & 1.4M               & 104K    \\ %note: 13K*4 for stereo camera, 13K*3 for gated camera, 13K for far infrared (FIR) camera (I don't count rectified images into it.).
Bounding boxes              & 1.4M               & 13.5K              \\
Frames                 & 390K               & 104K     \\ %note: 13K*6 for 64-channels lidar, 13K*2 for 32-channels lidar.
Point density per frame & 35K           &30K, 55K               \\ 
% Annotation                      &               & 100K               \\
Annotations        & 1.17M             & 212K                \\ \hline %note: nuscene: cat | grep | wc from "train+val" set. In "test" set, there is no "annotation" but bounded box ground truth. dense: in ".txt", each annotation form a line. 111K for stereo, 101K for gated.
% Night                       & \cmark                 & \cmark                  \\
% Rain                        & \cmark                 & \cmark                  \\
% Snow/Fog                    & \xmark                 & \cmark                  \\ \hline
\end{tabular}
\vspace{-2.5ex}
\label{tbl:dataset}
\end{table}

%
%The angles of the rotated anchors used by the RPN are set to -90$^{\circ}$, -45$^{\circ}$, 0$^{\circ}$, and 45$^{\circ}$. The aspect ratio of the anchors is set to 2.5 to conform to the length-width ratio of regular vehicles. The thresholds of IoU for detecting vehicles and backgrounds are set to 0.55 and 0.45, respectively. Before the last step of classification and regression, the size of RoI pooling is set to $7 \times 7$. The IoU threshold of NMS during testing is set to 0.2. 

\subsection{Evaluation Metrics}
    \label{sec:metric}
To evaluate the performance of object detection, follow the previous works, we consider the widely used metric mean Average Precision (mAP), along with the specialized nuScenes detection score (NDS) tailored for the nuScenes. 
Since semantic segmentation can be considered as a \textit{pixel-wise} classification task, we employ Intersection over Union (IoU) to measure the overlap between pixel set pairs of ground-truth and prediction.

\begin{itemize}
    % \item Intersection over Union (IoU): This item quantifies the degree of overlap between two ``boxes'' of groundtruth and prediction region. And the threshold to define a positive sample varies from $0.35$ to $0.65$ 
    % with a step size of $0.05$.
    % \item Intersection over Union (IoU). This item quantifies the degree of overlap between two ``boxes'' of groundtruth and prediction region. ``boxes'' can be 2D/3D region proposal box or even pointcloud sets.
    
    \item Average Precision (AP): Our predictions consists of 4 categories: True Positive (TP), True Negative (TN), False Positive (FP) and False Negative (FN). Based on whether our prediction agrees with the corresponding ground truth (T/F) and the condition of our prediction (P/N). Precision (prec) is then calculated as $\frac{|\text{TP}|}{|\text{TP}| + |\text{FP}|}$ to evaluate the likelihood of making false positive reports. %and Recall (Rec) is calculated as $\frac{|\text{TP}|}{|\text{TP}| + |\text{FN}|}$ to evaluate the likelihood of missing positive instances. In fact, precision is a non-negative monotonically decreasing function of recall ($\mathrm{prec}\left(r\right), r \in [0, 1].$) 
    Consequently, AP is defined as the integration over recall: $\text{AP}=\int_{0}^{1} \mathrm{prec}\left(r\right) \mathrm{d}r$.
    
    \item mAP: mAP is calculated by averaging the AP values across different thresholds and categories. These thresholds are based on IoU typically (like DENSE). However in NuScenes, the thresholds are center-distance based represented by a set of thresholds $\mathbb{D}$. The overall object category space, denoted as $\mathbb{C}$, is heavily biased and consists of 10 categories. The mAP can be expressed as:
    \begin{equation}
        \mathrm{mAP}=\frac{1}{|\mathbb{C}||\mathbb{D}|} \sum_{c \in \mathbb{C}} \sum_{d \in \mathbb{D}} \mathrm{AP}_{c, d}
    \end{equation}
    % \item NDS, it can be expressed with:
    % \begin{equation}
    %     \mathrm{NDS}=\frac{1}{10}[5 \cdot\mathrm{mAP}+\sum_{\mathrm{mTP} \in \mathbb{T P}}(1-\min (1, \mathrm{mTP}))]
    % \end{equation}
    % Errors of much more sophisticated factors are considered (\textit{e.g.}, volume, yaw orientation, velocity, \textit{etc}.) in mTP terms, which makes it a more comprehensive one. Take it in mind that mAP is biased to some extend indeed, for the AP of "Car" category, the most frequent instance category (contributes to $40+\%$.) is much higher than that of "Bike" category and these instances are rare to appear. Both metric is bounded between $0$ and $1$, a higher value is preferred for both metrics which are presented in percent format usually.
    % \item nuScenes detection score (NDS). It measures the error of much more sophisticated factors beside AP (\textit{e.g.}, yaw angle orientation, velocity, IoU in 3D space.), formally it is defined as:
    % \begin{equation}
    %     \mathrm{NDS}=\frac{1}{10}[5 \cdot\mathrm{mAP}+\sum_{\mathrm{mTP} \in \mathbb{T P}}(1-\min (1, \mathrm{mTP}))]
    % \end{equation}
    % % TP measures error in "orientation", "velocity", "scale", "center-distance", "attribute(object category)" totally
    % % 5 domain.
    % Where $\mathbb{TP}$ measures the error of aforementioned additional factors, and it can be larger than 1, thus it is bounded between $0$ and $1$.
    \item NDS: NDS is designed to address the limitations of mAP in capturing all aspects of general detection tasks, such as vehicle velocity. To overcome these limitations, it decomposes the detection error into individual normalized metric components, such as translation, orientation, etc.
    % : location, size, orientation, attributes, and velocity. Each of these components is normalized to be between 0 and 1, and their mean is used to compute the NDS. Formally, the NDS is defined as follows:
    % \begin{equation}
    %     \mathrm{NDS}=\frac{1}{10}[5 \cdot\mathrm{mAP}+ \textstyle{\sum_{\mathrm{mTP} \in \mathbb{T P}}}(1-\min (1, \mathrm{mTP}))]
    % \end{equation}
    %Where mTP denotes the meaned TP of all instances for each error aspect, and half of NDS is based on it.
\end{itemize}

All metrics are bounded between 0 and 1, with higher values indicating better performance.
We refer readers to the original papers~\cite{Caesar2020nuScenesAM, Bijelic2020SeeingTF} for more dataset-specific metric details.
% It should be noted that mAP is biased to some extent because the average precision for the ``car'' category, which is the most frequent instance category (contributing to over 40\% of instances), is much higher than that of the ``bike'' category, which is a rarer category. Both metrics are bounded between 0 and 1, with higher values indicating better performance.

%\rev{\subsection{Experimental Setup}}
    % Our network is implement in PyTorch, using the frameworks MMDetection and MMDetection3D\footnote{need github citation.}. For lidar modality, we quantify the space into $\left(0.075, 0.075, 0.2\right) \text{m}^{3}$ voxels, followed by VoxelNet as the backbone. For image modality, we use Swin-T~\cite{Liu2021SwinTH} as backbone to unify image into pseudo- lidar points. Training is conduct on a server with $8\times$ GeForce RTX 3090, while adaptation and assemble process is done on a Dell Vostro 7500 laptop offline. 

% 4 pages
\section{Evaluation}  \label{sec:evaluation}

In this section, we evaluate \name under various sensory variation and modality-missing scenarios.
\subsection{Benchmark}

One primary motivation of \name is to accommodate various distortion inputs. We now summarize several common distortion types which prevail in daily driving conditions:
%, \yuhang{since NuScenes only consist a small subset of samples lies in challenging conditions and DENSE only focus \textit{evaluation} under adverse conditions rather than \textit{train}, we simulate in rational methods:}
\begin{itemize}
        % \item  \paragraph{Lidar in fog}
        \item  \emph{Fog-induced distortion.} Fog affects lidar systems by distorting point clouds at short distances and reducing intensity information. We adopt the approach from \cite{Hahner2021FogSO} to model lidar as a Linear Time Invariant (LTI) system, and calibrate sensors according to the manuals as necessary. We use $\alpha$ to characterize the meteorological optical range (MOR) explicitly and capture the fog density implicitly.   

        \item  \emph{Snow-induced distortion.} Lidar in snow poses unique challenges compared to fog scenarios. In snow conditions, the air can be considered as a low-humidity medium with high-reflectivity snowflakes. 
        % Additionally, reflections from melting and wet ground introduce additional noise. 
        We adopt the approach~\cite{Hahner2022LiDARSS}, and utilize $\beta$ to represent the snowfall rate.
        \item \emph{Motion blur.} This effect is common in images when the vehicle moves rapidly.
        For simplicity, we use a convolution with Gaussian kernel to simulate this blurring effect. A smaller kernel size corresponds to a lower speed and less pronounced blurring, and vice versa.
        
        \item \emph{Exposure condition.} Auto exposure of cameras can result in poor image quality under poor illumination conditions. We use Gamma calibration to characterize different exposure levels. An image is classified as under-exposed for $\gamma \in [0, 1)$ and over-exposed for $\gamma \in (1, +\infty)$.
\end{itemize}

    % Other aforementioned variations are not considered further (e.g., flare, ego vehicle motion), 
    We adopt two models as baselines to demonstrate the capability of \name. The first one is a pre-trained BEVFusion model \textit{without further fine-tuning} (\textbf{w/o ft}). It is trained with a relatively high point density ($\approx 200\text{k points}$) 
    % and fine angular resolution ($\approx 0.4^{\circ}$) lidar input, 
    along with $6\times$ well-exposed RGB images from different perspective resized to $256\times 704$ pixels. 
    % As for the concerned sensor parameters, the training adopts neither fog nor snow condition, $\gamma$ value as $1$, and no motion blur. 
    No distortion is adopted during training, and the framework proposed in \S~\ref{ssec:contrastive_learning} is not applied. We also employ a \textit{fine-tuned model} (\textbf{full-model ft}) as a baseline, all parameters in the model can be fine-tuned.

    \subsection{Object Detection}
    \label{subsec: OD}
    %%%%%%%%
    %%%%%%%%note by yuhang: Prof Luo recommends to re-plot the figures such as Figure~\ref{fig:fog_impact}. which integrates multi metrics into one figure (with two y-axis locate at left and right), and plot a bar-chart indices memory/inference overhead as the right sub-figure to show the superior when compared to Full-model ft.
    %%%%%%%%
    We have demonstrated in \S~\ref{ssec:motivation_sensor_variation} that sensory input variation considerably impacts inference performance. To evaluate the task- and model-agnostic nature of \name, we first focus on the object detection task and lidar modality. We assess its effectiveness by systematically varying each type of lidar input to specific levels. Subsequently, we present the performance of \name for each variation setting.
    % We have shown in Section~\ref{ssec:motivation_sensor_variation} that variation in sensory inputs %(caused by exposure, motion, rotation rate etc) 
    % significantly degrades the inference performance.
    % % since the non-iid distributions during training and inference cannot be readily handled by the DNN model. 
    % To illustrate \name is non task-specific and non model-specific, we first concentrate on object detection task and lidar modality and evaluate the effectiveness by varying each type of lidar input to a specific level. In the following, we show the mAP and NDS of \name under each variation setting.

\begin{figure}[]
    % \vspace{-1.5ex}
    \centerline{
        \subfigure[mAP.]{
            \includegraphics[width = .5 \linewidth]{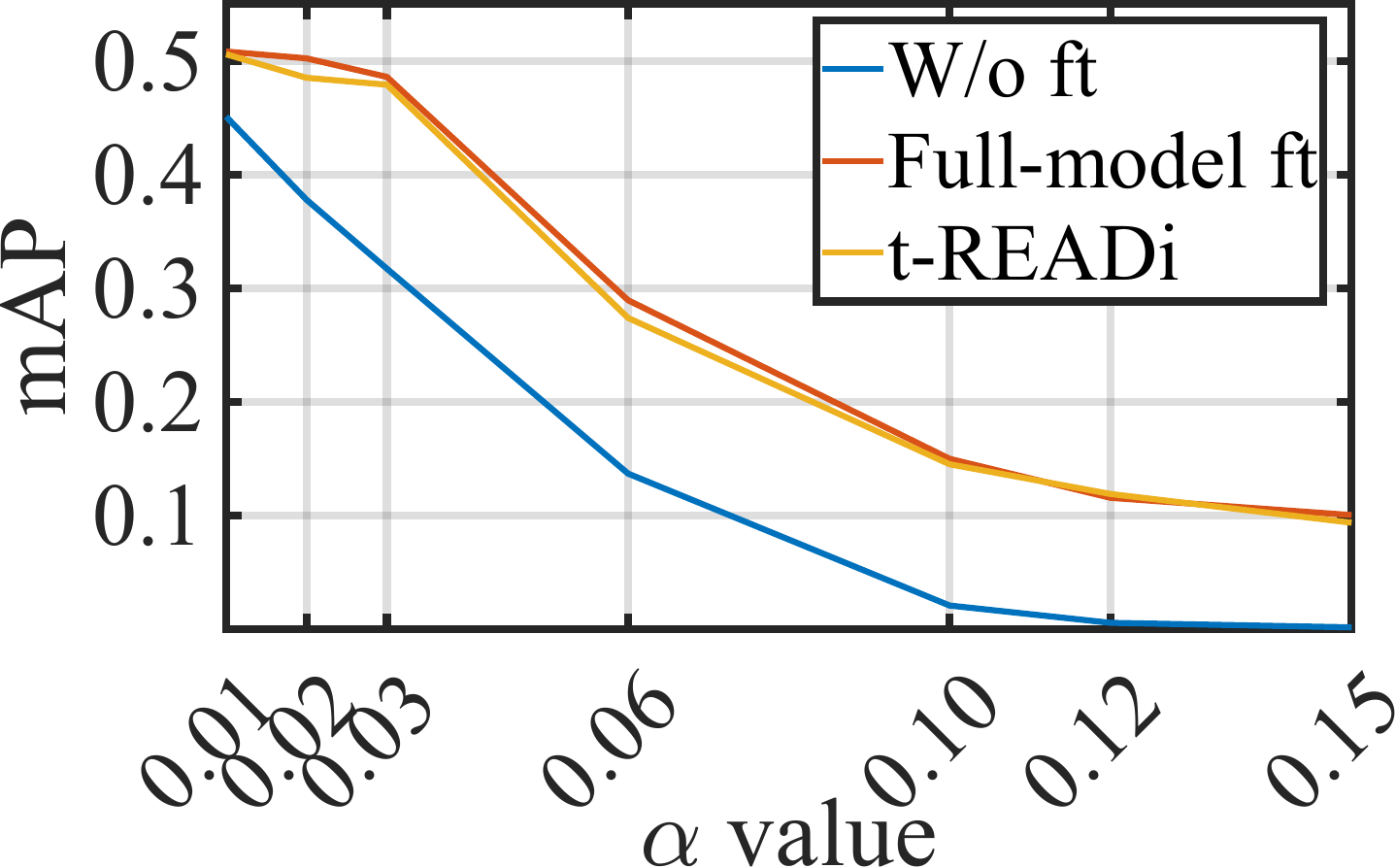}
            \label{subfig:readi_fog_mAP}
        }
        \hspace{-10pt}
        \subfigure[NDS.]{
            \includegraphics[width = .5 \linewidth]{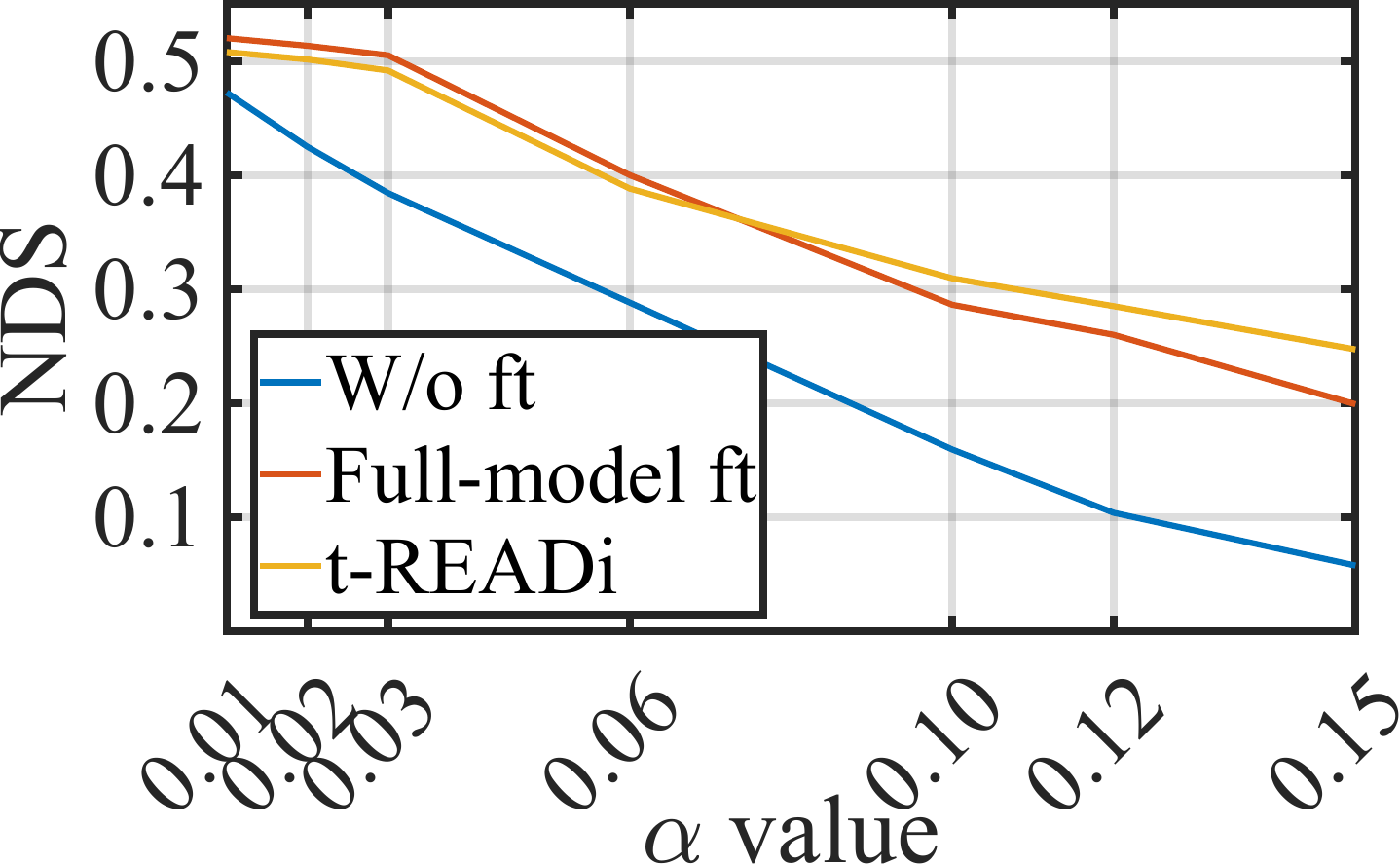}
            \label{subfig:readi_fog_NDS}
        }
    }
    \vspace{-2ex}
    \caption{\name maintains high performance when fog density deviates from pre-training.}
    \label{fig:fog_impact}
    \vspace{-2ex}
\end{figure}

%\vspace{-1ex}
\paragraph{Effects of Fog-induced Distortion.}
We introduce noise points by simulating scattering using fog density indicators of $\alpha \in \{0.01, 0.02, 0.03, 0.06, 0.1, 0.12, 0.15\}$. As a rule of thumb, when $\alpha=0.06$, the MOR is approximately $50$~\!m. Figure~\ref{fig:fog_impact} demonstrates the high inference accuracy of \name across all quantified fog density levels. In contrast, w/o ft experiences significant accuracy degradation when the fog density deviates from the training data. \name achieves $15\%$ average accuracy gain over the w/o ft model while only $2\%$ average accuracy loss over the full-model ft, but with much higher memory efficiency, at a cost of only $2.6\%$ memory and $2.1\%$ inference time overhead respectively. Our key observations are as follows. First, training a model from scratch with severely deviated data (e.g., $\alpha>0.1$) is risky and results in corruption at an early stage, leading to a result far from full-model ft. Second, only a small fraction of distorted points ($\approx5\%$ on average when $\alpha=0.03$) significantly impairs the inference accuracy. In these scenarios, radar data can provide more valuable insights.
% The result in Figure~\ref{fig:fog_impact} and Figure~\ref{subfig: OD_overhead} confirms the high inference accuracy of \name on all quantified fog density levels. In contrast, w/o ft suffers from significant accuracy degradation once the fog density deviates from the training one. \name achieves $15\%$ average accuracy gain over the w/o ft model while only $2\%$ average accuracy loss over the full-model ft, but with much higher memory efficiency, at a cost of only $2.6\%$ memory and $2.1\%$ inference time overhead respectively. We have two key observations: i) Training a model from scratch with severe deviate data (e.g., $\alpha>0.1$) is dangerous and will be corrupted in early stage and just yield a result far from tuned from a well pretrained one. ii) Only a very small fraction distorted points (e.g., only about $5\%$ on average when $\alpha=0.03$) will hamper the inference accuracy greatly, which indicates practical AV framework is a open question and radar can provide more insight in these scenarios.
% $\ln\left(20\right)/\alpha$
\begin{figure}[]
    % \vspace{-1.5ex}
    \centerline{
        \subfigure[mAP.]{
            \includegraphics[width = .5 \linewidth]{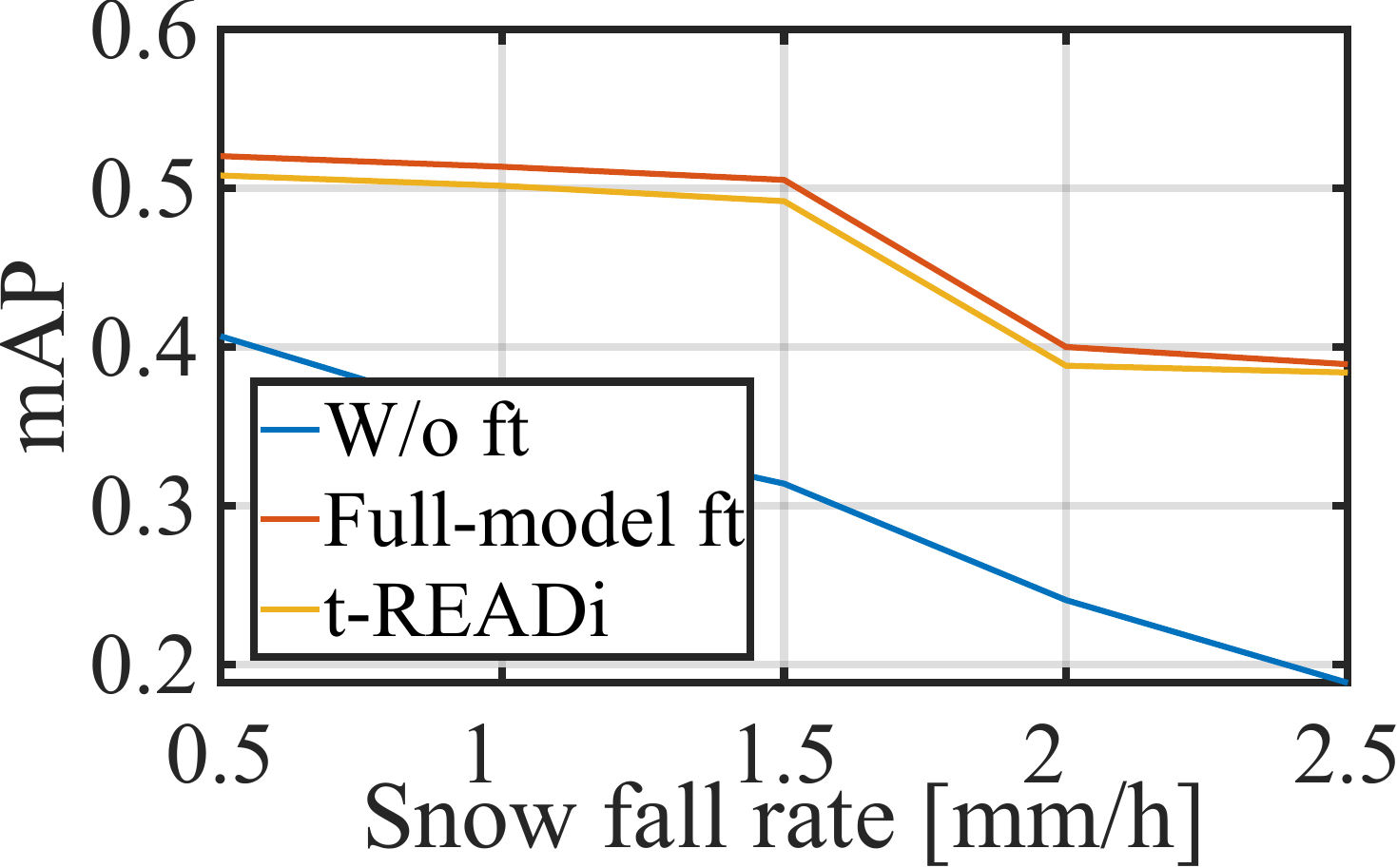}
            \label{subfig:readi_snow_mAP}
        }
        \hspace{-10pt}
        \subfigure[NDS.]{
            \includegraphics[width = .5 \linewidth]{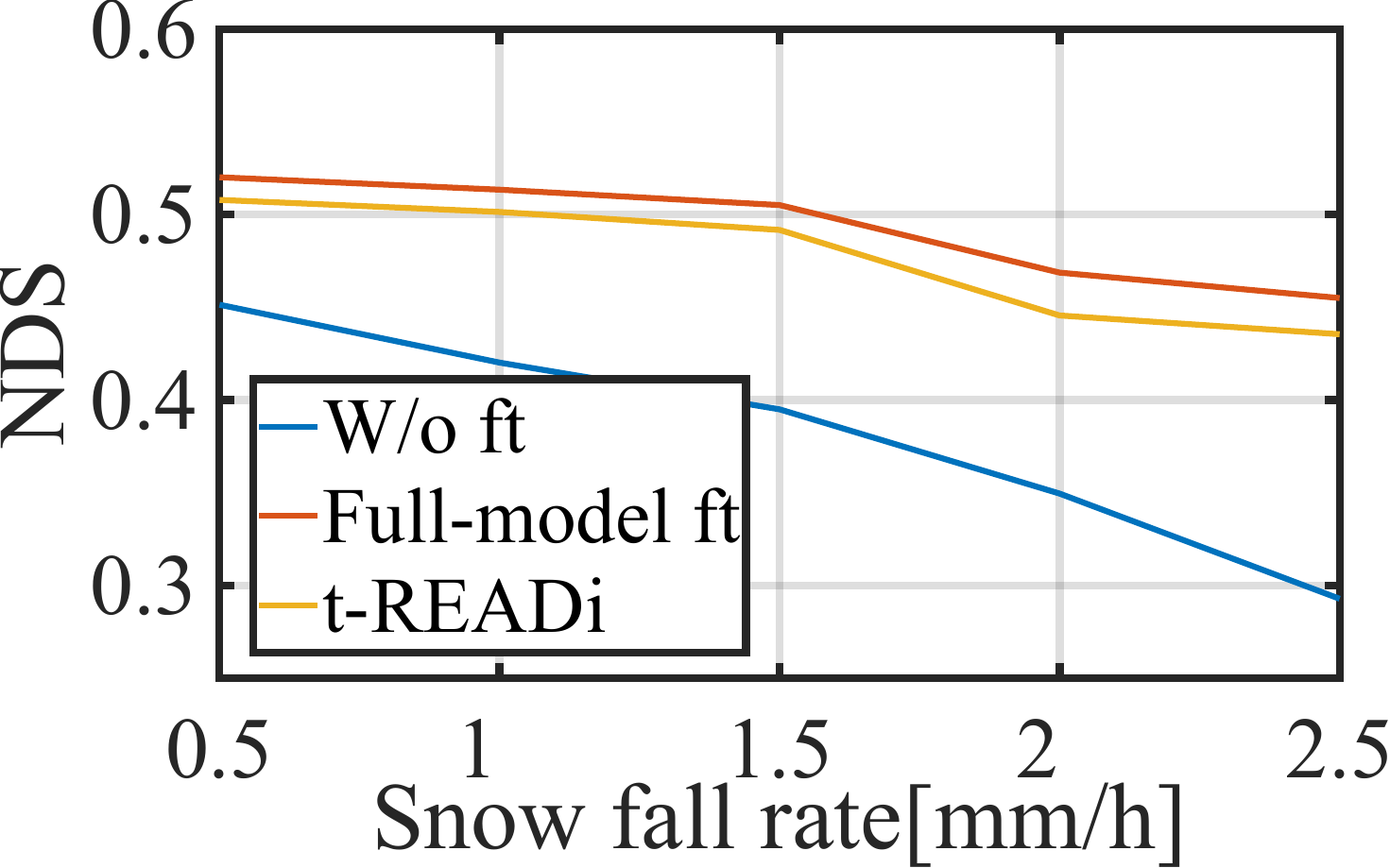}
            \label{subfig:readi_snow_NDS}
        }
    }
    \vspace{-2ex}
    \caption{\name maintains high performance when snow fall rate deviates from pre-training.}
    \label{fig:snow_impact}
    \vspace{-2ex}
\end{figure}

%\vspace{-1ex}
\paragraph{Effects of Snow-induced Distortion}
We generate snowflake-scattered and high-reflectivity points with the snowfall rate indicator $\beta \in \{0.5, 1.0, 1.5, 2.0, 2.5\}$. Figure~\ref{fig:snow_impact} demonstrates the high inference accuracy of \name across all quantified snowfall rate levels. \name achieves an average accuracy loss of only $2\sim3\%$ compared to the full-model ft, while offering higher memory efficiency. We note that snowflake scatter has less impact than reflection from melted snow, this also explains why fog affects lidar more than snow.

% We precompute snowflakes in the space and make scatters with high reflectivity where snow fall rate indicator $\beta \in \{0.5, 1.0, 1.5, 2.0, 2.5\}$. The result in Figure~\ref{fig:snow_impact} confirms the high inference accuracy of \name on all quantified snow fall rate levels, where \name achieves only $2\sim3\%$ average accuracy loss over the full-model ft, but with much higher memory efficiency. The effect of snowflake scatter is not as critical as melted snow reflect, This answers for the reason why heavy fog will hampers lidar more when compare to heavy snow. 
\begin{table*}[t]
\caption{3D average precision on DENSE dense fog split}
\vspace{-1.5ex}
\renewcommand{\arraystretch}{0.8}
\small
\centering
\begin{tabular}{llccc}
\hline
Model & Method                & Car AP@.5IoU            & Cyclist AP@.25IoU       & Pedestrian AP@.25IoU \\
                       & & easy    mod     hard    & easy    mod     hard    & easy    mod     hard  \\
\hline
PointRCNN~\cite{Shi_2019_CVPR} &  train with clear:       & 42.43   42.24   40.29   & 22.52   23.52   25.62   & 43.23   40.16   37.05 \\
& t-READi:                & 47.94   46.07   42.07   & 27.60   27.65   29.21   & 45.52   43.38   41.10 \\
& full ft:                & 48.52   46.77   42.19   & 27.66   27.94   29.37   & 45.65   43.23   41.15 \\
& train with dense fog:       & 49.31   47.33   42.94   & 27.89   27.89   29.29   & 45.79   43.47   41.33 \\
\hline
PV-RCNN~\cite{shi2020pv} & train with clear:       & 40.19   40.93   39.66   & 24.33   24.63   24.63    &42.67   41.04   39.59\\   

& t-READi:                & 46.69   47.38   46.51   & 29.63   28.50   28.22 &46.03   45.23   43.76\\

& full ft:                & 47.36   47.55   46.89   & 29.92   29.01   27.54       &46.51   45.44   44.91\\

& train with dense fog:   & 47.81   47.86   47.12       & 30.11   29.42   27.91       &46.73   45.61   45.12\\
\hline
\end{tabular}
\vspace{-2.5ex}
\label{tbl:test_with_dense}
\end{table*}

\subsection{Primary Results on DENSE Dataset}
To justify the methodology of \name as neither dataset- nor model-specific, we focus on lidar modality under practical dense fog conditions. We evaluate our approach using PointRCNN~\cite{Shi_2019_CVPR} and PV-RCNN~\cite{shi2020pv}, which differ from the model used in \S~\ref{subsec: OD} in terms of how raw points are represented and how proposals are generated. We train each model on the clear-split dataset (model $M_{c}$) and the dense-fog-split dataset from scratch. We then fine-tune or apply \name to $M_{c}$ and compare their performance on the dense-fog-split dataset.

The results of the class-wise AP metric are shown in Table~\ref{tbl:test_with_dense}, and we chose $r=4$ and $k=0$ as the parameters for \name, as neither is based on transformer. An interesting observation is dataset-specific: for DENSE, when testing a model trained on clear data with severe adverse weather data, the performance degradation ($\approx7$AP) is much smaller than that on NuScenes ($\approx15$mAP). We hypothesize that the multi-lidar setup in DENSE enhances its robustness and that the weather-based split achieves better consistency within the dataset. Despite this, \name demonstrates its generalizability: for every class, it incurs at most a $1\sim1.5\%$ AP loss compared to the full-model fine-tuning approach. Moreover, for subtle instances occupied by fewer points, the performance gap between \name and full-model fine-tuning is even more marginal.

\subsection{Semantic Segmentation}
\begin{figure}[]
%\vspace{-1.5ex}
    \centerline{
        \subfigure[Under various blur levels.]{
            \includegraphics[width = .5 \linewidth]{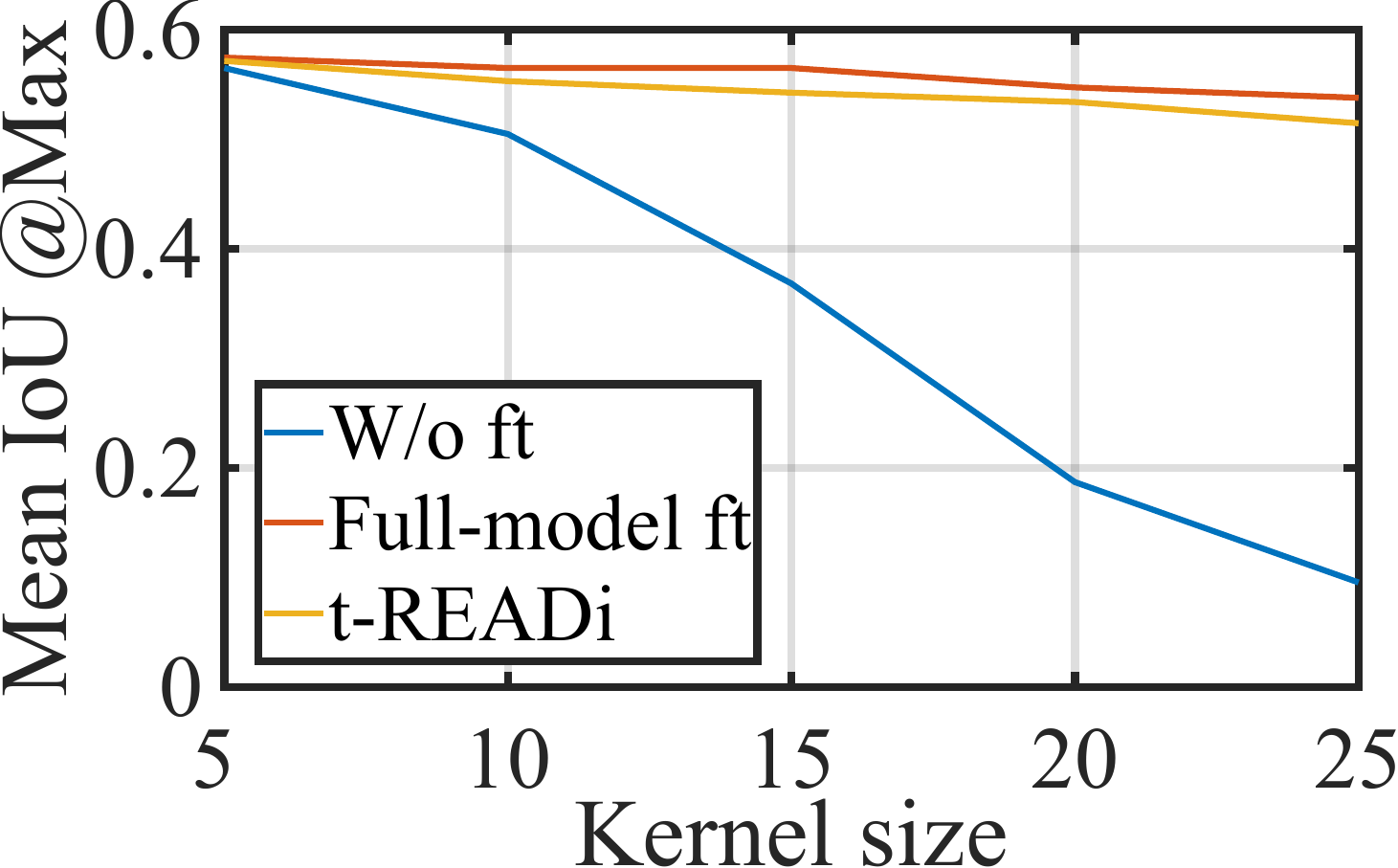}
            \label{subfig: readi_blur_seg}
        }
        \hspace{-10pt}
        \subfigure[Under various exposure levels.]{
            \includegraphics[width = .5 \linewidth]{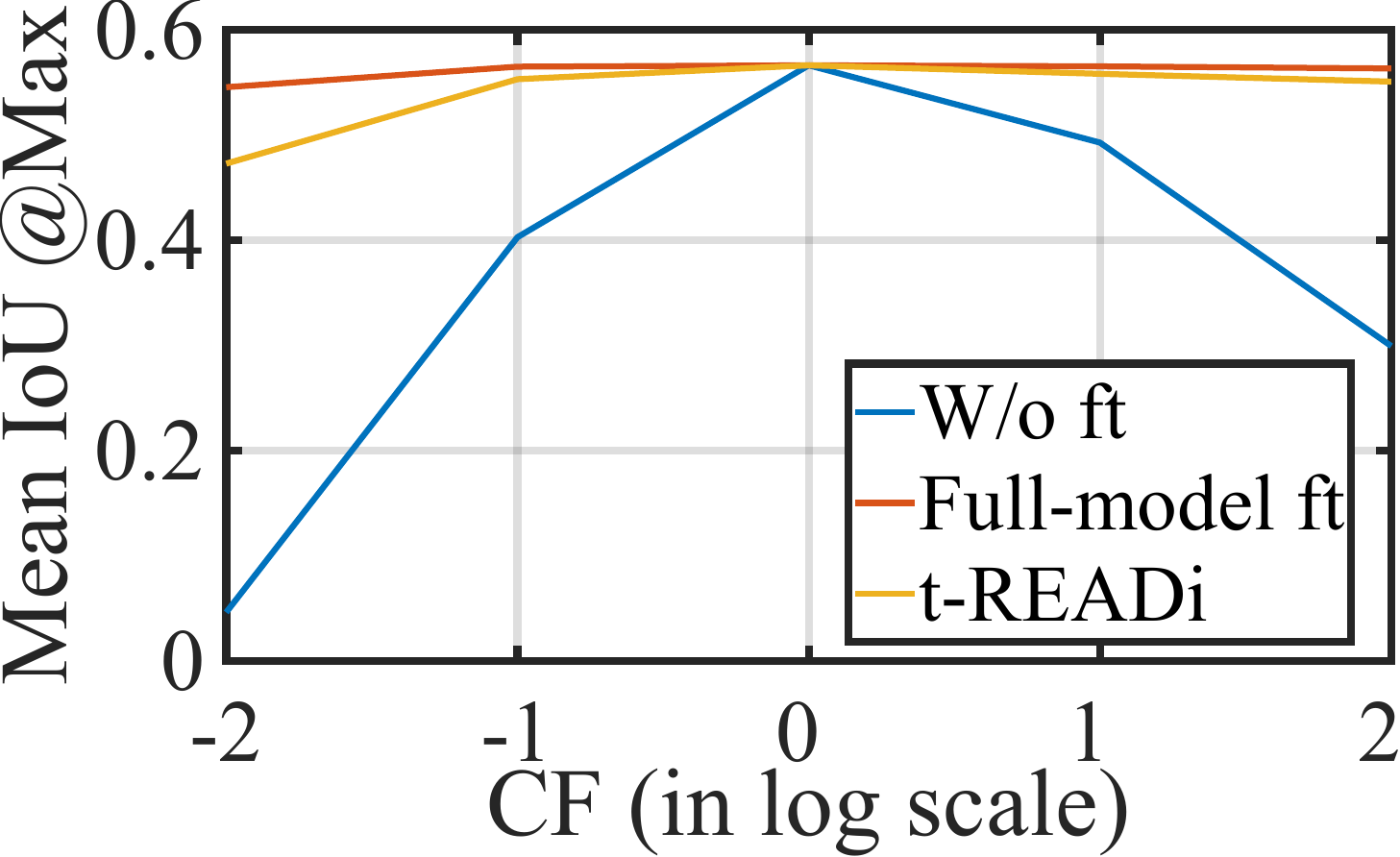}
            \label{subfig: readi_exposure_seg}
        }
    }
    \vspace{-2ex}
    \caption{\name maintains high performance when camera modality
    performance deviates from pre-training.}
    \label{fig: semantic_seg}
    \vspace{-2ex}
\end{figure}

\begin{figure}[]
%\vspace{-1.5ex}
    \centerline{
        \subfigure[Memory \& Load overhead.]{
            \includegraphics[width = .45 \linewidth]{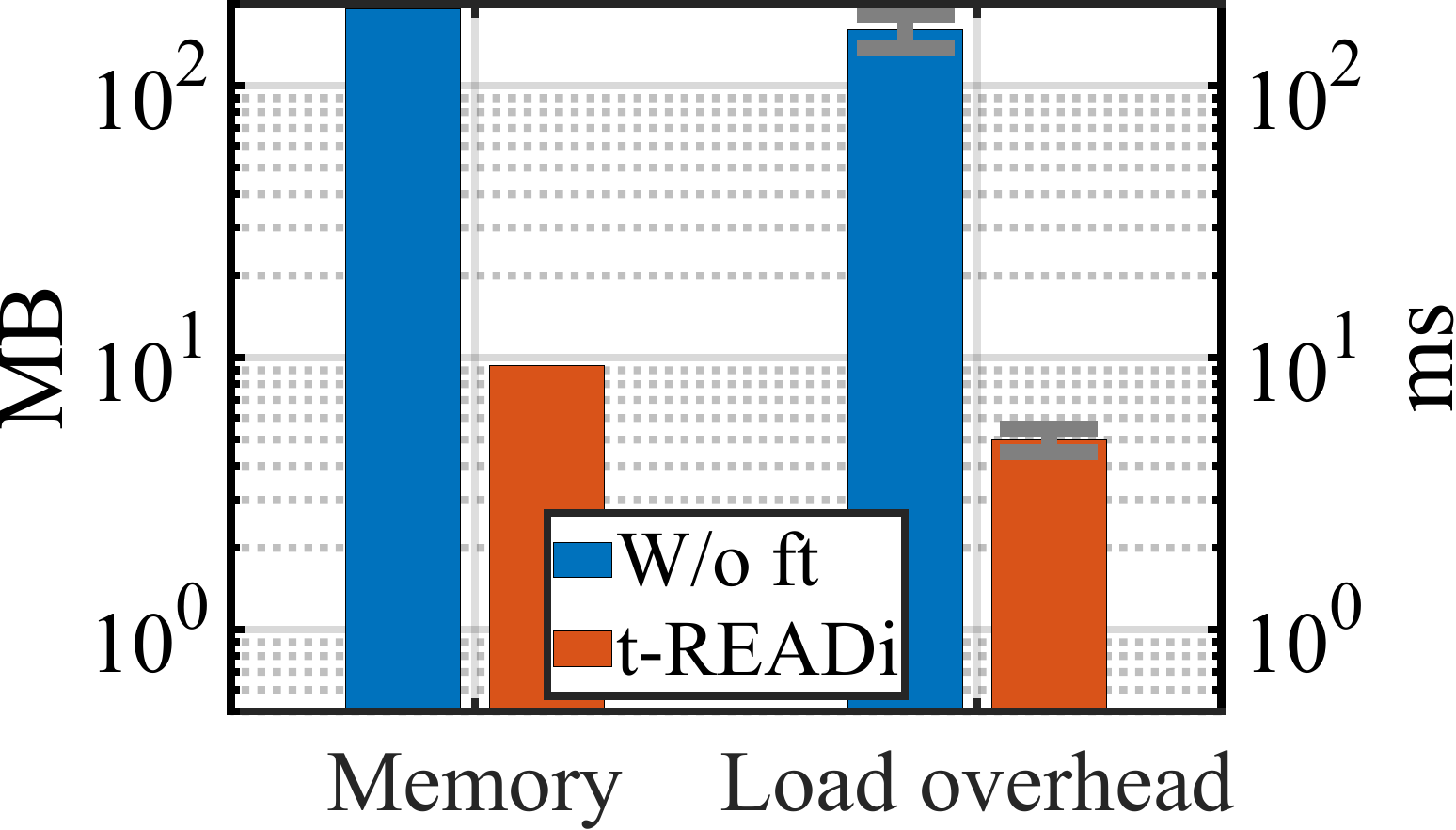}
            \label{subfig: Seg_overhead}
        }
    
    \hspace{-10pt}
    
        \subfigure[Inference overhead.]{
            \includegraphics[width = .55 \linewidth]{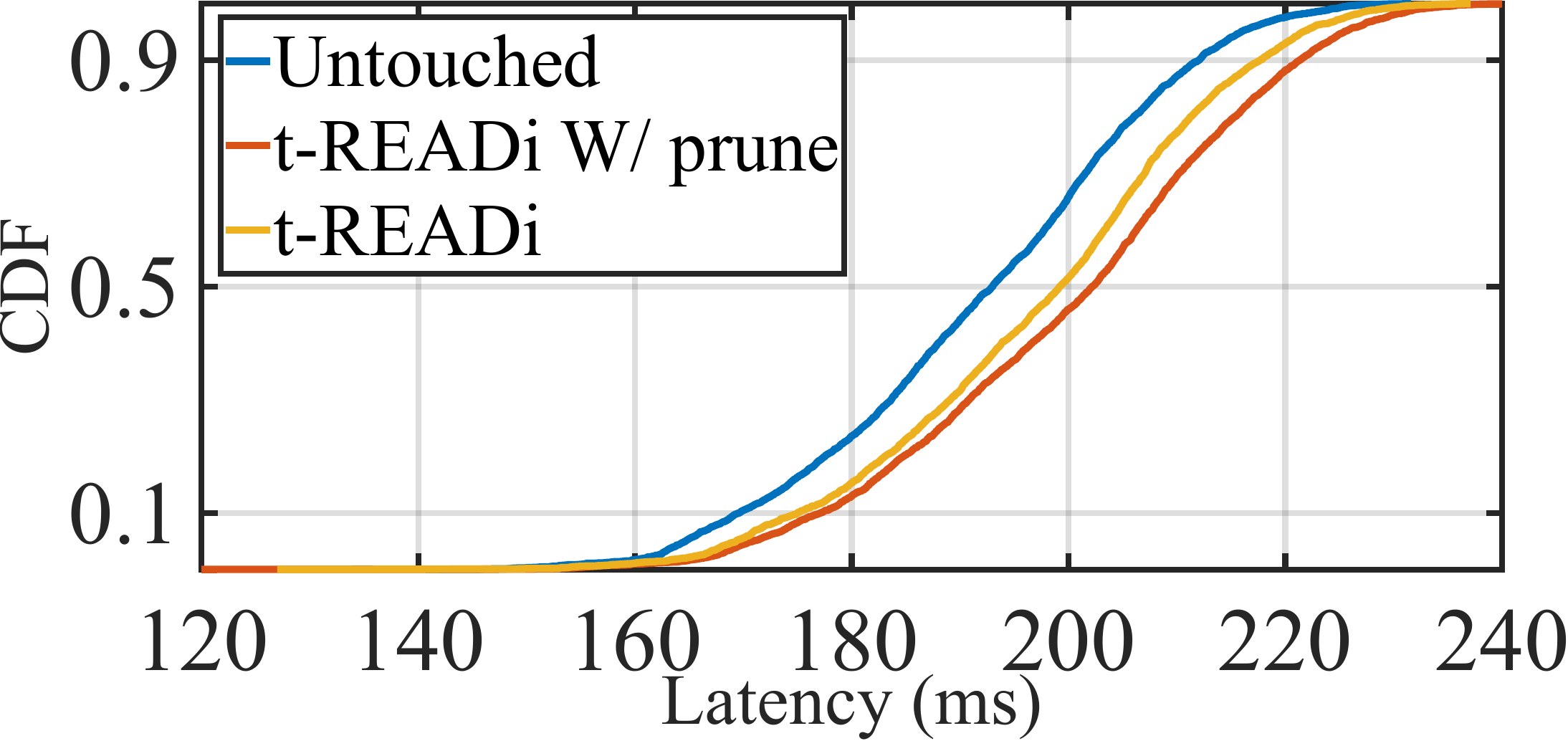}
            \label{subfig: inference_overhead}
        }
    }
    \vspace{-2ex}
    \caption{\name suffers from marginal overhead in additional memory or load/inference time.}
    \label{fig: overheads}
    \vspace{-2ex}
\end{figure}

We focus on the camera modality in the segmentation task, where the pre-trained model relies heavily on transformers. We pick the maximum IoU under each variation setting.
% We now focus on camera modality in segmentation task, where the pretrained model is much more transformer-intensive. In the following, we show the category-reduced (i.e., drivable area, carpark area, stop line, divider, walkway, and crosswalk) maximum IoU under each variation setting.
%\vspace{-1ex}
\paragraph{Effects of Motion Blur}
We apply different kernel sizes chose from $\{5, 10, 15, 20, 30\}$ to the original datasets to obtain blurred datasets.
% which actually emulate the case where the camera is compromised by vehicle ego. 
%%%%%%%%
%%%%%%%%note by yuhang: focal length of camera doesn't change at all in vehicle.
%%%%%%%%
In Figure~\ref{subfig: readi_blur_seg} and Figure~\ref{subfig: Seg_overhead}, \name demonstrates robustness even under heavy distortion, with only slight $1\sim2\%$ IoU loss compared to the full-model ft approach. Moreover, \name achieves this by updating only $3\%$ of the parameters. This approach also offers higher memory efficiency, with a memory overhead of $4.9\%$ and an load time overhead of $5.8\%$. On average, it takes 160~\!ms with a standard deviation of 25~\!ms to load a 200~\!MB model, whereas \name completes the switch within 10~\!ms in most cases.
% Figure~\ref{subfig: readi_blur_seg} and Figure~\ref{subfig: Seg_overhead} shows even under heavy distortion where pre-trained model functions very badly, \name achieves only $1\sim2\%$ IoU loss over the full-model ft, while it just updates $3\%$ parameters compared to that, with much higher memory efficiency, at a cost of only $4.9\%$ memory and $5.8\%$ inference time overhead respectively.

%

% \begin{figure}[]
% %\vspace{-1.5ex}
%     \centerline{
%         \subfigure[Object Detection variant.]{
%             % \includegraphics[width = .5 \linewidth]{tmp_imgs/OD_overhead.pdf}
%               \includegraphics[width = .5 \linewidth]{tmp_imgs/latency_multimodal.pdf}
%             \label{subfig: OD_overhead}
%         }
%         \subfigure[Segmentation variant.]{
%             \includegraphics[width = .5 \linewidth]{tmp_imgs/Seg_overhead.pdf}
%             \label{subfig: Seg_overhead}
%         }
%     }
%     \vspace{-2ex}
%     \caption{\name suffers from marginal overhead in additional memory or load/inference time.}
%     \label{fig: overheads}
%     \vspace{-2ex}
% \end{figure}

%\vspace{-1ex}
\paragraph{Effects of Poor Exposure}
Exposure variations in images can blur the boundaries between different textured zones. We apply pixel-wise gamma calibration with values of $\gamma \in\{0.25, 0.5, 1, 2, 4\}$ (Where "\textbf{CF}" denotes Calibration Factor for short in Figure~\ref{subfig: readi_exposure_seg}). To our surprise, when the image is heavily under-exposed (e.g., $\gamma=0.25$), \name shows a significant margin of $7$ IoU points compared to the full-model ft. 
% This can be attributed to the fact that \name tunes fewer parameters in the segmentation-oriented model, requiring more time to converge. 
Additionally, the properties of under-exposure effects may differ from over-exposure effects, as indicated by the marginal gap of approximately $1.5$ IoU points at $\gamma=4$. Furthermore, we observed that both \name and the full-model ft can improve segmentation performance when the camera is pushed to its limits, whereas this trend is not observed for detection and lidar modality. This discrepancy can be attributed to the relatively simple and less lossy compression applied to the camera-generated variations, while the fog/snow simulation on lidar introduces more noise. Exploring a more comprehensive framework for modeling the behavior of cameras under adverse conditions would be an interesting avenue for future research.

\subsection{Discussions on Inference Latency}
\label{subsec: IL}

\name significantly mitigates memory and load time overhead in scenarios necessitating multiple model variants, while the exact inference time per frame can be compensated. Leveraging the \verb|torch.profiler| interface in PyTorch, we compare the overall inference time distributions between the original model and its \name-enhanced counterpart, as illustrated in Figure~\ref{subfig: inference_overhead}. 
\name introduces an approximate $3.2\%$ overhead in both averaged and tail latency, while this increment exceeds the parameter injection overhead (calculated as $4.9\% - 3\% = 2.9\%$), it remains modest. Surprisingly, the tail-latency ($90$th percentile) to averaged-latency ratio is notably large ($\approx1.12$). Upon profiling the latency contributed by the encoder from the camera modality, we observe a corresponding ratio as low as $1.04$. Furthermore, we identify that the prolonged tail latency primarily stems from the quantization pipeline in the lidar modality encoder, which is beyond the scope of \name.

In addition, considering the significant variability in latency as advocated by~\cite{Lin2018Architectural}, tail latency emerges as a more pertinent metric for meeting real-time constraints. To alleviate tail latency within the context of \name, we take a simple step, employ pruning in the vast frozen layers. Specifically, we reset all parameters whose absolute value is less than $1\times10^{-3}$, resulting in approximately $10\%$ parameter pruning. We find that this approach nearly halves the overhead with no discernible accuracy degradation.
Moreover, there exists potential for further reduction in tail latency by applying more advanced techniques such as parameter quantization~\cite{zhu2017trained} to the vast frozen layers, along with the implementation of more efficient lidar point cloud quantization operators.

\subsection{Missing Sensing Modalities}

We have shown in \S~\ref{ssec:motivation_missing_modality} that missing modality incurs information loss, and filling the missing modality with 0's does not help because doing so only introduces bias into the network. Correspondingly, we propose to improve \name's robustness to missing modality by contrastive learning. We evaluate the robustness of \name to missing modalities by removing a certain number of sensors during inference and observe how the DNN's performance drops. In the following, we show the mAP and NDS of \name under different missing modality scenarios. On the $x$-axis of the figures, ``Full'' denotes there are no missing sensors, and ``number+C/L'' denotes the number of remaining cameras and lidars (there is a total of 6 cameras and 1 lidar). 

\begin{figure}[t]
    \centerline{
        \setcounter{subfigure}{0}
        \subfigure[mAP.]{
            \includegraphics[width = .5 \linewidth]{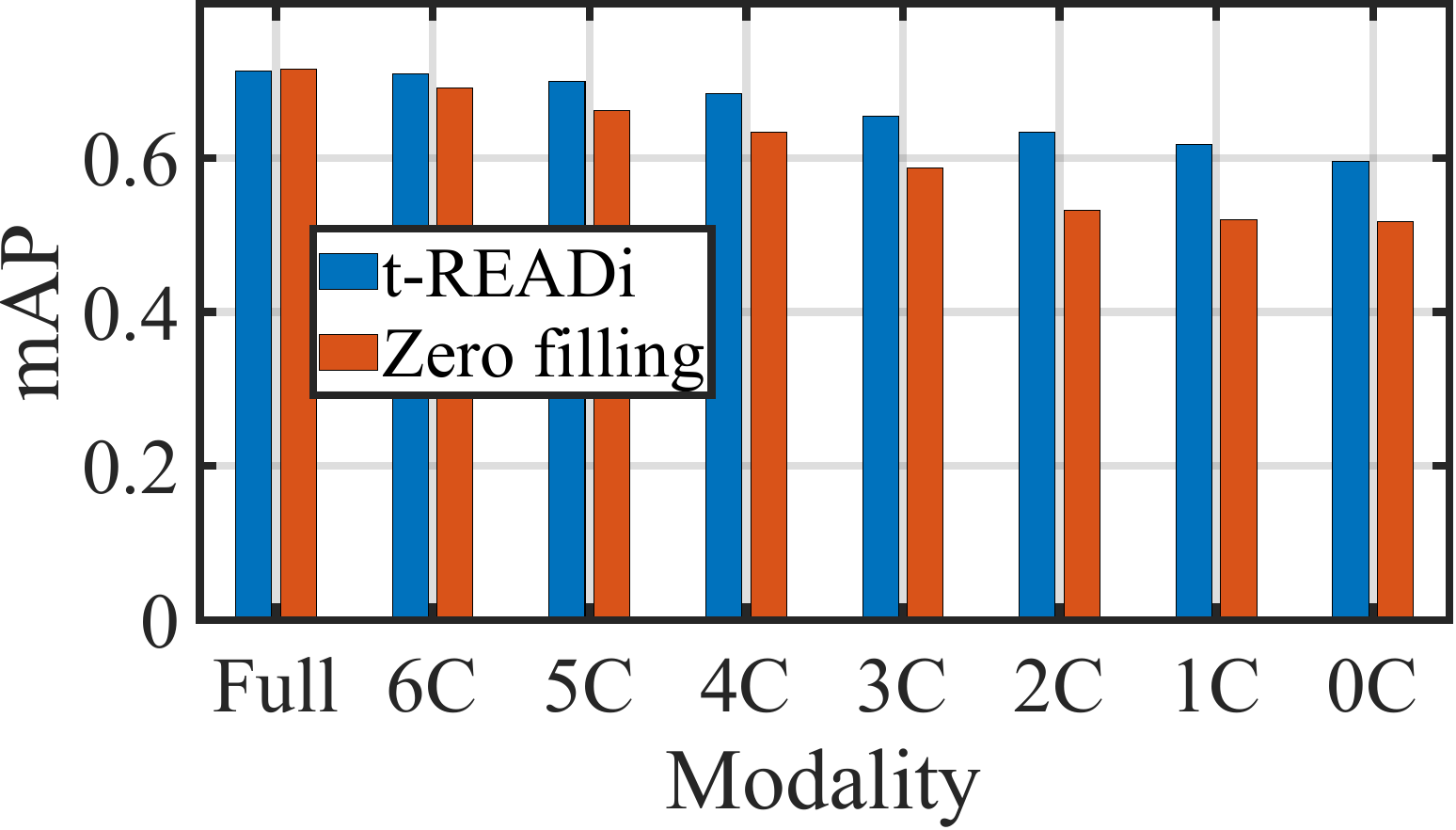}
            \label{subfig:camera_failure_map}
        }
        % \subfigure[NDS.]{
        %     \includegraphics[width = .5 \linewidth]{tmp_imgs/532_rev.pdf}
        %     \label{subfig:camera_failure_nds}
        % }
    \hspace{-10pt}
        % \setcounter{subfigure}{0}
        % \subfigure[mAP.]{
        %     \includegraphics[width = .5 \linewidth]{tmp_imgs/531_rev.pdf}
        %     \label{subfig:camera_failure_map}
        % }
        \subfigure[NDS.]{
            \includegraphics[width = .5 \linewidth]{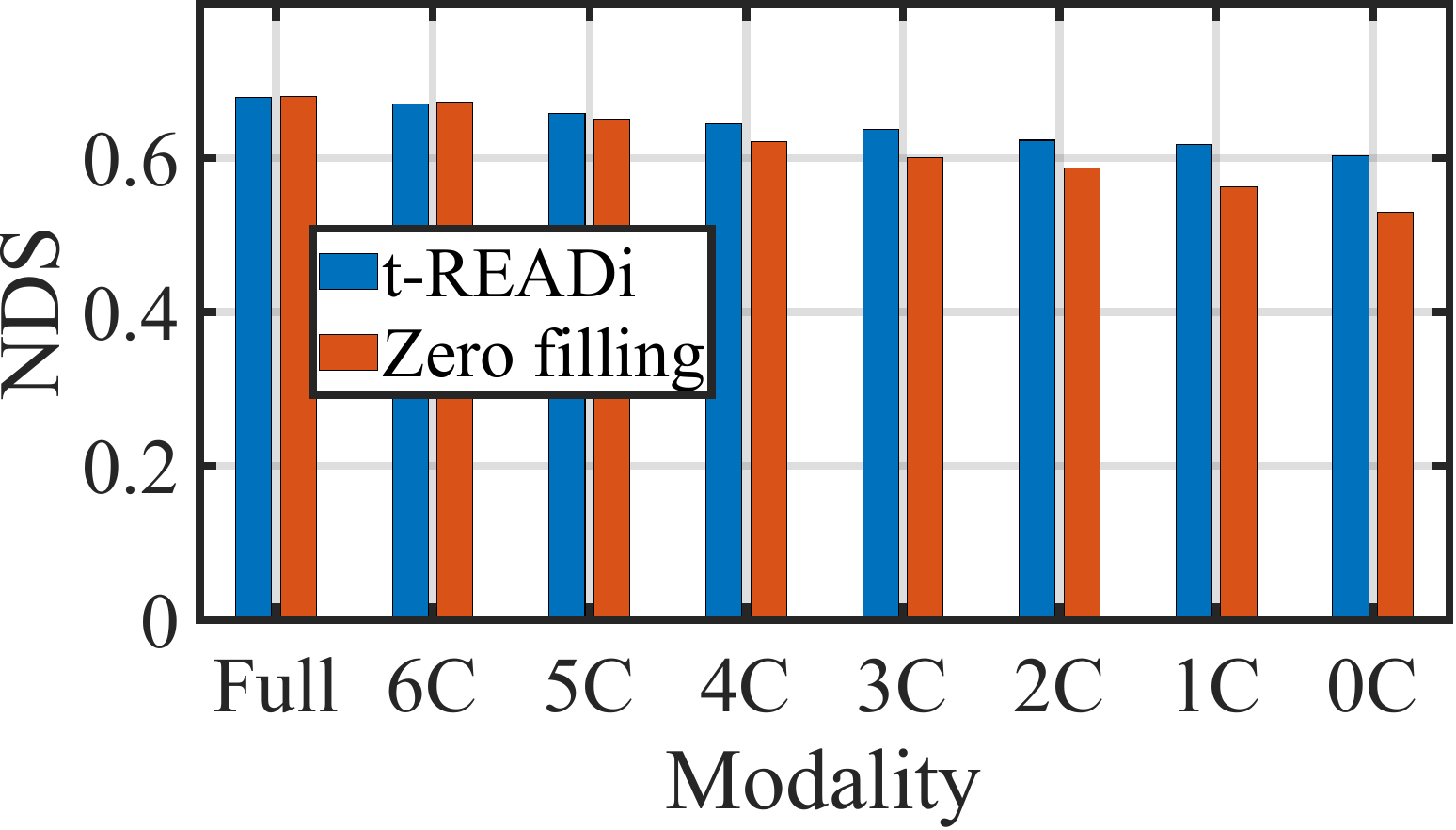}
            \label{subfig:camera_failure_nds}
        }
    }
    \vspace{-2.5ex}
    \caption{\name compensates for the information loss of missing cameras.}
    \label{fig:camera_failure}
    \vspace{-2.5ex}
\end{figure}

%\vspace{-1ex}
\paragraph{Missing camera.} We first present how our contrastive learning framework deals with missing cameras. It can be seen in Figure~\ref{fig:camera_failure} that mAP and NDS under contrastive learning decrease from 0.71 to 0.59 and from 0.68 to 0.60, respectively, and the mAP and NDS under zero filling decrease from 0.72 to 0.52 and from 0.68 to 0.53, respectively. Although we can still observe mAP and NDS decrease when contrastive learning is adopted, the decreased mAP and NDS is smaller than those of zero filling, proving \name's robustness to missing modalities. Moreover, we find that in some cases (e.g., ``3C'' for zero filling and ``1C'' for \name), \name can achieve comparable performance with zero filling with fewer camera sensors. The result shows that fewer sensors can be used by \name to achieve the same performance than previous solutions, thus proving \name's better efficiency. 

% \begin{figure}[t]
%     \centerline{
%         \setcounter{subfigure}{0}
%         \subfigure[mAP.]{
%             \includegraphics[width = .5 \linewidth]{tmp_imgs/531_rev.pdf}
%             \label{subfig:camera_failure_map}
%         }
%         \subfigure[NDS.]{
%             \includegraphics[width = .5 \linewidth]{tmp_imgs/532_rev.pdf}
%             \label{subfig:camera_failure_nds}
%         }
%     }
%     \vspace{-2.5ex}
%     \caption{\name compensates for the information loss of missing cameras.}
%     \label{fig:camera_failure}
%     \vspace{-2.5ex}
% \end{figure}

\begin{figure}[]
    \centerline{
        \subfigure[mAP.]{
            \includegraphics[width = .5 \linewidth]{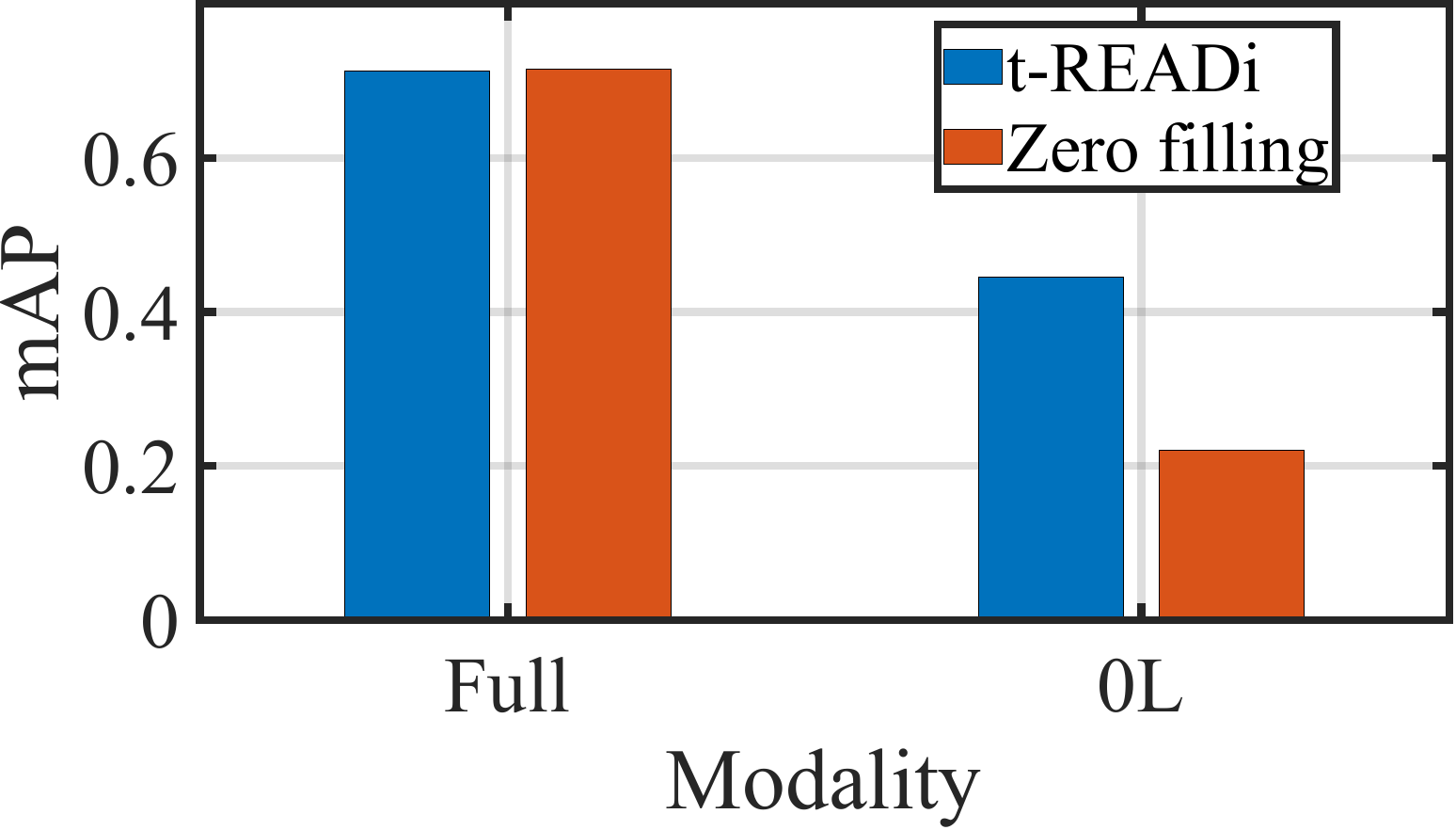}
            \label{subfig:lidar_failure_map}
        }
        % \subfigure[NDS.]{
        %     \includegraphics[width = .5 \linewidth]{tmp_imgs/534_rev.pdf}
        %     \label{subfig:lidar_failure_nds}
        % }
        \hspace{-10pt}
        % \subfigure[mAP.]{
        %     \includegraphics[width = .5 \linewidth]{tmp_imgs/533_rev.pdf}
        %     \label{subfig:lidar_failure_map}
        % }
        \subfigure[NDS.]{
            \includegraphics[width = .5 \linewidth]{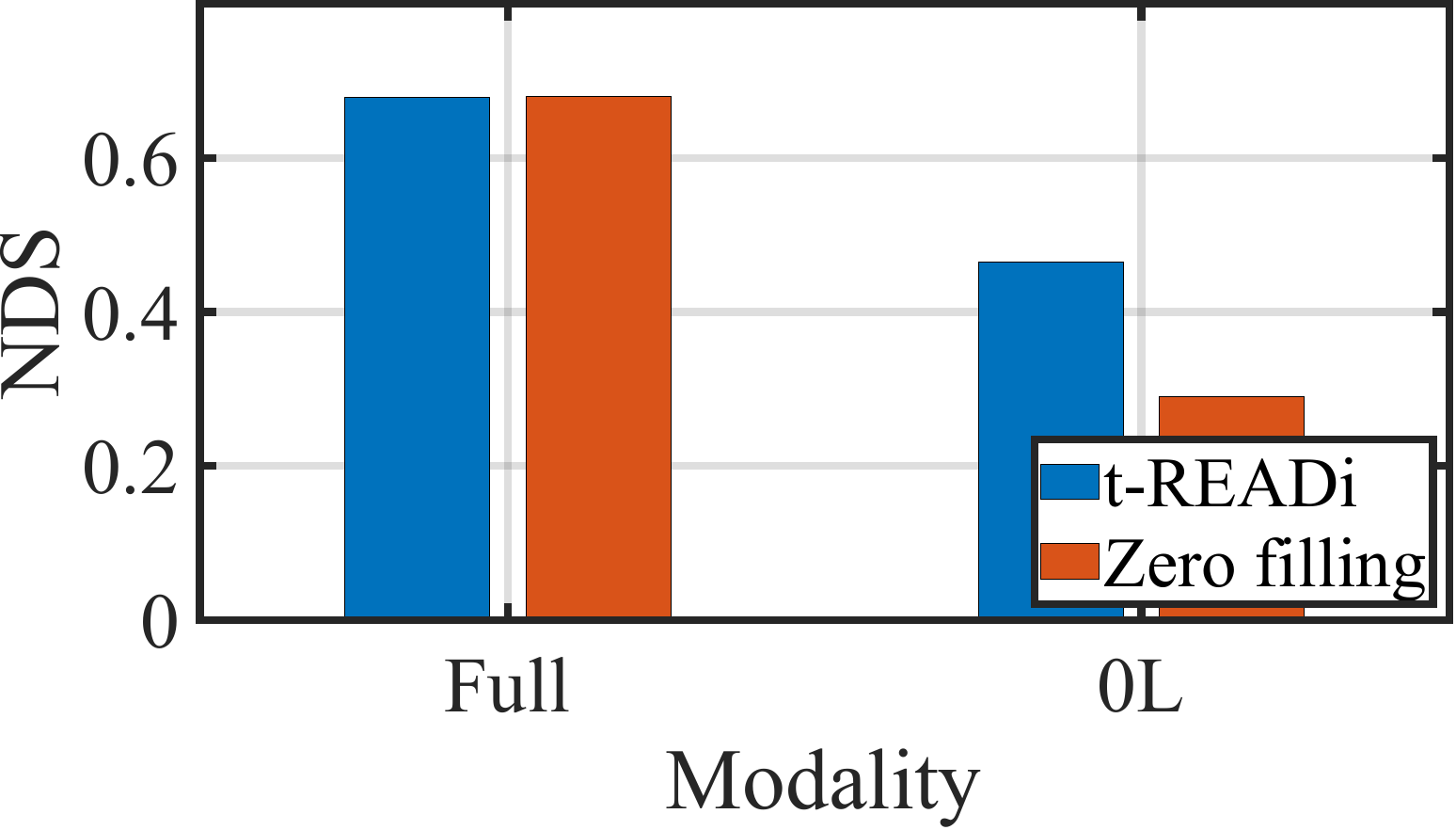}
            \label{subfig:lidar_failure_nds}
        }
    }
    \vspace{-2.5ex}
    \caption{\name compensates for the information loss of missing lidar.}
    \label{fig:lidar_failure}
    \vspace{-2.5ex}
\end{figure}

%\vspace{-1ex}
\paragraph{Missing lidar.} We then demonstrate \name's robustness to missing lidar. As Figure~\ref{subfig:lidar_failure_map} and~\ref{subfig:lidar_failure_nds} show, removing the lidar and filling the missing modality with 0's renders the system unusable since it only gives an mAP and NDS of 0.22 and 0.29, respectively. The poor performance can be attributed to the fact that lidar is the dominant sensor and removing it results in major information loss. As for the contrastive learning adopted by \name, we find that while maintaining the same performance under full modalities, it greatly improves \name's performance with missing lidar. The result demonstrates that the contrastive learning of \name can effectively compensate for missing lidar information by learning from the complementary radar modality.

% \begin{figure}[]
%     \centerline{
%         \subfigure[mAP.]{
%             \includegraphics[width = .5 \linewidth]{tmp_imgs/533_rev.pdf}
%             \label{subfig:lidar_failure_map}
%         }
%         \subfigure[NDS.]{
%             \includegraphics[width = .5 \linewidth]{tmp_imgs/534_rev.pdf}
%             \label{subfig:lidar_failure_nds}
%         }
%     }
%     \vspace{-2.5ex}
%     \caption{\name compensates for the information loss of missing lidar.}
%     \label{fig:lidar_failure}
%     \vspace{-2.5ex}
% \end{figure}

\begin{figure}[]
    \centerline{
        \subfigure[mAP.]{
            \includegraphics[width = .5 \linewidth]{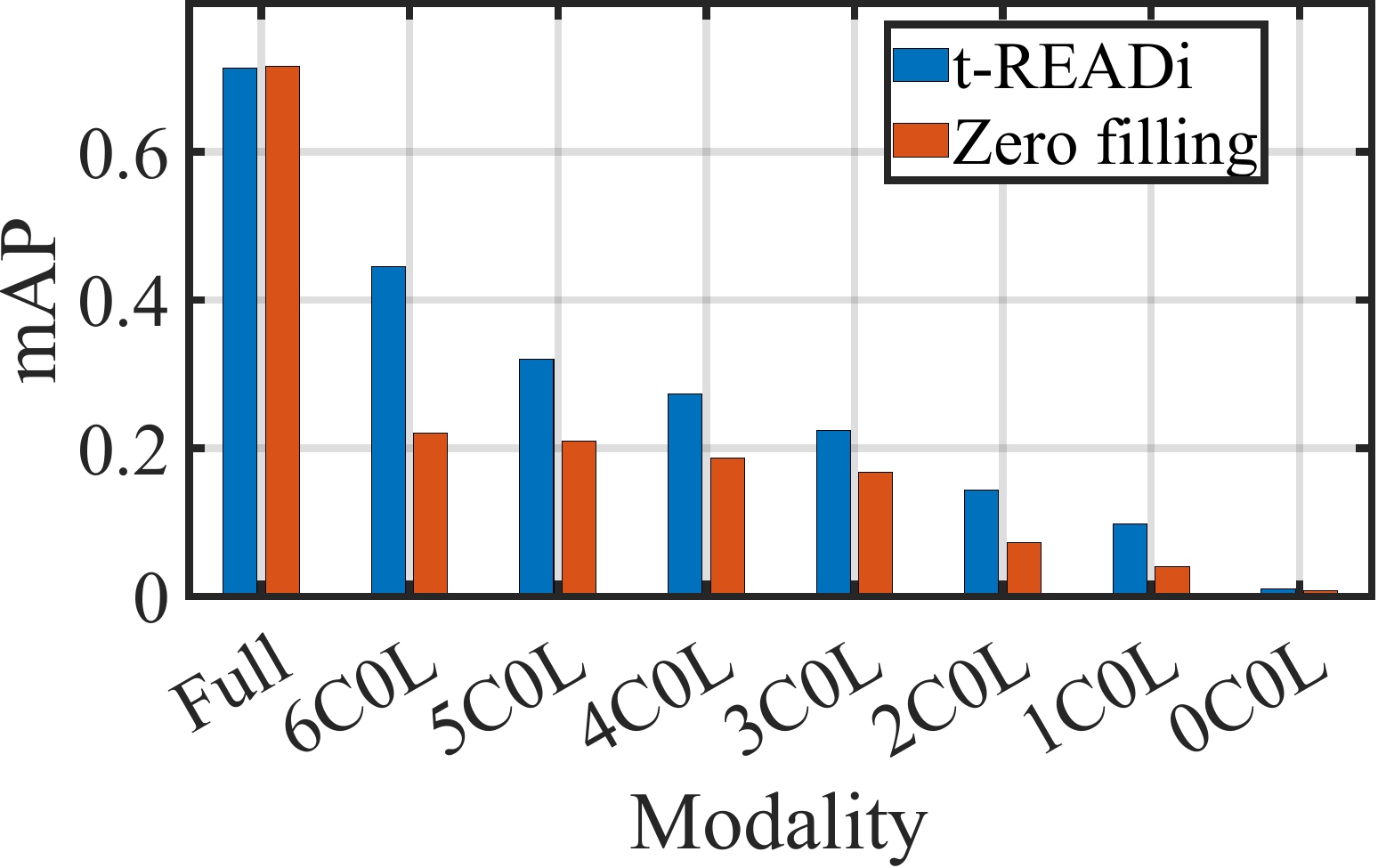}
            \label{subfig:mixing_failure_map}
        }
        % \subfigure[NDS.]{
        %     \includegraphics[width = .5 \linewidth]{tmp_imgs/536_rev.pdf}
        %     \label{subfig:mixing_failure_nds}
        % }
    \hspace{-10pt}
        % \subfigure[mAP.]{
        %     \includegraphics[width = .5 \linewidth]{tmp_imgs/535_rev.pdf}
        %     \label{subfig:mixing_failure_map}
        % }
        \subfigure[NDS.]{
            \includegraphics[width = .5 \linewidth]{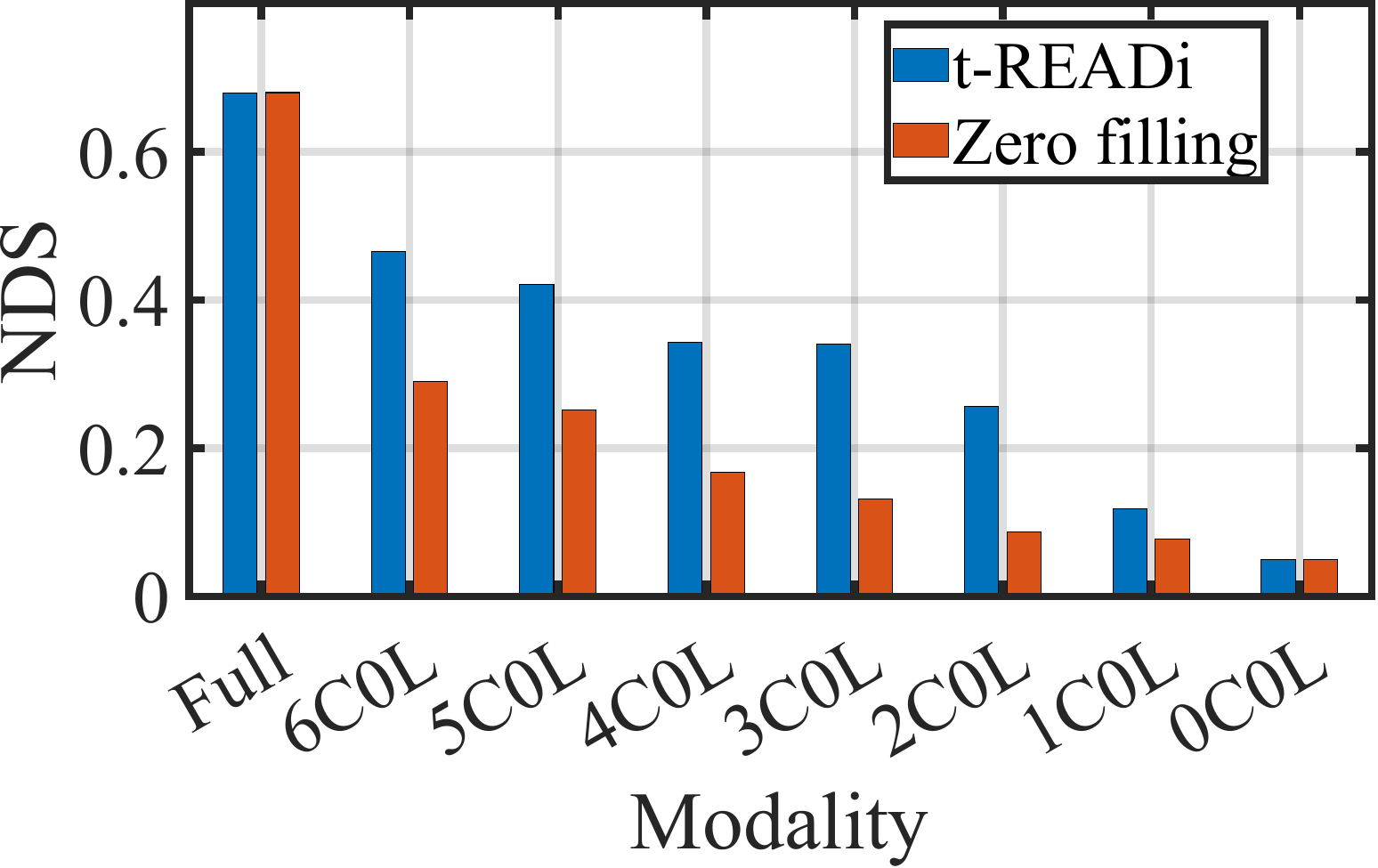}
            \label{subfig:mixing_failure_nds}
        }
    }
    \vspace{-2.5ex}
    \caption{\name compensates for both missing cameras and lidar information loss.}
    \label{fig:mixing_failure}
    \vspace{-2.5ex}
\end{figure}

%\vspace{-1ex}
\paragraph{Mixed missing sensors.} We also evaluate \name's robustness in cases where radar and lidar data are. It can be seen in Figure~\ref{fig:mixing_failure} that even when there is no radar or lidar data, \name still performs significantly better than the zero filling baseline in terms of mAP and NDS. This demonstrates that \name is able to effectively make use of remaining camera data, even when there are only a small number of cameras. Overall, our results show that \name offers both efficiency and robustness in multi-modal perception tasks.

% \begin{figure}[]
%     \centerline{
%         \subfigure[mAP.]{
%             \includegraphics[width = .5 \linewidth]{tmp_imgs/535_rev.pdf}
%             \label{subfig:mixing_failure_map}
%         }
%         \subfigure[NDS.]{
%             \includegraphics[width = .5 \linewidth]{tmp_imgs/536_rev.pdf}
%             \label{subfig:mixing_failure_nds}
%         }
%     }
%     \vspace{-2.5ex}
%     \caption{\name compensates for both missing cameras and lidar information loss.}
%     \label{fig:mixing_failure}
%     \vspace{-2.5ex}
% \end{figure}

%%%%%%%%
%%%%%%%%note by yuhang: subsection ``Performance and Efficiency Tradeoff" will be integrated into ``ablation study" and aforementioned ``OD", ``Seg".
%%%%%%%%

\subsection{Combating Cross-Domain Variations}

\begin{figure}[]
    \centerline{
        \includegraphics[width = .8\linewidth]{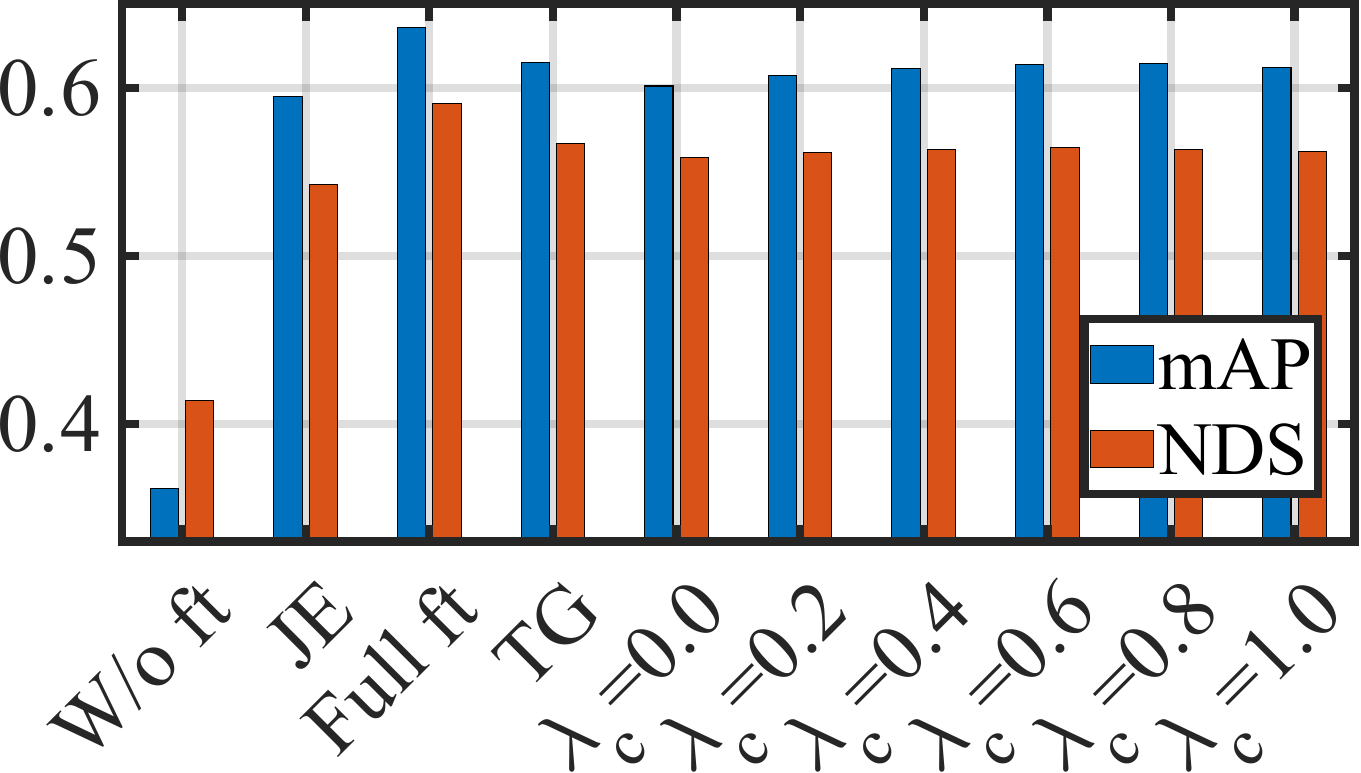}
    }
    \vspace{-2.5ex}
    \caption{\name is capable of handling multi-variation domain with mono-variation domain.}
    \label{fig: cross_domain}
    \vspace{-2.5ex}
\end{figure}

To evaluate \name's ability of handling of cross-domain variation using previously seen mono-domain variation, we extend the experiment in \S~\ref{subsec: OD}. The lidar/camera variation levels are set to $\alpha=0.03$ and $\gamma=0.5$ respectively. Our results are shown in Figure~\ref{fig: cross_domain}. Compared to the full model ft operation in column $3$ and tuning both variation domains together (referred to as \textbf{TG} in legend), \name achieves significant performance with mono-variation domain variants. Specifically, updating the exclusive layers, i.e., modality-specific encoders (referred to as \textbf{JE} in legend), already achieves an impressive performance loss of only $4\sim5$ mAP and NDS compared to full model ft. With interpolation, the local maximum is achieved at $\lambda_{c}=0.8$, reducing the gap to $2\sim2.5$ mAP and NDS loss. The approximation to the TG variant is remarkable, with a marginal gap of only $0.1$ and $0.4$ for mAP and NDS, respectively. In challenging cross-domain variation scenarios, we propose interpolating from mono-variation domains as a simple yet effective method, yielding sub-optimal performance but with greater scalability. This insight has the potential to be extended to more diverse sensor systems and over-the-air variations. When variations from different sensors are superimposed, \name achieves exceptional parameter efficiency compared to full model ft. It combines interpolation with variation-aware adaptation, ensuring scalability: the overall parameter space increases linearly with the number of variation quantization levels, regardless of the number of modalities.
% \name achieves extreme parameter efficient compared to full model ft, when various variation for different sensors superimposed together, with a combination of interpolating and variation-aware adaption. And it is more extendable as the overall parameter space grows linearly with variation quantization levels despite modalities number.
% For the exclusive layers $(\bigcup_{i \in \mathbf{N}}\mathcal{L}_{c_{i}}^{'}) \bigoplus (\bigcup_{j \in \mathbf{N}}\mathcal{L}_{l_{j}}^{'})$ (where $\bigoplus$ operator indicates symmetric difference operation), updating them is straightforward. However, for the overlapping layers $(\bigcup_{i \in \mathbf{N}}\mathcal{L}_{c_{i}}^{'}) \bigcap (\bigcup_{j \in \mathbf{N}}\mathcal{L}_{l_{j}}^{'})$, they are more complicated and we employ linear interpolation using $\lambda_{c} * \mathcal{P}_{c}\left(t\right) + \lambda_{l} * \mathcal{P}_{l}\left(t\right)$, where $t$ belongs to the overlapping layers and $\mathcal{P}_{\{c, l\}}\left(t\right)$ represents the model-specific parameter set for $t$. To simplify this process, we apply the additional constraint $\lambda_{c} + \lambda_{l}=1$.
% To verify whether \name can handle cross domain variation properly with already seen mono domain variation, we extend the experiment in Sec~\ref{subsec: OD} by applying the previous unused camera modality, and the lidar variation level is set to $\alpha=0.03$ while camera variation level is set to $\gamma=0.5$. Consider the separate updated parameter layer collection $\bigcup_{i \in \mathbf{N}}\mathcal{L}_{c_{i}}^{'}$ and $\bigcup_{j \in \mathbf{N}}\mathcal{L}_{l_{j}}^{'}$, for exclusive layers $(\bigcup_{i \in \mathbf{N}}\mathcal{L}_{c_{i}}^{'}) \bigoplus \\ (\bigcup_{j \in \mathbf{N}}\mathcal{L}_{l_{j}}^{'})$ (where $\bigoplus$ operator indicates symmetric difference operation), it is trivial to update them in whole, for overlap layers $(\bigcup_{i \in \mathbf{N}}\mathcal{L}_{c_{i}}^{'}) \bigcap (\bigcup_{j \in \mathbf{N}}\mathcal{L}_{l_{j}}^{'})$, which is much more complicated and we consider a linear interpolation $\lambda_{c} * \mathcal{P}_{c}\left(t\right) + \lambda_{l} * \mathcal{P}_{l}\left(t\right)$, where $t \in (\bigcup_{i \in \mathbf{N}}\mathcal{L}_{c_{i}}^{'}) \bigcap (\bigcup_{j \in \mathbf{N}}\mathcal{L}_{l_{j}}^{'})$, $\mathcal{P}_{\{c, l\}}\left(t\right)$ denotes model-specific parameter set specified by $t$. And here we apply additional constraint of $\lambda_{c} + \lambda_{l}=1$ to simplify this process.
% %%%%%%%%
%%%%%%%%note by yuhang: the brackets are obstacle.
%%%%%%%%

\subsection{Ablation Study and Micro-benchmark}
\label{ssec: ablation}
In this section, we present how the components of \name are fused properly and evaluate the critical hyperparameters to justify our choices. The settings are consistent with Sec~\ref{subsec: OD}, with parameter $\alpha$ set to $0.03$.
% In this section, we will present the components of \name are fused properly, and evaluate the effect of some critical hyperparameter to justify our option. The setting follows Sec~\ref{subsec: OD} and parameter $\alpha$ is set to $0.030$.

\begin{figure}[]
    \centerline{
        \includegraphics[width = .8\linewidth]{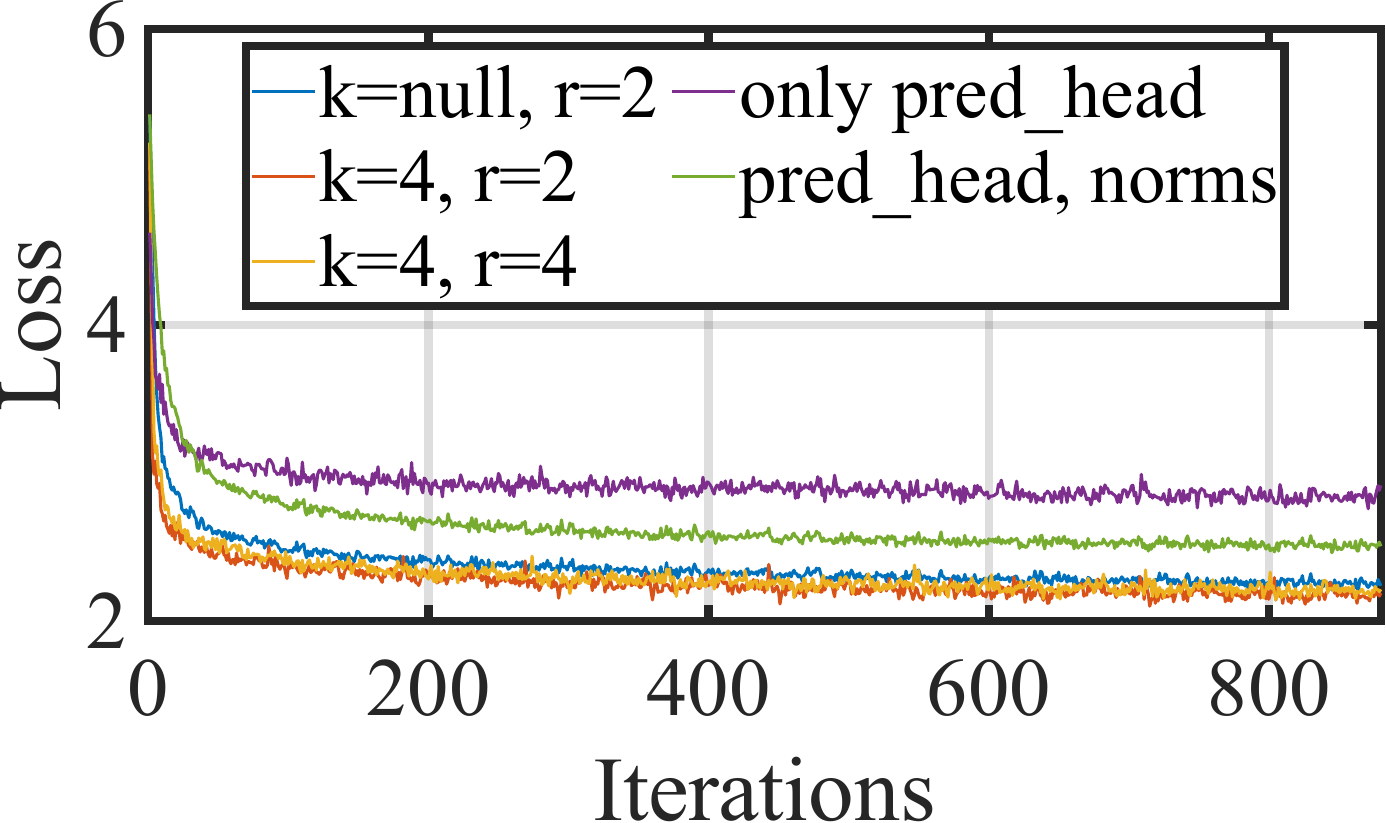}
    }
    \vspace{-2.5ex}
    \caption{Loss profile of various tuning schemes, as parameters of \name vary.}
    \label{fig: ablation_loss}
    \vspace{-2.5ex}
\end{figure}

% \paragraph{Loss profile perspective}
\paragraph{Ablation Study.} We first study the effectiveness of individual module of \name. %The key hyperparameters in this context are the intrinsic rank upper bound $k$ for enhancing the attention module and the squeeze ratio $r$ of adapters for enhancing the residual blocks. 
We set the default values of intrinsic rank upper bound $k$ and squeeze ratio $r$ to 4 and 2, respectively. Figure~\ref{fig: ablation_loss} provides a clear illustration: compared to the common practice of adjusting only a few downstream layers (typically lightweight prediction heads), \name demonstrates its effectiveness and efficiency by prioritizing the adjustment of Batch Normalization (BN) layers as the first step. This approach outperforms the conventional practice, which often reaches a plateau early on. Moreover, injecting transparent tiny modules further boosts \name’s performance. By injecting the tiny modules, \name achieves significant loss reduction in the first epoch compared to conventional schemes that require up to 10 epochs or more.

\paragraph{Performance and Overhead Trade-off} In addition, we conducted an evaluation of the effects of $k$ and $r$ on mAP, NDS, and memory budgets. Table~\ref{tbl: parameter_ver} presents a summary of the results. Here we have $4$ insights: 1) BN contribute a small percentage ($\approx0.12\%$) to memory usage, ensuring \name's memory efficiency. 2) Increasing $r$ for adapter squeezing memory usage but leads to fluctuating and slower convergence, we choose $2$ as the default ratio. 3) \name's low overhead is due to bulky 3D convolution modules, compared to 2D counterparts. This makes adapter less efficient for 2D convolution-intensive scenarios. While with the default configuration, the budget still remains within $5\%$. A no-projection fully connected layer was omitted in earlier versions of the adapter to save space, sacrificing a maginal gain of $0.5$ points. 4) A relatively small $k$ effectively mitigates rank collapse. Larger $k$ (e.g., $8$) worsen performance even more than not using this option. One may complain this improvement is more marginal, while can be attributed to the reduced reliance on the transformer and the minimal impact on the additional budget.

\paragraph{The Effectiveness of \name beyond Adding More Parameters} One might argue that the improvement achieved by \name is simply a result of adding more parameters. However, we show that this is not the case by comparing \name with an alternative configuration that allocates the budget to convolution modules instead of \name modules. The configuration has more parameters than \name, but it performs worse, as shown in the last row of Table~\ref{tbl: parameter_ver}. This demonstrates that \name is effective not because it has more parameters, but because it has a better organization of them.

% \paragraph{More than just additional parameters}
% To justify that the improvement achieved by \name is not simply a result of adding more parameters, we present an alternative organization of the budget in convolution modules (which just makes the model deeper) with a significant surplus. Please refer to the last row in Table~\ref{tbl: parameter_ver}. Despite the additional freely-tuned modules, \name outperforms the configuration with more parameters, highlighting its effectiveness.
% As the end of this section, to justify the improvement achieved by \name is not just the effect of stacking more parameters, we reorganize the budget in convolution modules with much more surplus as the alternative organization, please notice the last row in Table~\ref{tbl: parameter_ver}. Where although these additional modules are tuned freely, much more parameter only yield less competitive performance against \name, thus illustrates the effectiveness of \name.
\begin{table}[t]
    \centering
    \caption{Impact of module's parameters.}
    %\vspace{-1.5ex}
    %\renewcommand{\arraystretch}{0.9}
    \begin{tabular}{c|cccc}
        \hline & mAP & NDS & Exist budget & Add budget \\
        \hline 
        no op           & 31.71 & 38.43 & N/A       & N/A   \\
        just pred       & 37.11 & 41.91 & 0.63\%    & N/A   \\
        k, r=null       & 42.23 & 45.30 & 0.87\%    & N/A   \\
        \hline
        \textbf{k=4, r=2}& 47.90 & 49.17 & 0.87\%    & 1.83\%\\
        \hline
        k=4, r=4        & 47.23 & 48.62 & 0.87\%    & 1.00\%\\
        k=8, r=2        & 47.67 & 48.87 & 0.87\%    & 2.00\%\\
        alt org         & 46.43 & 47.62 & 0.87\%    & 3.51\%\\
        \hline
    \end{tabular}
    \vspace{-2.5ex}
    \label{tbl: parameter_ver}
\end{table}
\section{Related Work}\label{sec:related_work}
\paragraph{Multi-modal Sensor Fusion}

Multi-modal fusion is essential for autonomous driving systems' perception, combining data from multiple sensors to enhance accuracy and robustness. Major paradigms for multi-modal fusion can be categorized into early, deep, and late fusion based on the \textit{stage} (raw data, feature, proposal) where the fusion occurs~\cite{huang2022multi,xie2020pi,Chen2023RFMic,shi2022vips}. Each paradigm can employ different fusion policies, ranging from simple concatenation or element-wise addition to more sophisticated methods using differentiable and learnable functions. These traditional paradigms have limitations as they fuse all modalities simultaneously. Recent works have attempted to fuse modalities selectively, achieving more robust performance~\cite{Chen2022Selective}.

%\vspace{-1ex}
\paragraph{Multi-modal Detection \& Segmentation}
Detection and segmentation are vital tasks in autonomous driving perception. Multi-modal 3D detection and segmentation techniques leverage complementary modalities due to the deficiencies of the camera's depth information and the lidar's semantic information. For 3D multi-modal object detection, \cite{YingweiLi2022DeepFusionLD} uses cross-modal attention to fuse camera and lidar features. \cite{guan2022deepmix} combines edge-assisted 2D detection with on-device 3D boxes for lightweight, hybrid 3D object detection. For 3D semantic segmentation, \cite{AngelaDai20183DMVJ3} projects 2D CNN-extracted image features to 3D space and fuses them with lidar data for voxel-wise segmentation. In \cite{HangSu2018SPLATNetSL}, permutohedral lattice representation fuses multi-modal data for 3D semantic segmentation.

%\vspace{-1ex}
\paragraph{Combating Missing Modalities}
Multimodal systems face challenges with missing modalities, negatively impacting performance. Researchers proposed methods for missing modality imputation and performance improvement. \cite{tran2017missing} uses cascaded residual autoencoder (CRA) for missing modality imputation, stacking residual autoencoders (RAs) that iteratively output the difference between incomplete and complete data. \cite{pham2019found} learns robust joint representations by translating between modalities, using a translation network to establish a consistent, robust common space. This approach allows the system to handle missing modalities at test time by inferring them from available modalities.
\paragraph{Networking and system support for AI inference} 
The scope of support for AI spans the entire lifecycle of AI services, from \textit{data collection} and \textit{model training} to \textit{model inference}\cite{Shen2022NetIntel}. In the \textit{inference} phase, multi-dimensional QoS can be achieved. To ensure power efficiency and low latency for end users with limited computing capabilities and battery power, model partition techniques divide a DNN into multiple sub-models, embedding them into different network nodes to conduct inference collaboratively\cite{zhang2021Autodidactic,Teerapittayanon2017DDNN}. Additionally, model compression leverages techniques such as weight pruning~\cite{han2016deep,Lee2022DNNcomp}, parameter quantization~\cite{zhu2017trained}, and encoding~\cite{Reagen2017WeightlessLW} to fit DNNs into \textit{Application Specific Integrated Circuit} (ASIC), thereby overcoming memory and power constraints~\cite{han2016eie}. However, while a typical driving system is not as energy-constrained as energy-harvesting systems, real-time reaction to traffic conditions is critical, necessitating that processing always meets strict deadlines. Communication between the vehicle and edge servers (e.g., smart lampposts) is too costly, as an autonomous driving vehicle can generate hundreds of megabytes of raw sensor data per second. Most model compression techniques and ASIC designs focus on "single-branch" networks like AlexNet~\cite{Krizhevsky2012ImageNetCW}, VGG~\cite{simonyan2015deep}, while thoroughly compressing or designing ASICs for "multi-branch" networks like ResNet~\cite{he2016deep} is challenging. This is even more complex for attention-based advanced networks, which are the de facto solutions for autonomous driving~\cite{thinkautonomous}.

\paragraph{Efficient Fine-Tuning for LLM}
% \rev{Parameter efficient fine-tuning as a mainstream technique,} has variants like low-rank adaptation~\cite{hu2021lora} which update every low rank matrices at every iteration, 
% \rev{Several LLM turning method can achieve parameter efficient target.} Low-Rank Adaptation~\cite{hu2021lora} updates every parameter in every low-rank matrices at every iteration. Prefix-tuning~\cite{li2021prefixtuning} adds a series of trainable vectors, known as prefix tokens, to each layer in an LLM, these tokens can be tailored to specific tasks. Different from prefix-tuning, prompt tuning tokens can be inserted either as a prefix or anywhere within the input tokens. Some advanced technique~\cite{liu2022ptuning} introduces continuous prompts at each layer rather than at the input layer only, which boosts performance  for tasks related to natural language understanding. Beside, memory efficient fine-tuning is also a critical topic. The whole model can be first quantized or decomposited into low-precision data types variant and only LoRAs corresponding to specific downstream tasks are tuned~\cite{dettmers2023qlora}. In~\cite{kim2023memoryefficient}, scalar vectors as the byproduct of decomposition are updated as well. \name is greatly motivated by LoRA in terms of technique thanks to its scalability.
Several methods for tuning LLM are proposed to achieve parameter efficiency. Low-Rank Adaptation (LoRA)~\cite{hu2021lora} updates every parameter in low-rank matrices at each iteration. Prefix-tuning~\cite{li2021prefixtuning} adds a series of trainable vectors, known as prefix tokens, to each layer in LLM; these tokens can be tailored to specific tasks. Unlike prefix-tuning, prompt-tuning allows tokens to be inserted either as a prefix or anywhere within the input tokens. Advanced techniques, such as those introduced in~\cite{liu2022ptuning}, incorporate continuous prompts at each layer rather than solely at the input layer, boosting performance in natural language understanding tasks. Additionally, memory efficiency is also a critical topic. Models can first be quantized or decomposed into low-precision data types, and only the LoRAs corresponding to specific downstream tasks are tuned~\cite{dettmers2023qlora}. In~\cite{kim2023memoryefficient}, scalar vectors resulting from decomposition are also updated. Our method is greatly inspired by LoRA, thanks to its scalability.
% \subsection{Discussion} \label{ssec:discussion}
% Radar
\section{Discussion}\label{sec:discussion}
In this section, we discuss the existing limitations of \name's pipeline and potential techniques that could enhance \name when integrated.

\paragraph{High-Fidelity variations}
\name employs an \textit{online} version of camera modality variation to conserve disk space. However, an \textit{offline} approach, such as estimating the blur kernel and then generating artificial images accordingly, would be a more comprehensive method. Concurrently with \name, a study by \cite{Kong2023Robo3D} also explores lidar-perception robustness, simulating blur effects for lidar by introducing jitter noise subject to certain distributions.

% \paragraph{Compatibility with compressing techniques}
% \name achieve light memory footprint by deduplicate the majority parameters among various variants share the same structure and it doesn't rely on compressing techniques. Some basic techniques like weight pruning~\cite{han2016deep} can be applied into \name directly and boost the performance in memory-efficiency aspect further.

\paragraph{Additional Modalities}
\name currently focuses on camera and lidar binary modalities. When additional modalities are introduced, simple fusion policies, such as concatenation followed by element-wise multiplication~\cite{Bijelic2020SeeingTF}, have been shown to be less robust~\cite{Chen2022Selective}. Consequently, \name's robustness can be enhanced further through the adoption of more advanced fusion methods.

\paragraph{AV-DNN Accelerators}
\name functions akin to a gear mechanism, facilitating the swift switching of AV-DNNs rather than serving as an inference accelerator. 
Although it only marginally compromises inference latency, as discussed in \S~\ref{subsec: IL}, we have identified that the primary bottleneck in current inference latency resides within the lidar quantization pipeline, which falls outside the scope of \name. Addressing this bottleneck can be approached through software- or hardware-based optimizations, such as refining quantization parameters or offloading the quantization preprocessing to ASICs. Both methods are compatible with \name and hold significant potential for accelerating this stage.

% 0.25 pages
%\section{Discussion} \label{sec:discussion}
%\input{7_discussion}
% 0.25 page
\section{Conclusion}\label{sec:conclusion}
Taking an important step towards full driving automation, we have proposed \name in this paper for robust and efficient multimodal inference for autonomous driving. Employing a novel partial weight adaptation mechanism and a data imputation method by autoencoder, \name gracefully handles the heterogeneous data caused by sensor parameter variation and missing modality, thus releasing its full potential in the vehicle detection task. With extensive experiments under highly heterogeneous scenarios and comparisons with other baselines, we demonstrate the promising performance of \name in vehicle detection for autonomous driving. Overall, we believe that \name represents an important step forward in the development of robust and efficient multimodal inference for autonomous driving. As a potential future direction, we are looking forward to extending our t-READ to various applications such as distributed learning systems~\cite{lin2024adaptsfl,hu2024accelerating,zhang2024satfed,lyu2023optimal,lin2024efficient,zhang2024fedac}, multi-agent collaborative intelligence~\cite{luo2024hurry,zhao2024leo,yuan2024satsense,yuan2023graph}.
% We hope that our work will inspire further research in this exciting and important area.

% Though the datasets used for \name contain radar data, they provide post-processing data (via peak detection only) instead of raw data, which leads to extremely sparse radar point cloud density ($< 1\%$ of lidar point cloud density) not qualified to the tasks,
% % (i.e., object detection, semantic segmentation), 
% we only consider the camera and lidar as the inputs for multimodal fusion, while leaving the fusion of radar modal for future work.
% % of \sysname.
\ifCLASSOPTIONcompsoc
  \section*{Acknowledgments}
\else
  \section*{Acknowledgment}
\fi
This work is supported by National Key R\&D Program of China (No. 2023YFE0116600), the National Natural Science Foundation of China (Grant No. 62202276, 62232010), Shandong Science Fund for Excellent Young Scholars (No. 2022HWYQ-038), Shandong Science Fund (No. 2023TSGC0105), the research start-up grant from the Southern University of Science and Technology, and Shenzhen Higher Education Institutions Stable Support Program (No. 20231120215201001).
% We would like to thank the anonymous reviewers for their insightful comments. This work is supported by the National Key Research and Development Program of China (Grant No. 2021YFB3100400), National Natural Science Foundation of China (Grant No. 62202276, 61832012 and 62350410480), the Shandong Science Fund for Excellent Young Scholars (Grant No. 2022HWYQ-038), and the Future Young Scholars Project of Shandong University.

\bibliographystyle{IEEEtran}
\bibliography{reference}

\end{document}